\documentclass{article} 
\usepackage{iclr2020_conference,times}

\usepackage{amsmath,amsfonts,bm}

\def\eqref#1{equation~\ref{#1}}

\def\1{\bm{1}}

\DeclareMathAlphabet{\mathsfit}{\encodingdefault}{\sfdefault}{m}{sl}
\SetMathAlphabet{\mathsfit}{bold}{\encodingdefault}{\sfdefault}{bx}{n}

\def\sN{{\mathbb{N}}}

\def\sR{{\mathbb{R}}}

\usepackage{natbib}
\usepackage{hyperref}
\usepackage{url}
\usepackage{amsfonts}       

\usepackage{float}
\usepackage{graphicx}  

\usepackage{comment}

\usepackage[bottom]{footmisc}

\newcommand{\note}[1]{}  

\usepackage{array}
\newcolumntype{L}[1]{>{\raggedright\let\newline\\\arraybackslash\hspace{0pt}}m{#1}}
\newcolumntype{C}[1]{>{\centering\let\newline\\\arraybackslash\hspace{0pt}}m{#1}}
\newcolumntype{R}[1]{>{\raggedleft\let\newline\\\arraybackslash\hspace{0pt}}m{#1}}

\usepackage{csquotes}
\usepackage{multirow}
\usepackage{booktabs}
\usepackage{makecell}
\usepackage{amssymb}
\usepackage{amsmath}
\usepackage[ruled,vlined]{algorithm2e}

\title{Intrinsically Motivated Discovery of Diverse Patterns in Self-Organizing Systems}

\author{%
  Chris Reinke\thanks{Equal contribution.} , Mayalen Etcheverry\footnotemark[1] ,  Pierre-Yves Oudeyer \\
  Flowers Team\\
  Inria, Univ. Bordeaux, Ensta ParisTech (France) \\
  \texttt{\{chris.reinke,mayalen.etcheverry,pierre-yves.oudeyer\}@inria.fr} \\
  }

\iclrfinalcopy 
\begin{document}

\maketitle

\begin{abstract}
In many complex dynamical systems, artificial or natural, one can observe self-organization of patterns emerging from local rules. 
Cellular automata, like the Game of Life (GOL), have been widely used as abstract models enabling the study of various aspects of self-organization and morphogenesis, such as the emergence of spatially localized patterns. 
However, findings of self-organized patterns in such models have so far relied on manual tuning of parameters and initial states, and on the human eye to identify \enquote{interesting} patterns. 
In this paper, we formulate the problem of automated discovery of diverse self-organized patterns in such high-dimensional complex dynamical systems, as well as a framework for experimentation and evaluation. 
Using a continuous GOL as a testbed, we show that recent intrinsically-motivated machine learning algorithms (POP-IMGEPs), initially developed for learning of inverse models in robotics, can be transposed and used in this novel application area. 
These algorithms combine intrinsically-motivated goal exploration and unsupervised learning of goal space representations.
Goal space representations describe the \enquote{interesting} features of patterns for which diverse variations should be discovered.
In particular, we compare various approaches to define and learn goal space representations from the perspective of discovering diverse spatially localized patterns. 
Moreover, we introduce an extension of a state-of-the-art POP-IMGEP algorithm which incrementally learns a goal representation using a deep auto-encoder, and the use of CPPN primitives for generating initialization parameters.
We show that it is more efficient than several baselines and equally efficient as a system pre-trained on a hand-made database of patterns identified by human experts. \let\thefootnote\relax\footnote{Source code and supplementary material at \url{https://automated-discovery.github.io/}}

\end{abstract}

\section{Introduction}

Self-organization of patterns that emerge from local rules is a pervasive phenomena in natural and artificial dynamical systems \citep{ball1999self}.
It ranges from the formation of snow flakes, spots and rays on animal's skin, to spiral galaxies. 
Understanding these processes has boosted progress in many fields, ranging from physics to biology \citep{camazine2003self}. 
This progress relied importantly on the use of powerful and rich abstract computational models of self-organization \citep{kauffman1993origins}.
For example, cellular automata like Conway's Game of Life (GOL) have been used to study the emergence of spatially localized patterns (SLPs) \citep{gardener1970mathematical}, informing theories of the origins of life \citep{gardener1970mathematical, beer2004autopoiesis}.
SLPs, such as the famous glider in GOL \citep{gardner1983wheels}, are self-organizing patterns that have a local extension and can exist independently of other patterns.
However, finding such novel self-organized patterns, and mapping the space of possible emergent patterns, has so far relied heavily on manual tuning of parameters and initial states.
Moreover, the dependence of this exploration process on the human eye to identify \enquote{interesting} patterns is strongly limiting further advances. 

We formulate here the problem of automated discovery of a \textit{diverse} set of self-organized patterns in such high-dimensional, complex dynamical systems. This involves several challenges. 
A first challenge consists in determining a representation of patterns, possibly through learning, enabling to incentivize the discovery of diverse \enquote{interesting} patterns. Such a representation guides exploration by providing a measure of (di-)similarity between patterns. This problem is particularly challenging in domains where patterns  are high-dimensional as in GOL. In such cases, scientists have a limited intuition about what useful features are and how to represent them. Moreover, low-dimensional representations of patterns are needed for human browsing and the visualization of the discoveries. Representation learning shall both guide exploration, and be fed by self-collected data. 

A second challenge consists in how to automate exploration of high-dimensional, continuous initialization parameters to discover efficiently \enquote{interesting} patterns, such as SLPs, with a limited budget of experiments. 
Sample efficiency is important to enable the later use of such discovery algorithms for physical systems \citep{grizou2020curious}, where experimental time and costs are strongly bounded. 
For example, in the continuous GOL used in this paper as a testbed, initialization consists in determining the values of a real-valued, high-dimensional $256\times256$ matrix besides 7 additional dynamics parameters.
The possible variations of this matrix are too large for a simple random sampling to be efficient.
More structured methods are needed.


To address these challenges, we propose to leverage and transpose recent intrinsically motivated learning algorithms, within the family of population-based Intrinsically Motivated Goal Exploration Processes (POP-IMGEPs - denoted simply as IMGEPs below, \cite{baranes2013active,pere2018unsupervised}).
They were initially designed to enable autonomous robots to explore and learn what effects can be produced by their actions, and how to control these effects. 
IMGEPs self-define goals in a goal space that represents important features of the outcomes of actions, such as the position reached by an arm.
This allows the discovery of diverse novel effects within their goal representations.
It was recently shown how deep neuronal auto-encoders enabled unsupervised learning of goal representations in IMGEPs from raw pixel perception of a robot's visual scene \citep{laversanne2018curiosity}.
We propose to use a similar mechanism for automated discovery of patterns by unsupervised learning of a low-dimensional representation of features of self-organized patterns. This removes the need for human expert knowledge to define such representations.

Moreover, a key ingredient for sample efficient exploration of IMGEPs for robotics has been the use of structured motion primitives to encode the space of body motions \citep{pastor2013dynamic}.
We propose to use a similar mechanism to handle the generation of structured initial states in GOL-like complex systems, based on specialized recurrent neural networks (CPPNs) \citep{stanley2006exploiting}. 


In summary, we provide in this paper the following contributions: 
\begin{itemize}
\item We formulate the problem of automated discovery of diverse self-organized patterns in high-dimensional and complex game-of-life types of dynamical systems.
\item We show how to transpose POP-IMGEPs algorithms to address the associated joint challenge of learning to represent interesting patterns and discovering them in a sample efficient manner.
\item We compare various approaches to define or learn goal space representations for the sample efficient discovery of diverse SLPs in a continuous GOL testbed.
\item We show that an extension of a state-of-the-art POP-IMGEP algorithm, with incremental learning of a goal space using a deep auto-encoder, is equally efficient than a system pretrained on a hand-made database of patterns.
\end{itemize}

\section{Related Work}

\paragraph{Automated Discovery in Complex Systems}
Automated processes have been widely used to explore complex dynamical systems.
For example, evolutionary algorithms have been applied to search specific patterns or rules of cellular automata \citep{mitchell1996evolving,sapin2003research}.
However, their objective is to optimize a specific goal instead of discovering a diversity of patterns.
Another line of experiments represent active inquiry-based learning strategies which query which set of experiments to perform to improve a system model, i.e.\ a mapping from parameters to the system outcome.
Such strategies have been used in biology \citep{king2004functional,king2009automation}, chemistry \citep{raccuglia2016machine,reizman2016suzuki, duros2017human} and astrophysics \citep{richards2011active}.
However, these approaches have relied on expert knowledge, and focused on automated optimization of a pre-defined target property.
Here, we are interested to automatically discover and map a diversity of unseen patterns without prior knowledge of the system. 
An exception is the concurrent work of \cite{grizou2020curious}, which showed how a simple POP-IMGEP algorithm could be used to automate discovery of diverse patterns in oil-droplet systems. 
However, it used a low-dimensional input space, and a hand-defined low-dimensional representation of goal spaces, identified as a major limit of the system. 


\paragraph{Intrinsically motivated learning}
Intrinsically-motivated learning algorithms \citep{baldassarre2013intrinsically, baranes2013active} autonomously organize an agent's exploration curriculum in order to discover efficiently a maximally diverse set of outcomes the agent can produce in an unknown environment.
They are inspired from the way children self-develop open repertoires of skills and learn world models.  
Intrinsically Motivated Goal Exploration Processes (IMGEPs) \citep{baranes2013active, forestier2017intrinsically} are a family of curiosity-driven algorithms developed to allow efficient exploration of high-dimensional complex real world systems. 
Population-based versions of these algorithms, which leverage episodic memory, hindsight learning, and structured dynamic motion primitives to parameterize policies, enable sample efficient acquisition of high-dimensional skills in real world robots \citep{forestier2017intrinsically,rolf2010goal}. The discovered repertoires of diverse behaviors can be reused to solve hard exploration downstream tasks \citep{colas2018gep,conti2018improving}.
Recent work \citep{laversanne2018curiosity, pere2018unsupervised} studied how to automatically learn the goal representations with the use of deep variational autoencoders. 
However, training was done passively and in an early stage on a precollected set of available observations. 
Recent approaches \citep{nair2018visual, pong2019skew} introduced the use of an online training of \textit{variational autoencoders} VAEs to learn the important features of a goal space similar to the methods in this paper.
However, these approaches focused on the problem of sequential decisions in MDPs, incurring a cost on sample efficiency.
This problem is observed in various intrinsically motivated RL approaches \citep{bellemare2016unifying,burda2018exploration}.
These approaches are orthogonal to the automated discovery framework considered here with independent experiments allowing the use of memory-based sample efficient methods.
A related family of algorithms in evolutionary computation is novelty search \citep{lehman2008exploiting} and quality-diversity algorithms \citep{pugh2016quality}, which can be formalized as special kinds of population-based IMGEPs.

\paragraph{Representation learning}
We are using representation learning methods to learn autonomously goal spaces for IMGEPs.
Representation learning aims at finding low-dimensional explanatory factors representing high-dimensional input data \citep{bengio2013representation}.
It is a key problem in many areas in order to understand the underlying structure of complex observations. 
Many state-of-the-art methods \citep{tschannen2018recent} have built on top of Deep VAEs \citep{kingma2013auto}, using varying objectives and network architectures. 
However, studies of the interplay between representation learning and autonomous data collection through exploration of an environment have been limited so far. 


\section{Algorithmic Methods for Automated Discovery}


\subsection{Population-based Intrinsically Motivated Goal Exploration Processes}

\begin{figure}[t!]
  \centering
  \includegraphics[scale=1]{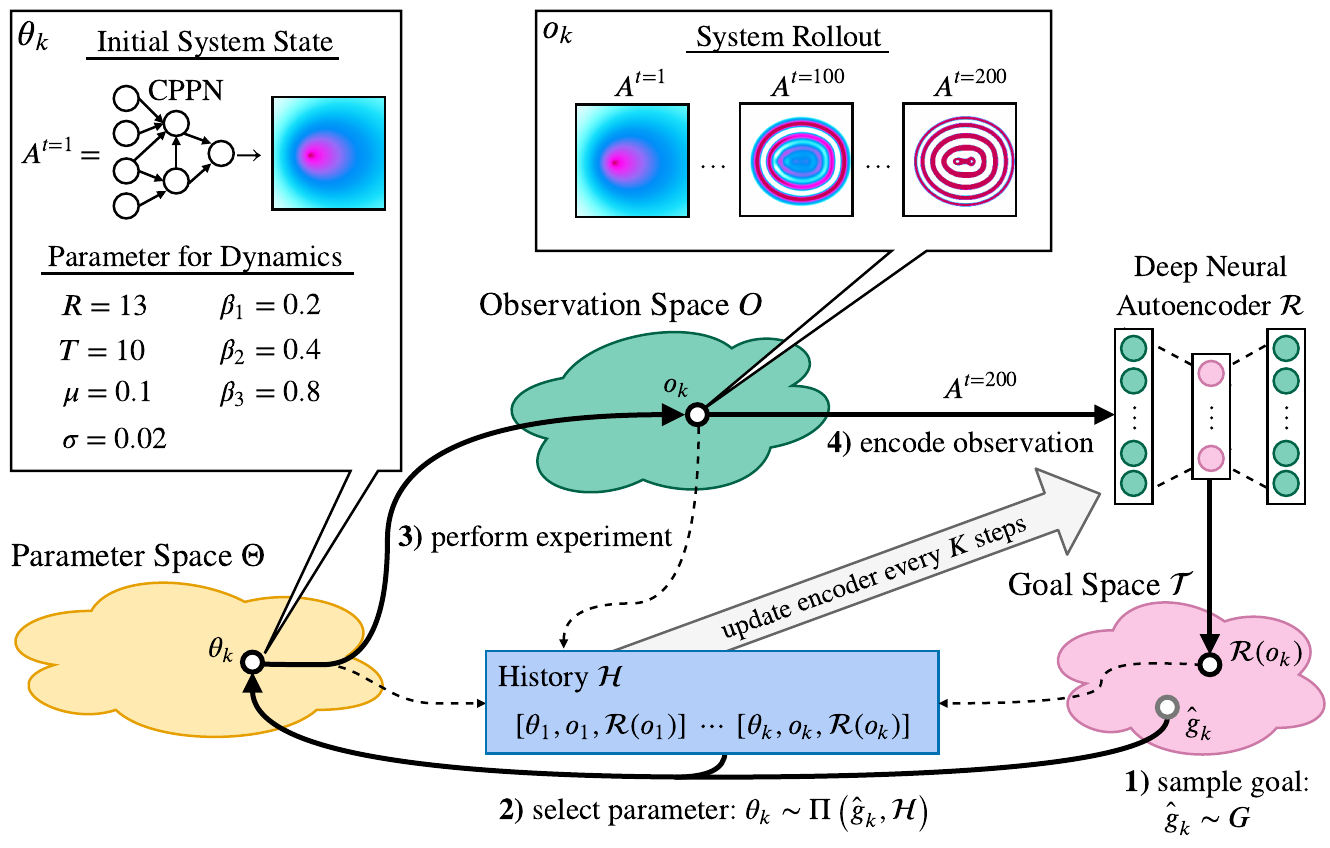}
  \caption{Population-based intrinsically motivated goal exploration process with incremental learning of a goal space (IMGEP-OGL algorithm) used to explore a continuous GOL model.}
  \label{fig:imgep}
\end{figure}

An IMGEP is an algorithmic process generating a sequence of experiments to explore the parameters of a system by targeting self-generated goals (Fig.~\ref{fig:imgep}). 
It aims to maximize the diversity of observations from that system within a budget of $N$ experiments. 
In population-based IMGEPs, an explicit memory of the history $\mathcal{H}$ of experiments and observations is used to guide the process.

The systems are defined by three components.
A parameter space $\Theta$ corresponding to the controllable system parameters $\theta$.
An observation space $O$ where an observation $o$ is a vector representing all the signals captured from the system.
For this paper, the observations are a time series of images which depict the morphogenesis of activity patterns.  
Finally, an unknown environment dynamic $D$: $\Theta \rightarrow O$ which maps parameters to observations.

To explore a system, an IMGEP uses a goal space $\mathcal{T}$ that represents relevant features of its observations,
computed by an encoding function $\hat{g} = \mathcal{R}(o)$.
For an exploration of patterns, such features may describe their form or extension.
The exploration process iterates $N$ times through: 
1) sample a goal from a goal sampling distribution $g \sim G(\mathcal{H})$; 
2) infer corresponding parameter $\theta$ using a parameter sampling policy $\Pi = \Pr(\theta; g, \mathcal{H})$;
3) roll-out an experiment with $\theta$, observe outcome $o$, compute encoding $\mathcal{R}(o)$;
4) store $(\theta,o ,\mathcal{R}(o))$ in history $\mathcal{H}$.
The parameter sampling policy $\Pi$ and in some cases the goal sampling distribution $G$ take into account previous explorations stored in history $\mathcal{H}$.
Therefore, the history is first populated through exploring $N_{init}$ randomly sampled parameters.
After this initial phase the described intrinsically motivated goal exploration process starts.

Different goal and parameter sampling mechanisms can be used within this architecture \citep{baranes2013active,forestier2016modular}. 
In the experiments below, goals are sampled uniformly over a hyperrectangle defined in $\mathcal{T}$.
The hyperrectangle is chosen large enough to allow a sampling of a large goal diversity. 
The parameters are sampled by 1) given a goal, selecting the parameter from the history whose corresponding outcome is most similar in the goal space; 2) then mutating it by a random process. 

\subsection{Online Learning of Goal Spaces with Deep Auto-Encoders}

For IMGEPs the definition of the goal space $\mathcal{T}$ and its corresponding encoder $\mathcal{R}$ are a critical part, because it biases exploration of the target system.
One approach is to define a goal space by selecting features manually, for example by using computer vision algorithms to detect the positions of a pattern and its form.
The diversity found by the IMGEPs will then be biased along these pre-defined features.
A limit of this approach is its requirement of expert knowledge to select helpful features, particularly problematic in environments where experts do not know in advance what features are important, or how to formulate them.

Another approach is to learn goal space features by unsupervised representation learning. 
The aim is to learn a mapping $\mathcal{R}(o)$ from the raw sensor observations $o$ to a compact latent vector $\mathbf{z} \in \mathbb{R}^d$. 
This latent mapping can be used as a goal space where a latent vector $\mathbf{z} = g$ is interpreted as a goal.
Previous IMGEP approaches already learned successfully their goal spaces with VAEs \citep{laversanne2018curiosity, pere2018unsupervised}.
However, the goal spaces were learned before the start of the exploration from a prerecorded dataset of observations from the target environment. 
During the exploration the learned representations were kept fixed.
A problem with this pretraining approach is that it limits the possibilities to discover novel patterns beyond the distribution of pretraining examples, and requires expert knowledge.

\begin{algorithm}[t!]
\DontPrintSemicolon
Initialize goal space encoder VAE $\mathcal{R}$ with random weights\\
\For{$i\leftarrow 1$ \KwTo $N$}{
\If(\tcp*[f]{Initial random iterations to populate $\mathcal{H}$}){$i<N_{init}$} {
Sample $\theta \sim \mathcal{U}(\Theta)$
}
\Else(\tcp*[f]{Intrinsically motivated iterations}){
Sample a goal $g \sim \mathcal{G}(\mathcal{H})$ based on space represented by $\mathcal{R}$ \\
Choose $\theta \sim \Pi(g, \mathcal{H})$  \\
}
Perform an experiment with $\theta$ and observe $o$ \\
Encode reached goal $\hat{g} = \mathcal{R}(o)$\\

Append $(\theta, o, \hat{g})$ to the history $\mathcal{H}$ \\
\BlankLine
\If(\tcp*[f]{Periodically train the network}){$i \mod K == 0$} {
\For{E epochs} {Train $\mathcal{R}$ on observations in $\mathcal{H}$ with importance sampling}
\For(\tcp*[f]{Update the database of reached goals}){$(\theta, o, \hat{g}) \in \mathcal{H}$}{ $\mathcal{H}[\hat{g}] \gets \mathcal{R}(o)$}
}
}
\caption{IMGEP-OGL}
\label{algo:IMGEP-OGL}
\end{algorithm}

In this paper we attempt to address this problem by continuously adapting the learned representation to the novel observations encountered during the exploration process. 
For this purpose, we propose an online goal space learning IMGEP (IMGEP-OGL), which learns the goal space incrementally during the exploration process (Algorithm~\ref{algo:IMGEP-OGL}).
The training procedure of the VAE is integrated in the goal sampling exploration process by first initializing the VAE with random weights (Appendix \ref{c:sm_imgep_learned_goalspaces}). The VAE network is then trained every $K$ explorations for $E$ epochs on the observation collected in the history $\mathcal{H}$. 
Importance sampling is used to give more weight to recently discovered patterns by using a weighted random sampler. 
It samples for 50\% of the training batch samples patterns from the last $K$ iterations and for the other 50\% patterns from all other previous iterations.

\subsection{Structuring the parameter space in IMGEPs: from DMPs to CPPNs}
\label{c:methods_cppn}

A key role in the generation of patterns in dynamical systems is their initial state $A^{t=1}$.
IMGEPs sample these initial states and apply random perturbations to them during the exploration. 
For the experiments in this paper this state is a two-dimensional grid with $256\times256$ cells.
Performing directly a random sampling of the $256\times256$ grid cells results in initial patterns that resemble white noise. 
Such random states result mainly in the emergence of global patterns that spread over the whole state space, complicating the search for spatially localized patterns. 
This effect is analogous to a similar problem in the exploration of robot controllers. Direct sampling of actions for individual actuators at a microscopic time scale is usually inefficient.
A key ingredient for sample efficient exploration has been the use of structured primitives (dynamic motion primitives - DMPs) to encode the space of possible body motions \citep{pastor2013dynamic}. 

We solved the sampling problem for the initial states by transposing the idea of structured primitives.
Indeed, \enquote{actions} consist here in deciding the parameters of an experiment, including the initial state.
We propose to use compositional pattern producing networks (CPPNs) \citep{stanley2006exploiting} to produce structured initial patterns similar do DMPs.
CPPNs are recurrent neural networks that allow the generation of structured initial states. (Appendix~\ref{c:sm_sampling_lenia_parameter}, Fig.~\ref{fig:cppn_evolution_examples}).
They are defined by their network structure (number of neurons, connections between neurons) and their connection weights. 
Thus, instead of using directly an initial state as part of the parameters $\theta$ to control the dynamical system, a CPPN is used.
If a system roll-out for the parameters $\theta$ are performed, then the initial state $A^{t=1}$ is generated by the CPPN. 
Moreover, instead of sampling and mutating directly an initial state, the weights and structure of the CPPN are randomly generated and mutated. 
Please note, the number of parameters in $\theta$ is therefore not fixed and open-ended (yet starts small) because the structure and the number of weights of randomly sampled and mutated CPPNs differ.

\newpage
\section{Experimental methods}

We describe here the continuous Game of Life (Lenia) we use as a testbed representing a large class of high-dimensional dynamical systems, as well as the experimental procedures, the evaluation methods used to measure diversity and detect SLPs, and the used algorithmic baselines and ablations.
Implementation details and parameter settings for all procedures are given in Appendix~ \ref{c:sm_implementation_details}.  

\subsection{Continous Game of Life as a testbed}
\label{c:experiments_lenia}

\begin{figure}[t]
\centering
\setlength\tabcolsep{1pt}
\renewcommand{\arraystretch}{0.5}
\begin{tabular}{@{} cc c|c cc c|c cc @{}}
 \multicolumn{2}{c}{\makecell{(a) Evolution in Lenia \\ from CPPN to animal}}  &
 &
 &
 \multicolumn{2}{c}{\makecell{(b) Lenia animals \\ discovered by IMGEP-OGL}}  &
 &
 &
 \multicolumn{2}{c}{\makecell{(c) Lenia animals \\ discovered by \cite{chan2019lenia}}}\\ 
 
 \fbox{\includegraphics[width=2.1cm]{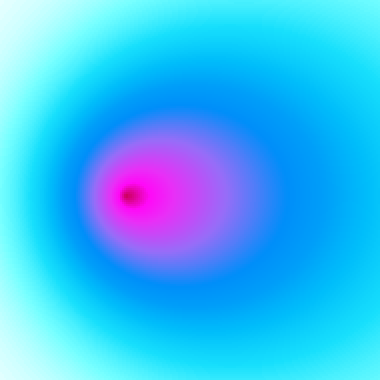}} &
 \fbox{\includegraphics[width=2.1cm]{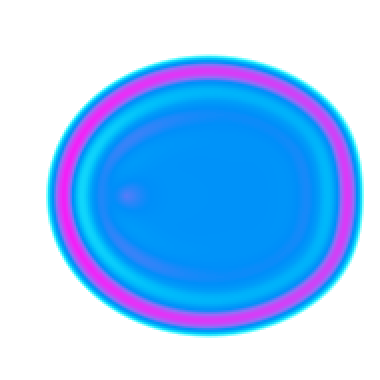}} &
 &
 &
 \fbox{\includegraphics[width=2.1cm]{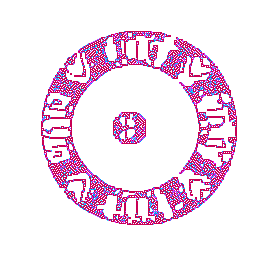}} &
 \fbox{\includegraphics[width=2.1cm]{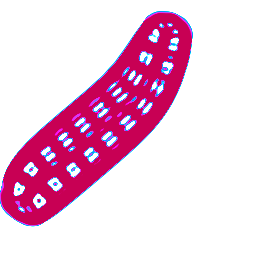}} &
 &
 &
 \fbox{\includegraphics[width=2.1cm]{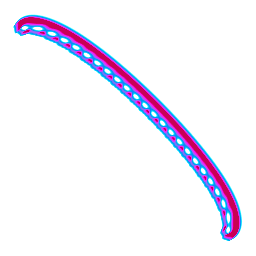}} &
 \fbox{\includegraphics[width=2.1cm]{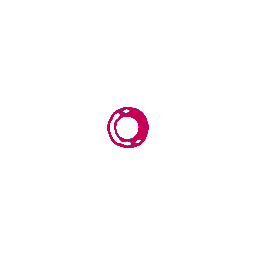}} \\
 \small{t=1} &
 \small{t=50} &
 &
 &
 &
 &
 &
 &
 &
 \\
 
 \fbox{\includegraphics[width=2.1cm]{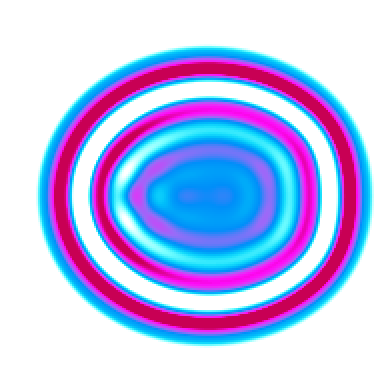}} &
 \fbox{\includegraphics[width=2.1cm]{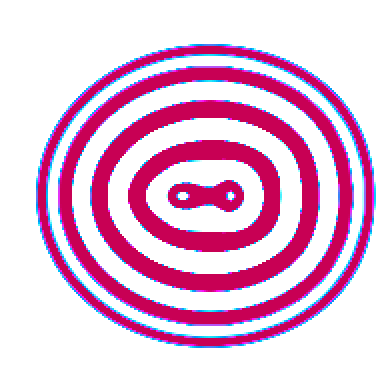}} &
 &
 &
 \fbox{\includegraphics[width=2.1cm]{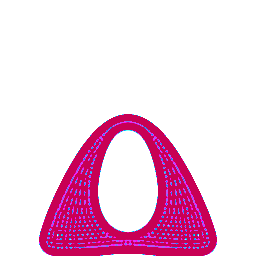}} &
 \fbox{\includegraphics[width=2.1cm]{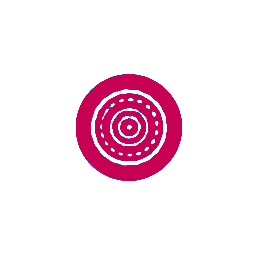}} &
 &
 &
 \fbox{\includegraphics[width=2.1cm]{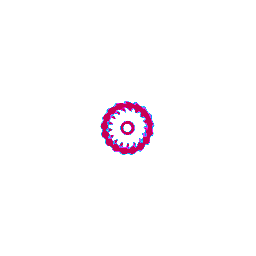}} &
 \fbox{\includegraphics[width=2.1cm]{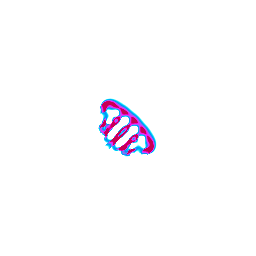}} \\
  \small{t=100} &
 \small{t=200} &
 &
 &
 &
 &
 &
 &
 &
 \\

\end{tabular}
 \caption{Example patterns produced in the continuous GOL system (Lenia). Illustration of the dynamical  morphing from an initial CPPN image to an animal (a). The automated discovery (b) is able to find similar complex animals as a human-expert manual search (c) by \cite{chan2019lenia}.} 
 \label{fig:examples_animals}
\end{figure}

\note{Definition}
Lenia \citep{chan2019lenia} is a continuous cellular automaton \citep{wolfram1983statistical} similar to Conway's Game of Life \citep{gardener1970mathematical}. 
Lenia, in particular, represents a high-dimensional complex dynamical system where diverse visual structures can self-organize and yet are hard to find by manual exploration. It features the richness of Turing-complete game-of-life models.
It is therefore well suited to test the performance of pattern exploration algorithms for unknown and complex systems. The fact that GOL models have been used widely to study self-organization in various disciplines, ranging from physics to biology and economics \citep{bak1989self}, also supports their generality and potential of reuse of our approach for discovery in other computational or wet high-dimensional systems. 

Lenia consists of a two-dimensional grid of cells $A \in [0,1]^{256 \times 256}$ where the state of each cell is a real-valued scalar activity $A^t(x) \in [0, 1]$.
The state of cells evolves over discrete time steps $t$ (Fig.~\ref{fig:examples_animals}, a).
The activity change is computed by integrating the activity of neighboring cells.
Lenia's behavior is controlled by its initial pattern $A^{t=1}$ and several settings that control the dynamics of the activity change ($R, T, \mu, \sigma, \beta_1, \beta_2 ,\beta_3$). 
  
\note{Morphogenesis of patterns and animals}
Lenia can be understood as a self-organizing morphogenetic system.
Its parameters for the initial pattern $A^{t=1}$ and dynamics control determine the development of morphological patterns. Lenia can produce diverse patterns with different dynamics (stable, non-stable or chaotic).
Most interesting, spatially localized coherent patterns that resemble in their shapes microscopic animals can emerge (Fig. \ref{fig:examples_animals}, b, c).
These pattern types, which we will denote \enquote{animals} as a short name, are a key reason scientists have used GOL models to study theories of the origins of life \citep{gardener1970mathematical, beer2004autopoiesis}.
Therefore, in our evaluation method based on measures of diversity (see below), we will in particular study the performance of IMGEPs, and the impact of using various approaches for goal space representation, on finding a diversity of \textit{animal} patterns.
We implemented for this purpose different pattern classifiers to analyze the exploration results.
Initially we differentiate between dead and alive patterns.
A pattern is dead if the activity of all cells are either $0$ or $1$.
Alive patterns are separated into animals and non-animals.
Animals are a connected areas of positive activity which are finite, i.e.\ which do not infinitely cross several borders.
All other patterns are non-animals whose activity usually spreads over the whole state space.

\subsection{Evaluation based on the diversity of Patterns}

The algorithms are evaluated based on their discovered diversity of patterns. 
Diversity is measured by the spread of the exploration in an \textit{analytic behavior space}.
This space is externally defined by the experimenter as in previous evaluation approaches in the IMGEP literature.
For example, in \cite{pere2018unsupervised}, the diversity of discovered effects of a robot that manipulates objects is measured by binning the space of object positions and counting the number of bins discovered. 
A difference here is that the experimenter does not have access to an easily interpretable hand-defined low-dimensional representation of possible patterns, equivalent to the cartesian coordinate of rigid objects. 
The space of raw observations $O$, i.e.\ the final Lenia patterns $A^{t=200}$, is also too high-dimensional for a meaningful measure of spread in it.
We constructed therefore an external evaluation space.
First, a latent representation space was build through the training of a $\beta$-VAE \citep{higgins2017beta} to learn the important features over a large dataset of 42500 Lenia patterns identified during the many experiments over all evaluated algorithms.  
This large dataset enabled to cover a diversity of patterns orders of magnitude larger than what could be found in any single algorithm experiment, which experimental budget was order of magnitude smaller. 
We then augmented that space by concatenating hand-defined features (the same features were used for a hand-defined goal space IMGEP).

For each experiment, all explored patterns were projected into the analytic behavior space. 
The diversity of the patterns is then measured by discretizing the space into bins of equal size by splitting each dimension into $7$ sections (results were found to be robust to the number of bins per dimension, see Appendix \ref{c:sm_definition_analytic_spaces}). This results in $7^{13}$ bins.
The number of bins in which at least one explored entity falls is used as a measure for diversity.

We also measured the diversity in the space of parameters $\Theta$ by constructing an \textit{analytic parameter space}.
The 15 features of this space consisted of Lenia's parameters ($R$, $T$, $\mu$, $\sigma$, $\beta_1$, $\beta_2$, $\beta_3$) and the latent representation of a $\beta$-VAE.
The $\beta$-VAE was trained on a large dataset of initial Lenia states ($A^{t=1}$) used over the experimental campaign. This diversity measures also used 7 bins per dimension.

\subsection{Algorithms}
The exploration behaviors of different IMGEP algorithms were evaluated and compared to a random exploration. 
The IMGEP variants differ in their way how the goal space is defined or learned.

\textbf{Random exploration:}
The IMGEP variants were compared to a random exploration that sampled randomly for each of the $N$ exploration iterations the parameters $\theta$ including the initial state $A^{t=1}$.

\textbf{IMGEP-HGS - Goal exploration with a hand-defined goal space:}
The first IMGEP uses a hand-defined goal space that is composed of 5 features used in \cite{chan2019lenia}.
Each feature measures a certain property of the final pattern $A^{t=200}$ that emerged in Lenia:
1) the sum over the activity of all cells, 
2) the number of activated cells, 
3) the density of the activity center, 
4) an asymmetry measure of the pattern and 
5) a distribution measure of the pattern.

\textbf{IMGEP-PGL - Goal exploration with a pretrained goal space:}
For this IMGEP variant the goal space was learned with a $\beta$-VAE approach on training data before the exploration process started.
The training set consisted of 558 Lenia patterns: half were animals that have been manually identified by \cite{chan2019lenia}; the other half randomly generated with CPPNs (see Section~\ref{c:experiments_procedure}).

\textbf{IMGEP-OGL - Goal exploration with online learning of the goal space:} Algorithm~\ref{algo:IMGEP-OGL}.

\textbf{IMGEP-RGS - Goal exploration with a random goal space:} An ablated IMGEP using a goal space based on the encoder of a VAE with random weights.

\subsection{Experimental Procedure and hyperparameters}
\label{c:experiments_procedure}

For each algorithm 10 independent repetitions of the exploration experiment were conducted.
Each experiment consisted of $N = 5000$ exploration iterations. 
This number was chosen to be compatible with the application of the algorithms in physical experimental setups similar to \cite{grizou2020curious}, planned in future work.
For IMGEP variants the first $N_{\textrm{init}} = 1000$ iterations used random parameter sampling to initialize their histories $\mathcal{H}$.
For the following $4000$ iterations each IMGEP approach sampled a goal $g$ via an uniform distribution over its goal space.
The ranges for sampling in the hand-defined goal space (HGS) are defined in Table~\ref{tbl:sm_hgs_goalspace_ranges} (Appendix \ref{c:sm_hgs}).
The ranges for the $\beta$-VAE based goal spaces (PGL, OGL) were set to $[-3, 3]$ for each of their latent variables.
Then, the parameter $\theta_k$ from a previous exploration in $\mathcal{H}$ was selected whose reached goal $\hat{g}_k$ had the minimum euclidean distance to the current goal $g$ within the goal space.
This parameter was then mutated to generate the parameter $\theta$ that was explored.

The parameters $\theta$ consisted of a CPPN (Section~\ref{c:methods_cppn}) that generates the initial state $A^{t=1}$ for Lenia and the settings defining Lenia's dynamics: $\theta = [\textrm{CPPN} \rightarrow A^{t=1}, R, T, \mu, \sigma, \beta_1, \beta_2 ,\beta_3]$.  
CPPNs were initialized and mutated by a random process that defines their structure and connection weights as done by \cite{stanley2006exploiting}.
The random initialization of the other Lenia settings for the dynamics was done by an uniform distribution and their mutation by a Gaussian distribution around the original values.
The meta parameters to initialize and mutate the parameters were the same for all algorithms.
They were manually chosen without optimizing them for a specific algorithm.

\section{Results}

We address several questions evaluating the ability of IMGEP algorithms to identify a diverse set of patterns, and in particular diverse \enquote{animal} patterns (i.e. spatially localized patterns). 

\begin{figure}[b!]
\centering
\setlength\tabcolsep{1pt}
\begin{tabular}{@{}cc@{}}
 (a) Diversity in Parameter Space 
 &
 (b) Diversity in Behavior Space 
 \\
 \includegraphics[width=0.49\linewidth]{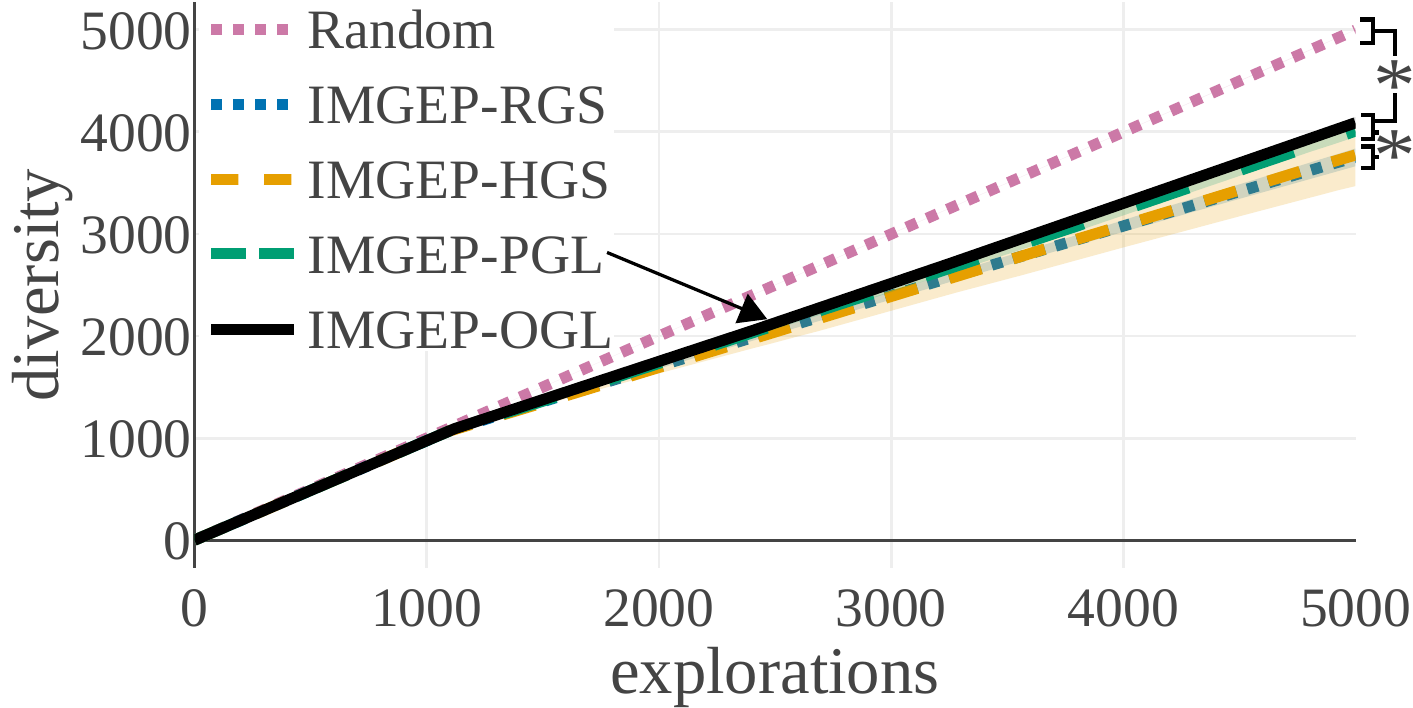} 
 &
 \includegraphics[width=0.49\linewidth]{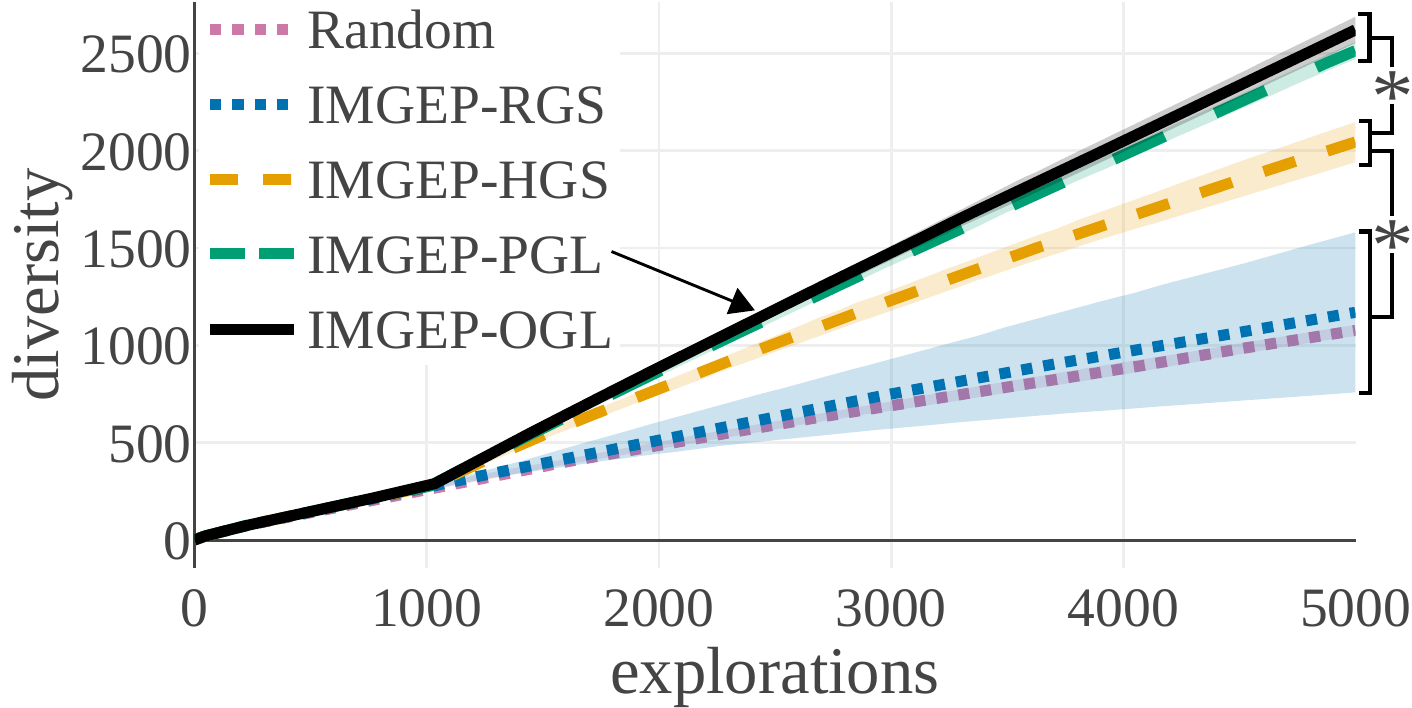} \\
 
 (c) Behavior Space Diversity for Animals \vspace{0.1cm}
&
(d) Behavior Space Diversity for Non-Animals
\\

 \includegraphics[width=0.49\linewidth]{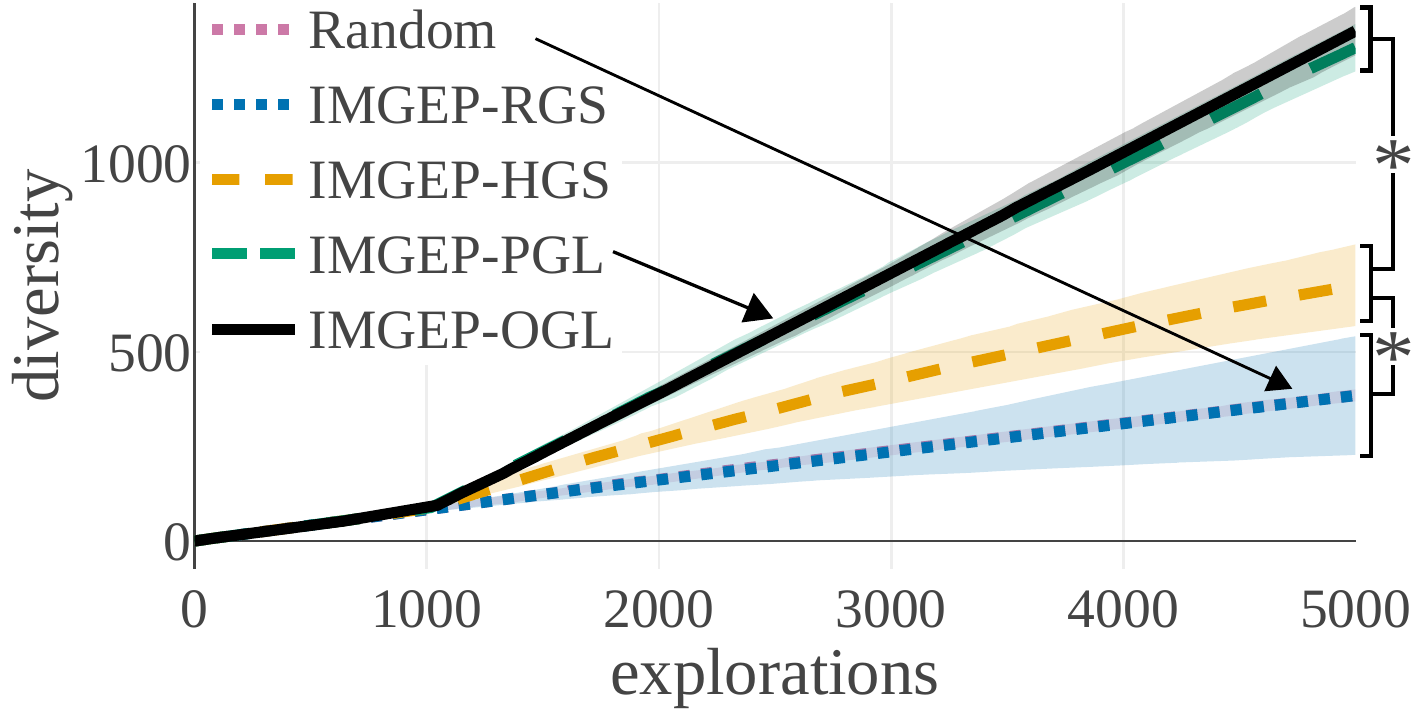}
 &
 \includegraphics[width=0.49\linewidth]{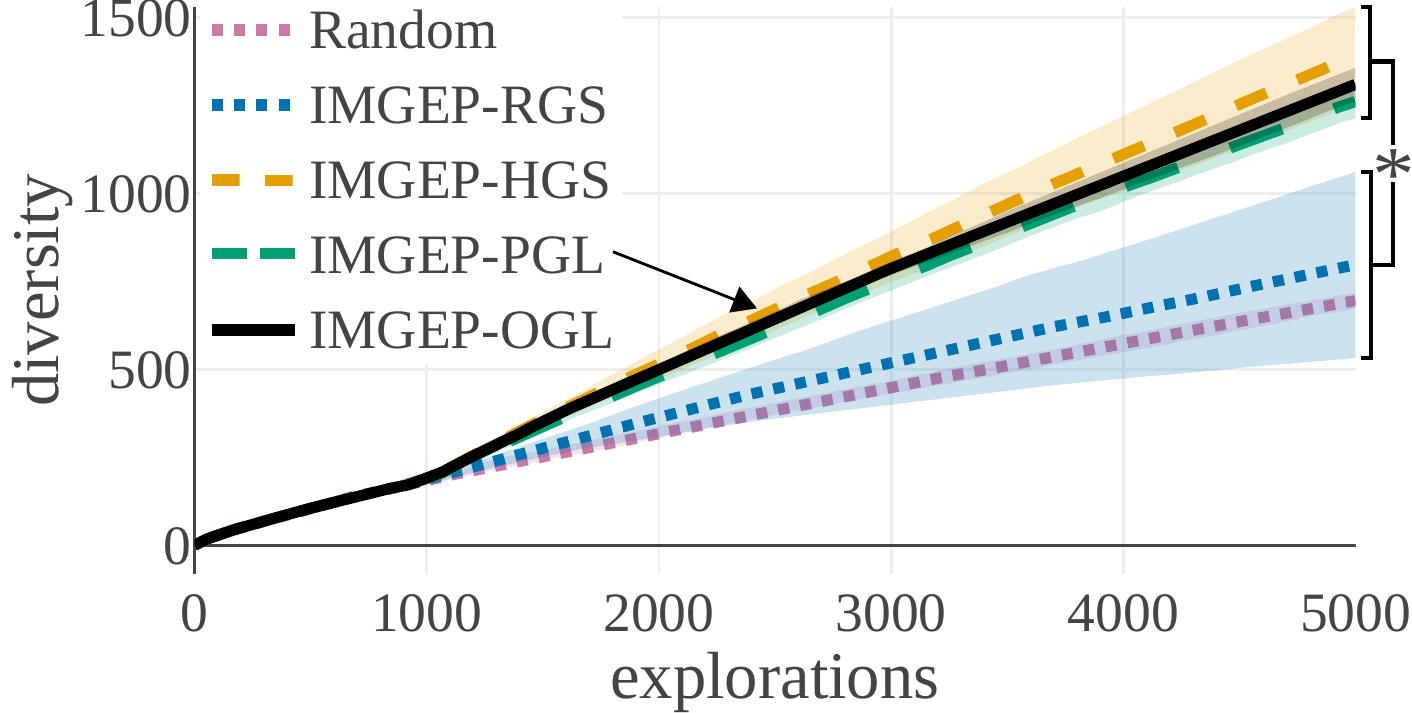}

\end{tabular}
 \caption{
 (a) Although random explorations reach the highest diversity in the analytic parameter space, (b) IMGEPs reach a higher diversity in the analytic behavior space (except when using random representations). 
 (c) IMGEPs with a learned goal space discovered a larger diversity of animals compared to a hand-defined goal space. 
 (d) Learned goal spaces are as efficient as a hand-defined space for finding diverse non-animals. 
 Overall, IMGEPs with unsupervised learning of goal features are efficient to discover a \textit{diversity of diverse
patterns}.
Depicted is the average diversity ($n=10$) with the standard deviation as shaded area (for some not visible because it is too small).
The final diversity is significantly different (Welch’s t-test, $p < 0.01$) for algorithms between the braces plotted on the right of each figure.
See Figs.~\ref{fig:identified_patterns_random} to \ref{fig:identified_patterns_ogl} for a qualitative visual illustration of these results.
}
 \label{fig:results_diversity}
\end{figure}


\paragraph{Does goal exploration outperform random parameter exploration?} 
In robotics/agents contexts where scenes are populated with rigid objects, various forms of goal exploration algorithms outperform random parameter exploration \citep{laversanne2018curiosity}. 
We checked whether this still holds in continuous GOL which have very different properties.
Measures of the diversity in the analytic behavior space confirmed the advantage of IMGEPs with hand-designed (HGS) or learned goal spaces (PGL/OGL) over random explorations (Fig.~\ref{fig:results_diversity}, b). 
The organization resulting from goal exploration is also visible through the diversity in the space of parameters.
IMGEPs focus their exploration on subspaces that are useful for targeting new goals.
In contrast, random parameter exploration is unguided, resulting in a higher diversity in the parameter space (Fig.~\ref{fig:results_diversity}, b). 

\paragraph{What is the impact of learning a goal space vs. using random or hand-defined features?}
We compared also the performance of random VAE goal spaces (RGS) to learned goal spaces (PGL/OGL). 
For reinforcement learning problems, using intrinsic reward functions based on random features of the observations 
can result in a high or boosted performance \citep{burda2018large,burda2018exploration}.
In our context however, using random features (RGS) collapsed the performance of goal exploration, and did not even
outperform random parameter exploration for all kinds of behavioral diversity (Fig.~\ref{fig:results_diversity}).
Results also show that hand-defined features (HGS) produced significantly less global diversity and less \enquote{animal} diversity than using learned features (PGL/OGL). 
However, HGS found an equal diversity of \enquote{non-animals}.
These results show that in this domain, the goal-space has a critical influence on the type and diversity of patterns 
discovered. 
Furthermore, unsupervised learning is an efficient approach to discover a \textit{diversity of diverse patterns}, i.e.\ both efficient at finding diverse animals and diverse non-animals. 

\paragraph{Is pretraining on a database of expert patterns necessary for efficient discovery of diverse animals?}
A possibility to bias exploration towards patterns of interest, such as \enquote{animals}, is to pretrain a goal space with
a pattern dataset hand-made by an expert.
Here PGL is pretrained with a dataset containing 50\% animals. 
This leads PGL to discover a high diversity of animals. 
However, the new online approach (OGL) is as efficient as PGL to discover diverse patterns (Fig.~\ref{fig:results_diversity}, b,c,d).
Taken together, these results uncover an interesting bias of using learned features with a VAE architecture, which strongly incentivizes efficient discovery of diverse spatially localized patterns.

\begin{figure}[t]
\centering
\begin{tabular}{@{} c c @{}}
(a) IMGEP-HGS Goal Space  & (b) IMGEP-OGL Goal Space \\

\includegraphics[width=0.49\linewidth]{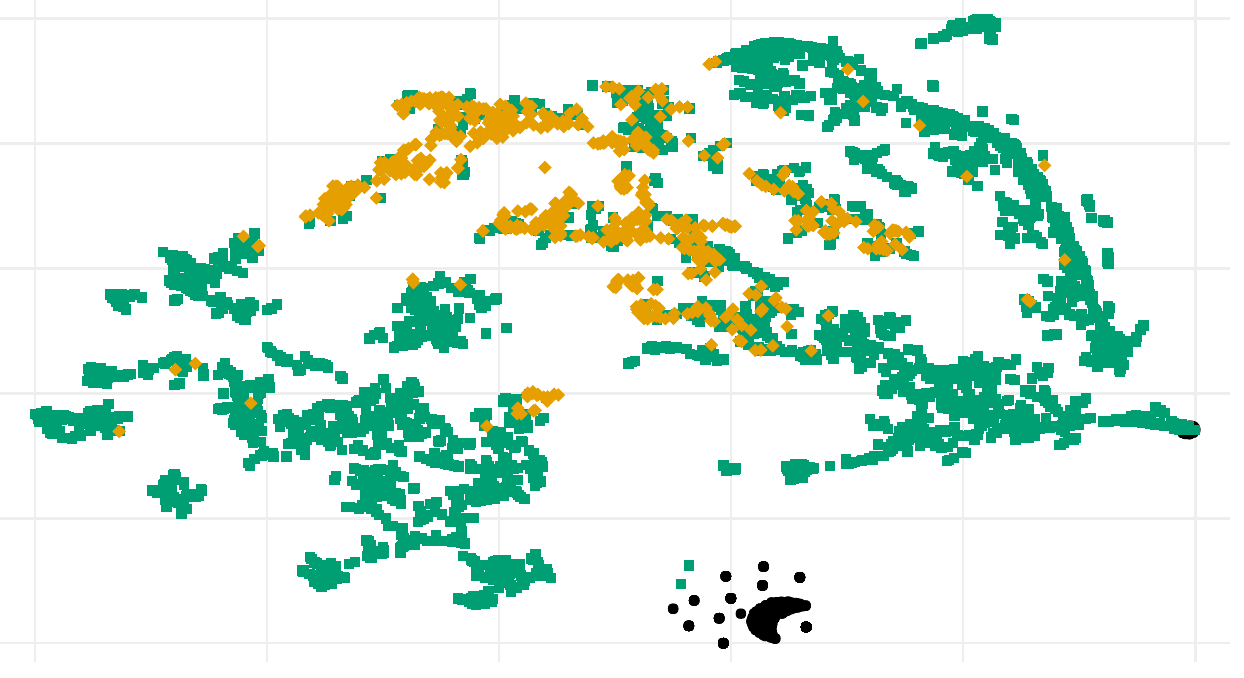} & \includegraphics[width=0.49\linewidth]{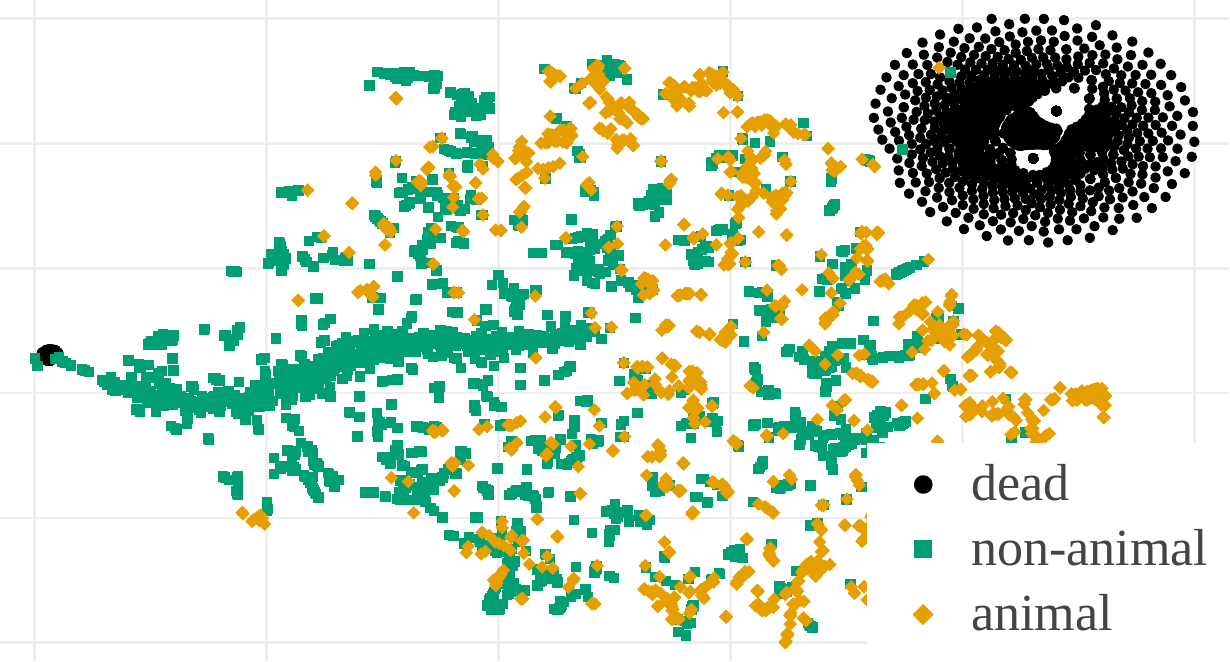} \\
 
\end{tabular}
 \caption{
 (a) Hand-defined and (b) learned goal spaces have major differences shown here by a t-SNE visualization. 
 The different number and size of clusters of animals or non-animals can explain the differences in their resulting diversity between the algorithms (Fig.~\ref{fig:results_diversity}).}
 \label{fig:results_goal_space}
\end{figure}

\paragraph{How do goal space representations differ?}
We analyzed the goal spaces of the different IMGEP variants to understand their behavior by visualizing their reached goals in a two-dimensional space.  
T-SNE \citep{maaten2008visualizing} was used to reduce the high-dimensional goal spaces.

The hand-defined (HGS) and learned (OGL) goal spaces show strong differences between each other (Fig.~\ref{fig:results_goal_space}).
We believe this explains their different abilities to find either a high diversity of non-animals or animals (Fig.~\ref{fig:results_diversity}, c, d).
The goal space of the HGS shows large areas and several clusters for non-animal patterns (Fig.~\ref{fig:results_goal_space}, a).
Animals form only few and nearby clusters.
Thus, the hand-defined features seem poor to discriminate and describe animal patterns in Lenia.
As a consequence, when goals are uniformly sampled within this goal space during the exploration process, then more goals are generated in regions that describe non-animals.
This can explain why HGS explored a higher diversity of non-animal patterns but only a low diversity of animal patterns.
In contrast, features learned by OGL capture better factors that differentiate animal patterns.
This is indicated by the several clusters of animals that span a wide area in its goal space (Fig.~\ref{fig:results_goal_space}, b).

We attribute this effect to the difficulty of VAEs to capture sharp details \citep{zhao2017towards}. 
They therefore represent mainly the general form of the patterns but not their fine-grained structures.
Animals differ often in their form whereas non-animals occupy often the whole cell grid and differ in their fine-grained details.
The goal spaces learned by VAEs seem therefore better suited for exploring diverse sets of animal patterns.

\section{Conclusion}

We formulated a novel application area for machine learning: the problem of automatically discovering self-organized patterns in complex dynamical systems with high-dimensions both in the parameter space and in the observation space.
We showed that this problem calls for advanced methods requiring the dynamic interaction between sample efficient autonomous exploration and unsupervised representation learning. 
We demonstrated that population-based IMGEPs are a promising algorithmic framework to address this challenge, by showing how it can discover diverse self-organized  patterns in a continuous GOL. 
In particular, we introduced a new approach for learning a goal space representation online via data collected during the exploration process.
It enables sample efficient discovery of diverse sets of animal-like patterns, similar to those identified by human experts and yet without relying on such prior expert knowledge (Fig.~\ref{fig:examples_animals}). 
We also analyzed the impact of goal space representations on the diversity and types of discovered patterns.  

The continuous game of life shares many properties with other artificial or natural complex systems, explaining why GOL models have been used in many disciplines to study self-organization, see \cite{bak1989self}.
We therefore believe this study shows the potential of IMGEPs for automated discovery in other systems encountered in physics, chemistry or even computer animation. 
In further work, we aim to apply this approach in robotized wet experiments such as the one presented in \cite{grizou2020curious} addressing the fundamental understanding of how proto-cells can self-organize.
We are also working together with bio-chemists to map the space of behaviors of certain complex biochemistry systems for which no appropriate model exists.
Once such a map has been discovered, it can be used to optimize the system for target properties by leveraging the diversity of patterns found through the unsupervised discovery process. 

\subsubsection*{Acknowledgments}

We thank Bert Wang-Chak Chan for his helpful discussions about the Lenia system. 
We also thank Jonathan Grizou and Cedric Colas for useful comments on the paper.

\bibliography{iclr2020_conference.bib}

\begin{thebibliography}{46}
\providecommand{\natexlab}[1]{#1}
\providecommand{\url}[1]{\texttt{#1}}
\expandafter\ifx\csname urlstyle\endcsname\relax
  \providecommand{\doi}[1]{doi: #1}\else
  \providecommand{\doi}{doi: \begingroup \urlstyle{rm}\Url}\fi

\bibitem[Bak et~al.(1989)Bak, Chen, and Creutz]{bak1989self}
Per Bak, Kan Chen, and Michael Creutz.
\newblock Self-organized criticality in the'game of life.
\newblock \emph{Nature}, 342\penalty0 (6251):\penalty0 780, 1989.

\bibitem[Baldassarre \& Mirolli(2013)Baldassarre and
  Mirolli]{baldassarre2013intrinsically}
Gianluca Baldassarre and Marco Mirolli.
\newblock \emph{Intrinsically motivated learning in natural and artificial
  systems}.
\newblock Springer, 2013.

\bibitem[Ball(1999)]{ball1999self}
Philip Ball.
\newblock \emph{The self-made tapestry: pattern formation in nature}, volume
  198.
\newblock Oxford University Press Oxford, 1999.

\bibitem[Baranes \& Oudeyer(2013)Baranes and Oudeyer]{baranes2013active}
Adrien Baranes and Pierre-Yves Oudeyer.
\newblock Active learning of inverse models with intrinsically motivated goal
  exploration in robots.
\newblock \emph{Robotics and Autonomous Systems}, 61\penalty0 (1):\penalty0
  49--73, 2013.

\bibitem[Beer(2004)]{beer2004autopoiesis}
Randall~D Beer.
\newblock Autopoiesis and cognition in the game of life.
\newblock \emph{Artificial Life}, 10\penalty0 (3):\penalty0 309--326, 2004.

\bibitem[Bellemare et~al.(2016)Bellemare, Srinivasan, Ostrovski, Schaul,
  Saxton, and Munos]{bellemare2016unifying}
Marc Bellemare, Sriram Srinivasan, Georg Ostrovski, Tom Schaul, David Saxton,
  and Remi Munos.
\newblock Unifying count-based exploration and intrinsic motivation.
\newblock In \emph{Advances in Neural Information Processing Systems}, pp.\
  1471--1479, 2016.

\bibitem[Bengio et~al.(2013)Bengio, Courville, and
  Vincent]{bengio2013representation}
Yoshua Bengio, Aaron Courville, and Pascal Vincent.
\newblock Representation learning: A review and new perspectives.
\newblock \emph{IEEE transactions on pattern analysis and machine
  intelligence}, 35\penalty0 (8):\penalty0 1798--1828, 2013.

\bibitem[Burda et~al.(2019{\natexlab{a}})Burda, Edwards, Pathak, Storkey,
  Darrell, and Efros]{burda2018large}
Yuri Burda, Harri Edwards, Deepak Pathak, Amos Storkey, Trevor Darrell, and
  Alexei~A. Efros.
\newblock Large-scale study of curiosity-driven learning.
\newblock 2019{\natexlab{a}}.

\bibitem[Burda et~al.(2019{\natexlab{b}})Burda, Edwards, Storkey, and
  Klimov]{burda2018exploration}
Yuri Burda, Harrison Edwards, Amos Storkey, and Oleg Klimov.
\newblock Exploration by random network distillation.
\newblock In \emph{International Conference on Learning Representations},
  2019{\natexlab{b}}.

\bibitem[Camazine et~al.(2003)Camazine, Deneubourg, Franks, Sneyd, Bonabeau,
  and Theraula]{camazine2003self}
Scott Camazine, Jean-Louis Deneubourg, Nigel~R Franks, James Sneyd, Eric
  Bonabeau, and Guy Theraula.
\newblock \emph{Self-organization in biological systems}.
\newblock Princeton university press, 2003.

\bibitem[Chan(2019)]{chan2019lenia}
Bert Wang-Chak Chan.
\newblock Lenia: Biology of artificial life.
\newblock \emph{Complex Systems}, 28\penalty0 (3):\penalty0 251–--286, 2019.

\bibitem[Colas et~al.(2018)Colas, Sigaud, and Oudeyer]{colas2018gep}
C{\'e}dric Colas, Olivier Sigaud, and Pierre-Yves Oudeyer.
\newblock Gep-pg: Decoupling exploration and exploitation in deep reinforcement
  learning algorithms.
\newblock In \emph{International Conference on Machine Learning (ICML)}, 2018.

\bibitem[Conti et~al.(2018)Conti, Madhavan, Such, Lehman, Stanley, and
  Clune]{conti2018improving}
Edoardo Conti, Vashisht Madhavan, Felipe~Petroski Such, Joel Lehman, Kenneth
  Stanley, and Jeff Clune.
\newblock Improving exploration in evolution strategies for deep reinforcement
  learning via a population of novelty-seeking agents.
\newblock In \emph{Advances in neural information processing systems}, pp.\
  5027--5038, 2018.

\bibitem[Duros et~al.(2017)Duros, Grizou, Xuan, Hosni, Long, Miras, and
  Cronin]{duros2017human}
Vasilios Duros, Jonathan Grizou, Weimin Xuan, Zied Hosni, De-Liang Long,
  Haralampos~N Miras, and Leroy Cronin.
\newblock Human versus robots in the discovery and crystallization of gigantic
  polyoxometalates.
\newblock \emph{Angewandte Chemie}, 129\penalty0 (36):\penalty0 10955--10960,
  2017.

\bibitem[Forestier \& Oudeyer(2016)Forestier and Oudeyer]{forestier2016modular}
S{\'e}bastien Forestier and Pierre-Yves Oudeyer.
\newblock Modular active curiosity-driven discovery of tool use.
\newblock In \emph{2016 IEEE/RSJ International Conference on Intelligent Robots
  and Systems (IROS)}, pp.\  3965--3972. IEEE, 2016.

\bibitem[Forestier et~al.(2017)Forestier, Mollard, and
  Oudeyer]{forestier2017intrinsically}
S{\'e}bastien Forestier, Yoan Mollard, and Pierre-Yves Oudeyer.
\newblock Intrinsically motivated goal exploration processes with automatic
  curriculum learning.
\newblock \emph{preprint arXiv:1708.02190}, 2017.

\bibitem[Gardener(1970)]{gardener1970mathematical}
Martin Gardener.
\newblock Mathematical games: The fantastic combinations of john conway's new
  solitaire game" life,".
\newblock \emph{Scientific American}, 223:\penalty0 120--123, 1970.

\bibitem[Gardner et~al.(1983)Gardner, Gardner, Gardner, and
  Gardner]{gardner1983wheels}
Martin Gardner, Martin Gardner, Martin Gardner, and Martin Gardner.
\newblock \emph{Wheels, life, and other mathematical amusements}, volume~86.
\newblock WH Freeman New York, 1983.

\bibitem[Glorot \& Bengio(2010)Glorot and Bengio]{glorot2010understanding}
Xavier Glorot and Yoshua Bengio.
\newblock Understanding the difficulty of training deep feedforward neural
  networks.
\newblock In \emph{Proceedings of the thirteenth international conference on
  artificial intelligence and statistics}, pp.\  249--256, 2010.

\bibitem[Grizou et~al.(2020)Grizou, Points, Sharma, and
  Cronin]{grizou2020curious}
Jonathan Grizou, Laurie~J Points, Abhishek Sharma, and Leroy Cronin.
\newblock A curious formulation robot enables the discovery of a novel
  protocell behavior.
\newblock \emph{Science Advances}, 6\penalty0 (5):\penalty0 eaay4237, 2020.

\bibitem[Higgins et~al.(2017)Higgins, Matthey, Pal, Burgess, Glorot, Botvinick,
  Mohamed, and Lerchner]{higgins2017beta}
Irina Higgins, Loic Matthey, Arka Pal, Christopher Burgess, Xavier Glorot,
  Matthew Botvinick, Shakir Mohamed, and Alexander Lerchner.
\newblock beta-vae: Learning basic visual concepts with a constrained
  variational framework.
\newblock In \emph{International Conference on Learning Representations},
  volume~3, 2017.

\bibitem[Jolliffe(1986)]{Jolliffe1986}
I.~T. Jolliffe.
\newblock \emph{Principal Component Analysis and Factor Analysis}, pp.\
  115--128.
\newblock Springer New York, New York, NY, 1986.
\newblock ISBN 978-1-4757-1904-8.
\newblock \doi{10.1007/978-1-4757-1904-8_7}.

\bibitem[Kauffman(1993)]{kauffman1993origins}
Stuart~A Kauffman.
\newblock \emph{The origins of order: Self-organization and selection in
  evolution}.
\newblock OUP, 1993.

\bibitem[King et~al.(2004)King, Whelan, Jones, Reiser, Bryant, Muggleton, Kell,
  and Oliver]{king2004functional}
Ross~D King, Kenneth~E Whelan, Ffion~M Jones, Philip~GK Reiser, Christopher~H
  Bryant, Stephen~H Muggleton, Douglas~B Kell, and Stephen~G Oliver.
\newblock Functional genomic hypothesis generation and experimentation by a
  robot scientist.
\newblock \emph{Nature}, 427\penalty0 (6971):\penalty0 247, 2004.

\bibitem[King et~al.(2009)King, Rowland, Oliver, Young, Aubrey, Byrne, Liakata,
  Markham, Pir, Soldatova, et~al.]{king2009automation}
Ross~D King, Jem Rowland, Stephen~G Oliver, Michael Young, Wayne Aubrey, Emma
  Byrne, Maria Liakata, Magdalena Markham, Pinar Pir, Larisa~N Soldatova,
  et~al.
\newblock The automation of science.
\newblock \emph{Science}, 324\penalty0 (5923):\penalty0 85--89, 2009.

\bibitem[Kingma \& Ba(2014)Kingma and Ba]{kingma2014adam}
Diederik~P Kingma and Jimmy Ba.
\newblock Adam: A method for stochastic optimization.
\newblock \emph{preprint arXiv:1412.6980}, 2014.

\bibitem[Kingma \& Welling(2013)Kingma and Welling]{kingma2013auto}
Diederik~P Kingma and Max Welling.
\newblock Auto-encoding variational bayes.
\newblock \emph{arXiv:1312.6114}, 2013.

\bibitem[Laversanne-Finot et~al.(2018)Laversanne-Finot, Pere, and
  Oudeyer]{laversanne2018curiosity}
Adrien Laversanne-Finot, Alexandre Pere, and Pierre-Yves Oudeyer.
\newblock Curiosity driven exploration of learned disentangled goal spaces.
\newblock In \emph{Proceedings of The 2nd Conference on Robot Learning},
  volume~87 of \emph{PLMR}, 2018.

\bibitem[Lehman \& Stanley(2008)Lehman and Stanley]{lehman2008exploiting}
Joel Lehman and Kenneth~O Stanley.
\newblock Exploiting open-endedness to solve problems through the search for
  novelty.
\newblock In \emph{ALIFE}, pp.\  329--336, 2008.

\bibitem[Maaten \& Hinton(2008)Maaten and Hinton]{maaten2008visualizing}
Laurens van~der Maaten and Geoffrey Hinton.
\newblock Visualizing data using t-sne.
\newblock \emph{Journal of machine learning research}, 9\penalty0
  (Nov):\penalty0 2579--2605, 2008.

\bibitem[Mitchell et~al.(1996)Mitchell, Crutchfield, Das,
  et~al.]{mitchell1996evolving}
Melanie Mitchell, James~P Crutchfield, Rajarshi Das, et~al.
\newblock Evolving cellular automata with genetic algorithms: A review of
  recent work.
\newblock In \emph{Proceedings of the First International Conference on
  Evolutionary Computation and Its Applications (EvCA’96)}, volume~8. Moscow,
  1996.

\bibitem[Nair et~al.(2018)Nair, Pong, Dalal, Bahl, Lin, and
  Levine]{nair2018visual}
Ashvin~V Nair, Vitchyr Pong, Murtaza Dalal, Shikhar Bahl, Steven Lin, and
  Sergey Levine.
\newblock Visual reinforcement learning with imagined goals.
\newblock In \emph{Advances in Neural Information Processing Systems}, pp.\
  9191--9200, 2018.

\bibitem[Pastor et~al.(2013)Pastor, Kalakrishnan, Meier, Stulp, Buchli,
  Theodorou, and Schaal]{pastor2013dynamic}
Peter Pastor, Mrinal Kalakrishnan, Franziska Meier, Freek Stulp, Jonas Buchli,
  Evangelos Theodorou, and Stefan Schaal.
\newblock From dynamic movement primitives to associative skill memories.
\newblock \emph{Robotics and Autonomous Systems}, 61\penalty0 (4):\penalty0
  351--361, 2013.

\bibitem[P{\'e}r{\'e} et~al.(2018)P{\'e}r{\'e}, Forestier, Sigaud, and
  Oudeyer]{pere2018unsupervised}
Alexandre P{\'e}r{\'e}, S{\'e}bastien Forestier, Olivier Sigaud, and
  Pierre-Yves Oudeyer.
\newblock {Unsupervised Learning of Goal Spaces for Intrinsically Motivated
  Goal Exploration}.
\newblock In \emph{{ICLR2018 - 6th International Conference on Learning
  Representations}}, Vancouver, Canada, April 2018.

\bibitem[Pong et~al.(2019)Pong, Dalal, Lin, Nair, Bahl, and
  Levine]{pong2019skew}
Vitchyr~H Pong, Murtaza Dalal, Steven Lin, Ashvin Nair, Shikhar Bahl, and
  Sergey Levine.
\newblock Skew-fit: State-covering self-supervised reinforcement learning.
\newblock \emph{preprint arXiv:1903.03698}, 2019.

\bibitem[Pugh et~al.(2016)Pugh, Soros, and Stanley]{pugh2016quality}
Justin~K Pugh, Lisa~B Soros, and Kenneth~O Stanley.
\newblock Quality diversity: A new frontier for evolutionary computation.
\newblock \emph{Frontiers in Robotics and AI}, 3:\penalty0 40, 2016.

\bibitem[Raccuglia et~al.(2016)Raccuglia, Elbert, Adler, Falk, Wenny, Mollo,
  Zeller, Friedler, Schrier, and Norquist]{raccuglia2016machine}
Paul Raccuglia, Katherine~C Elbert, Philip~DF Adler, Casey Falk, Malia~B Wenny,
  Aurelio Mollo, Matthias Zeller, Sorelle~A Friedler, Joshua Schrier, and
  Alexander~J Norquist.
\newblock Machine-learning-assisted materials discovery using failed
  experiments.
\newblock \emph{Nature}, 533\penalty0 (7601):\penalty0 73, 2016.

\bibitem[Reizman et~al.(2016)Reizman, Wang, Buchwald, and
  Jensen]{reizman2016suzuki}
Brandon~J Reizman, Yi-Ming Wang, Stephen~L Buchwald, and Klavs~F Jensen.
\newblock Suzuki--miyaura cross-coupling optimization enabled by automated
  feedback.
\newblock \emph{Reaction chemistry \& engineering}, 1\penalty0 (6):\penalty0
  658--666, 2016.

\bibitem[Richards et~al.(2011)Richards, Starr, Brink, Miller, Bloom, Butler,
  James, Long, and Rice]{richards2011active}
Joseph~W Richards, Dan~L Starr, Henrik Brink, Adam~A Miller, Joshua~S Bloom,
  Nathaniel~R Butler, J~Berian James, James~P Long, and John Rice.
\newblock Active learning to overcome sample selection bias: application to
  photometric variable star classification.
\newblock \emph{The Astrophysical Journal}, 744\penalty0 (2):\penalty0 192,
  2011.

\bibitem[Rolf et~al.(2010)Rolf, Steil, and Gienger]{rolf2010goal}
Matthias Rolf, Jochen~J Steil, and Michael Gienger.
\newblock Goal babbling permits direct learning of inverse kinematics.
\newblock \emph{IEEE Transactions on Autonomous Mental Development}, 2\penalty0
  (3):\penalty0 216--229, 2010.

\bibitem[Sapin et~al.(2003)Sapin, Bailleux, and
  Jean-Jacques]{sapin2003research}
Emmanuel Sapin, Olivier Bailleux, and Chabrier Jean-Jacques.
\newblock Research of a cellular automaton simulating logic gates by
  evolutionary algorithms.
\newblock In \emph{European Conference on Genetic Programming}, pp.\  414--423.
  Springer, 2003.

\bibitem[Stanley(2006)]{stanley2006exploiting}
Kenneth~O Stanley.
\newblock Exploiting regularity without development.
\newblock In \emph{Proceedings of the AAAI Fall Symposium on Developmental
  Systems}, pp.\ ~37. AAAI Press Menlo Park, CA, 2006.

\bibitem[Stanley \& Miikkulainen(2002)Stanley and
  Miikkulainen]{stanley2002efficient}
Kenneth~O Stanley and Risto Miikkulainen.
\newblock Efficient evolution of neural network topologies.
\newblock In \emph{Proceedings of the 2002 Congress on Evolutionary
  Computation.}, volume~2. IEEE, 2002.

\bibitem[Tschannen et~al.(2018)Tschannen, Bachem, and
  Lucic]{tschannen2018recent}
Michael Tschannen, Olivier Bachem, and Mario Lucic.
\newblock Recent advances in autoencoder-based representation learning.
\newblock \emph{preprint arXiv:1812.05069}, 2018.

\bibitem[Wolfram(1983)]{wolfram1983statistical}
Stephen Wolfram.
\newblock Statistical mechanics of cellular automata.
\newblock \emph{Rev. of mod. physics}, 55\penalty0 (3):\penalty0 601, 1983.

\bibitem[Zhao et~al.(2017)Zhao, Song, and Ermon]{zhao2017towards}
Shengjia Zhao, Jiaming Song, and Stefano Ermon.
\newblock Towards deeper understanding of variational autoencoding models.
\newblock \emph{preprint arXiv:1702.08658}, 2017.

\end{thebibliography}
\bibliographystyle{iclr2020_conference}


\appendix

\section{Additional Results and Figures}
\label{c:sm_additional_results}

We provide visualizations of discovered patterns, as well as of learned goal-spaces. 
A dataset and an interactive visualization of all discovered patterns can be found at \url{https://automated-discovery.github.io/}. 
The visualization serves as a support in qualitatively evaluating the diversity of discovered patterns. 
Integrated into an interactive interface, these graphs are also useful for a human to easily explore and visualize the different types of found patterns.
In addition, we provide statistical results quantifying the proportion of sub types of discoverd patterns.

\subsection{Discovered Patterns}

\begin{figure}[p!]
  \centering
  \textbf{IMGEP-RGS}
  \includegraphics[width=\textwidth]{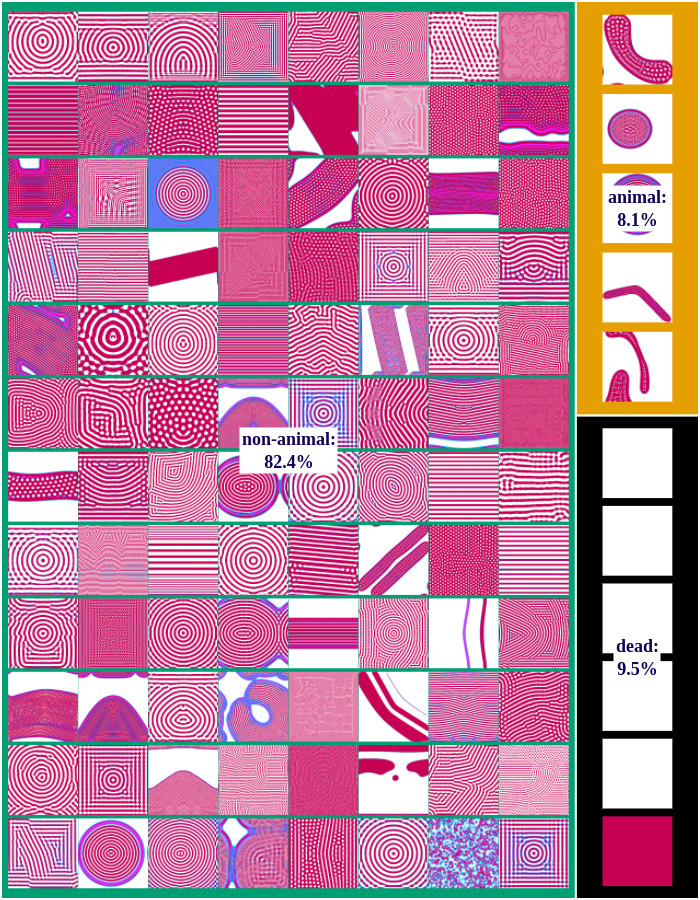}
  \caption{Randomly selected examples of patterns discovered by the IMGEP-RGS algorithm during a single exploration with 5000 iterations.}
  \label{fig:identified_patterns_rgs}
\end{figure}

Fig.~\ref{fig:identified_patterns_random} to \ref{fig:identified_patterns_ogl} illustrate examples of discovered patterns per class (animal, non-animal, dead) for each algorithm.
The patterns have been randomly sampled from the results of a single exploration repetition experiment. 
Please note that the shown examples represent only a small fraction of all discovered patterns.
A database with all patterns can be found on the website for the paper.

\subsection{Visualization of Goal Spaces}

\begin{figure}[p!]
\centering
\setlength\tabcolsep{12pt}
\resizebox{\textwidth}{!}{%
\begin{tabular}{@{} l c c @{}}
& \huge{PCA}  & \huge{T-SNE}\\
& ~  & ~\\

  \huge{\textsc{rgs}} 
  & \makecell{\includegraphics{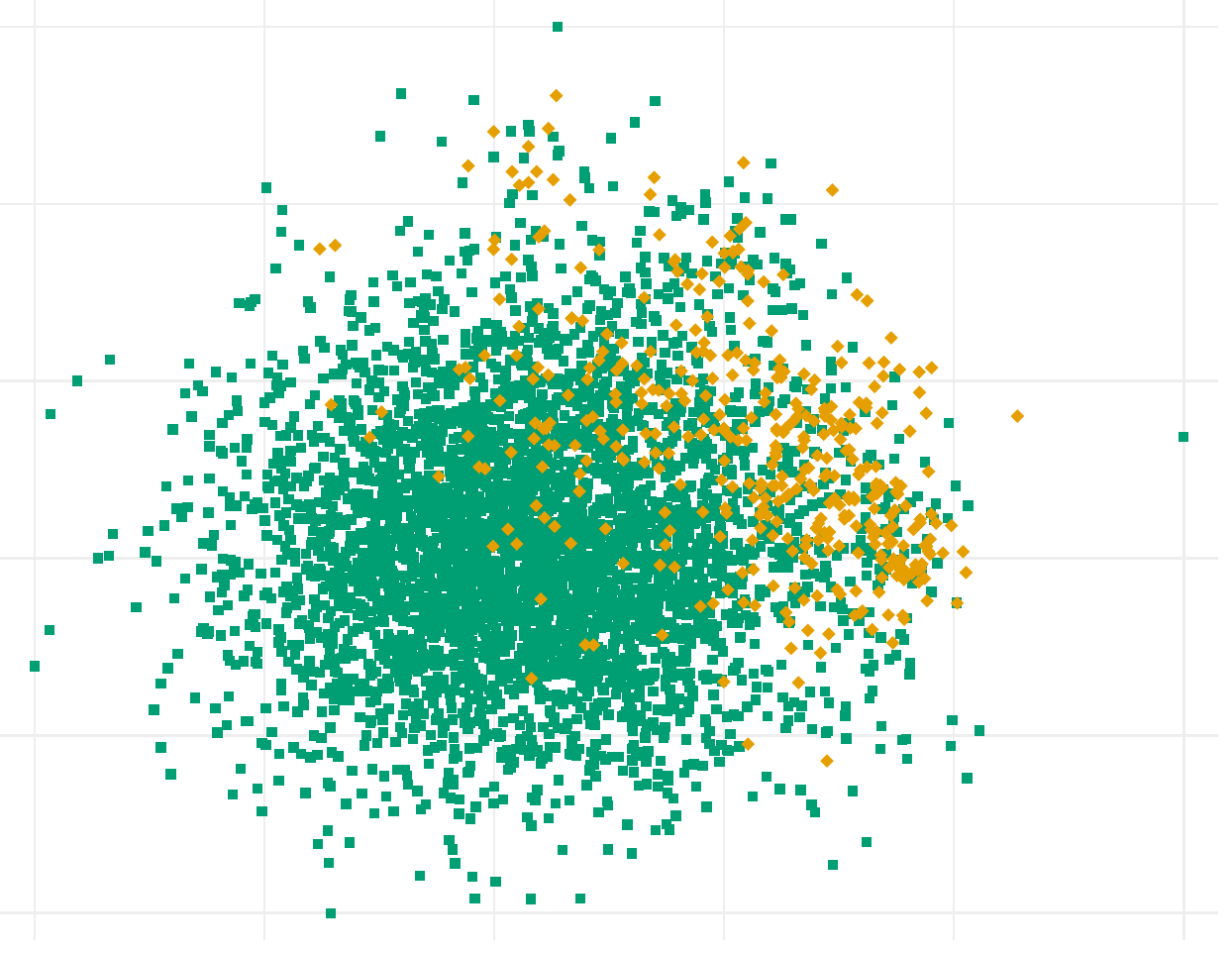}} 
  & \makecell{\includegraphics{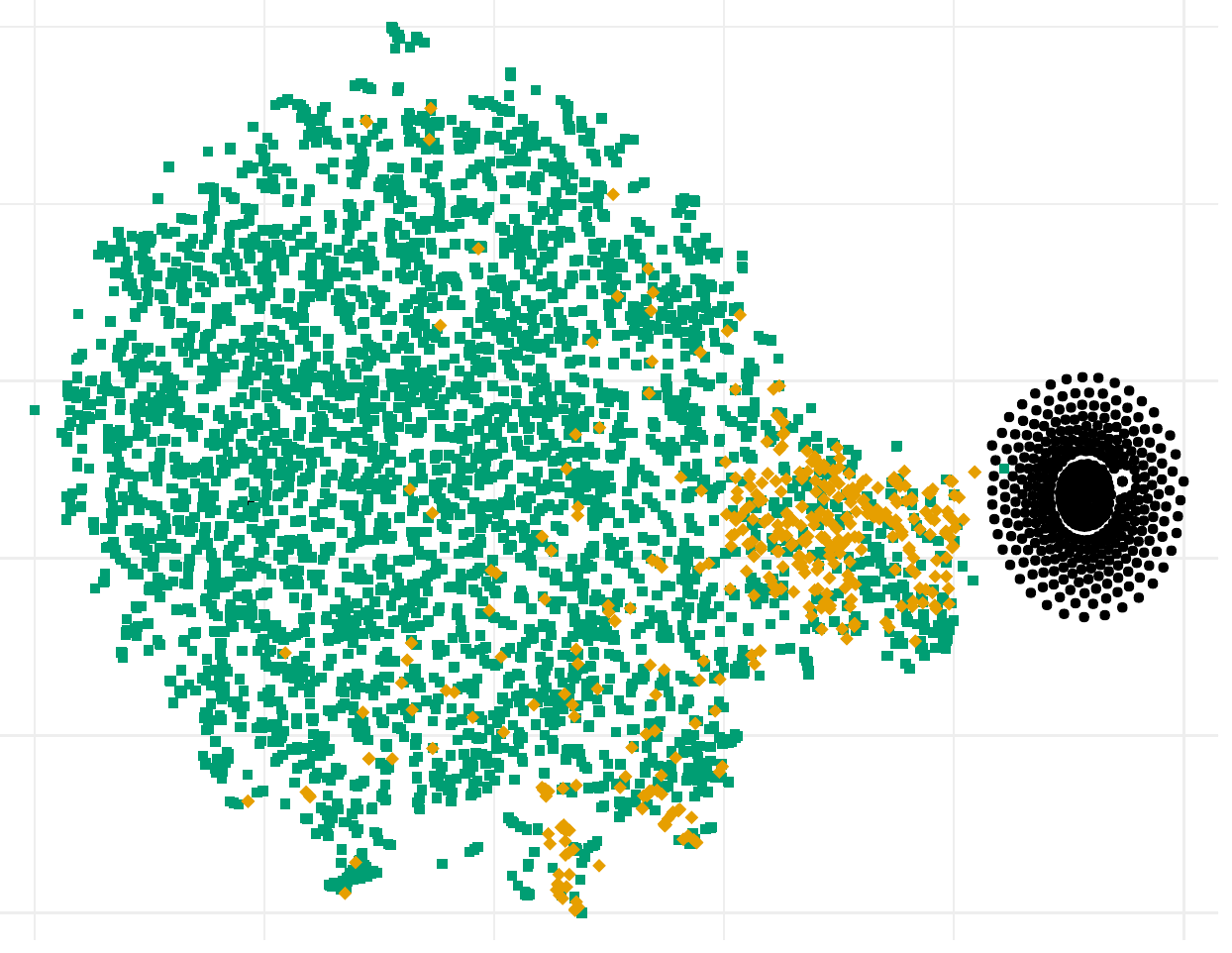}}\\
 
  \huge{\textsc{hgs}} 
  & \makecell{\includegraphics{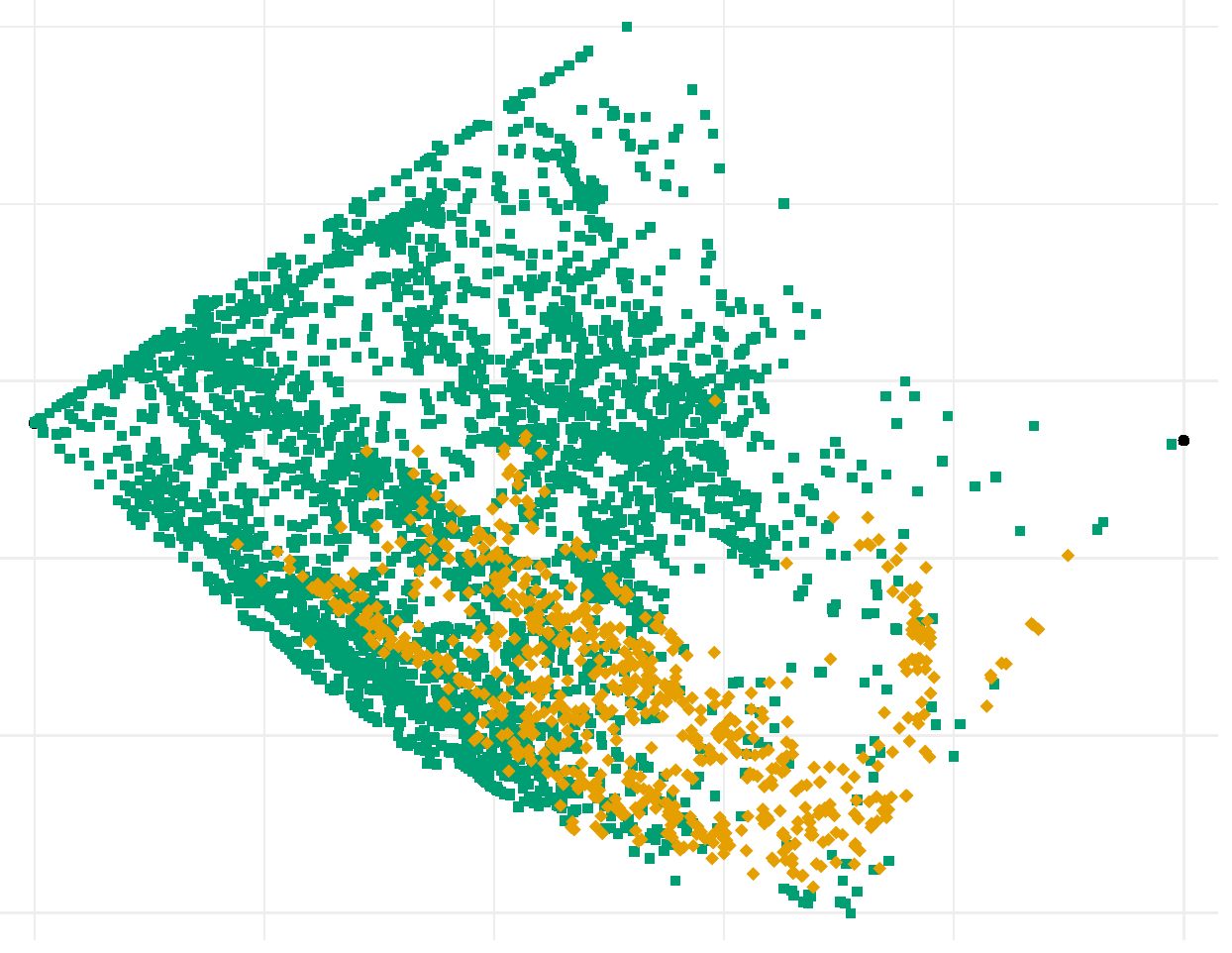}} 
  & \makecell{\includegraphics{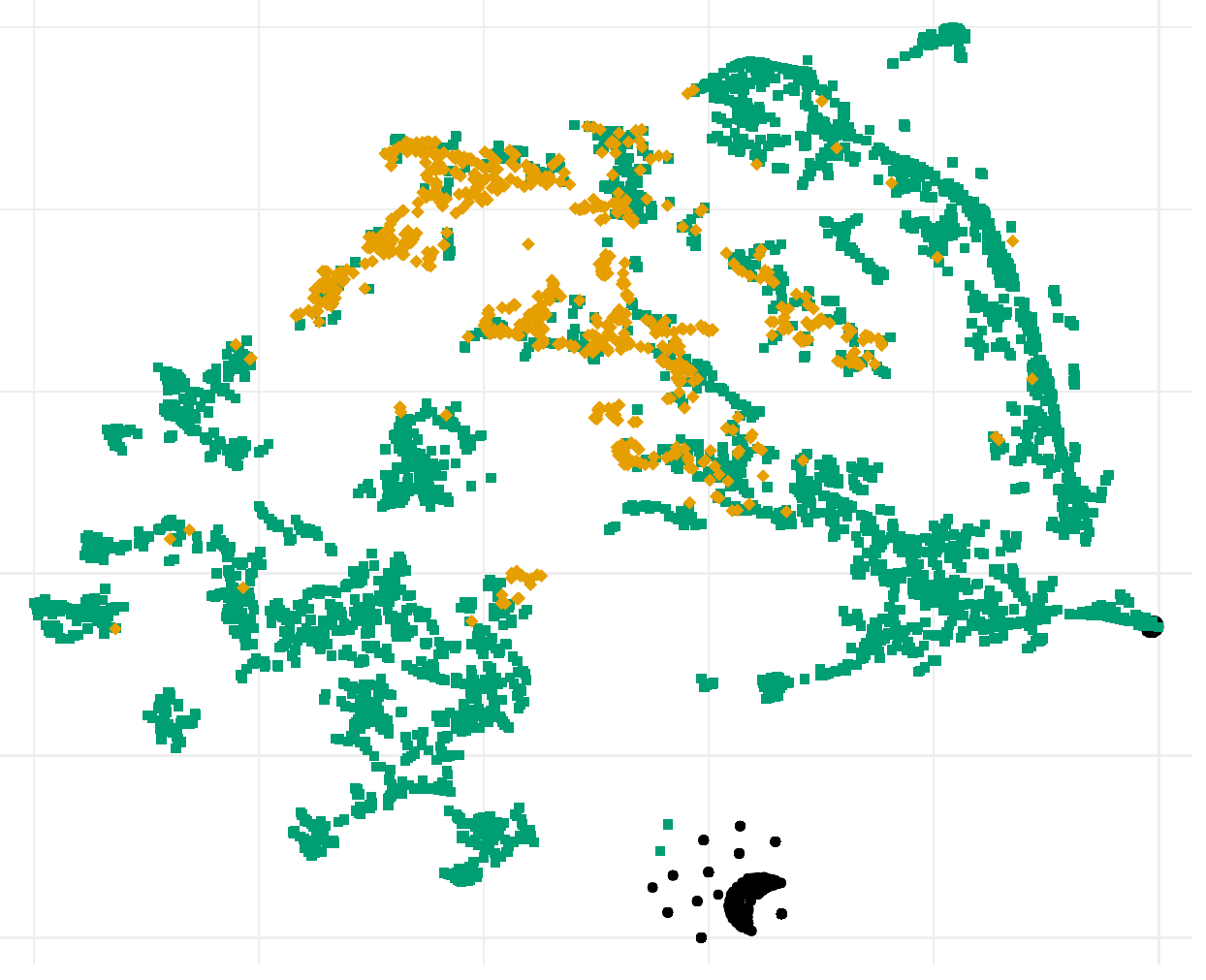}}\\

 \huge{\textsc{pgl}}
  & \makecell{\includegraphics{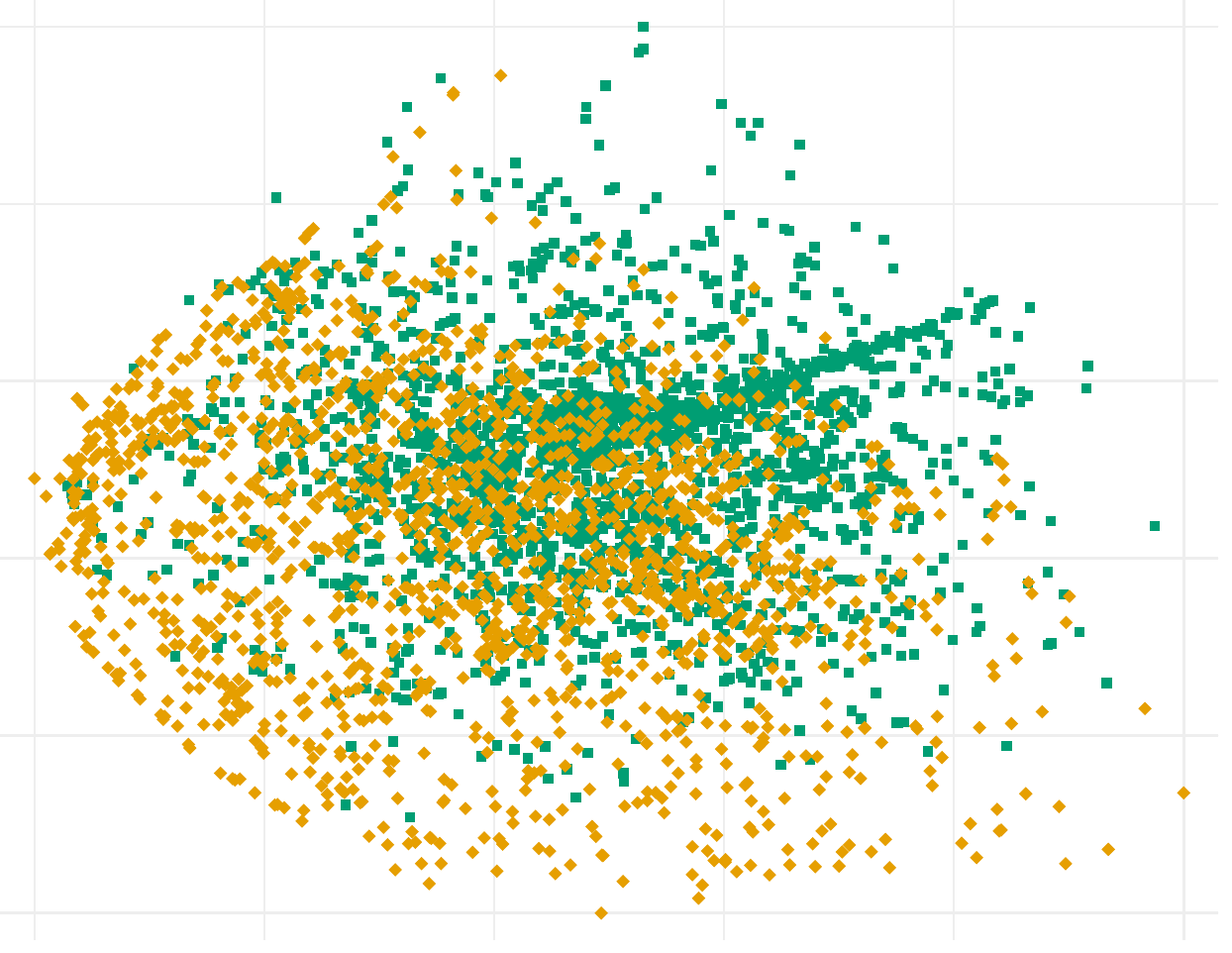}} 
  & \makecell{\includegraphics{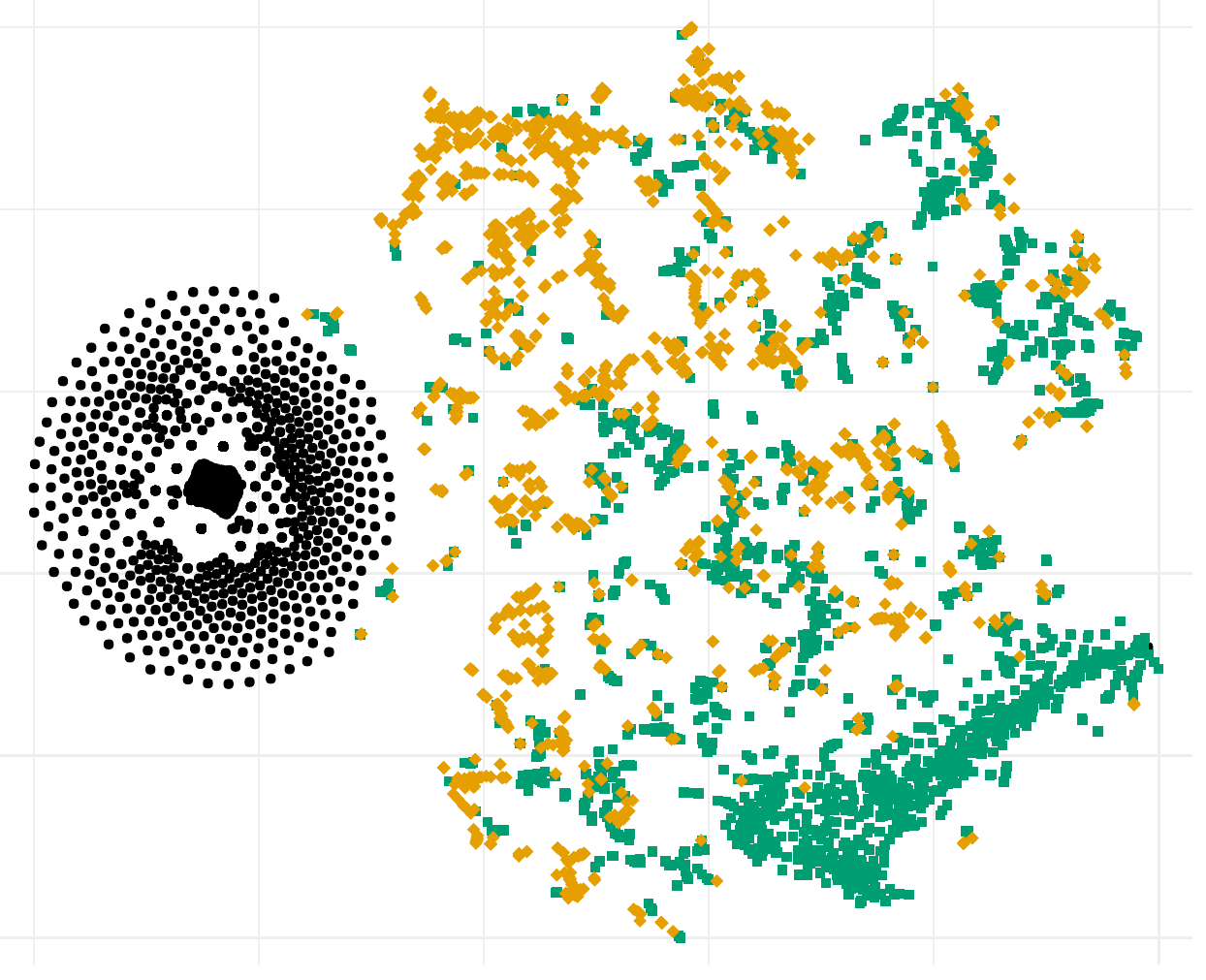}}\\

  \huge{\textsc{ogl}}
  & \makecell{\includegraphics{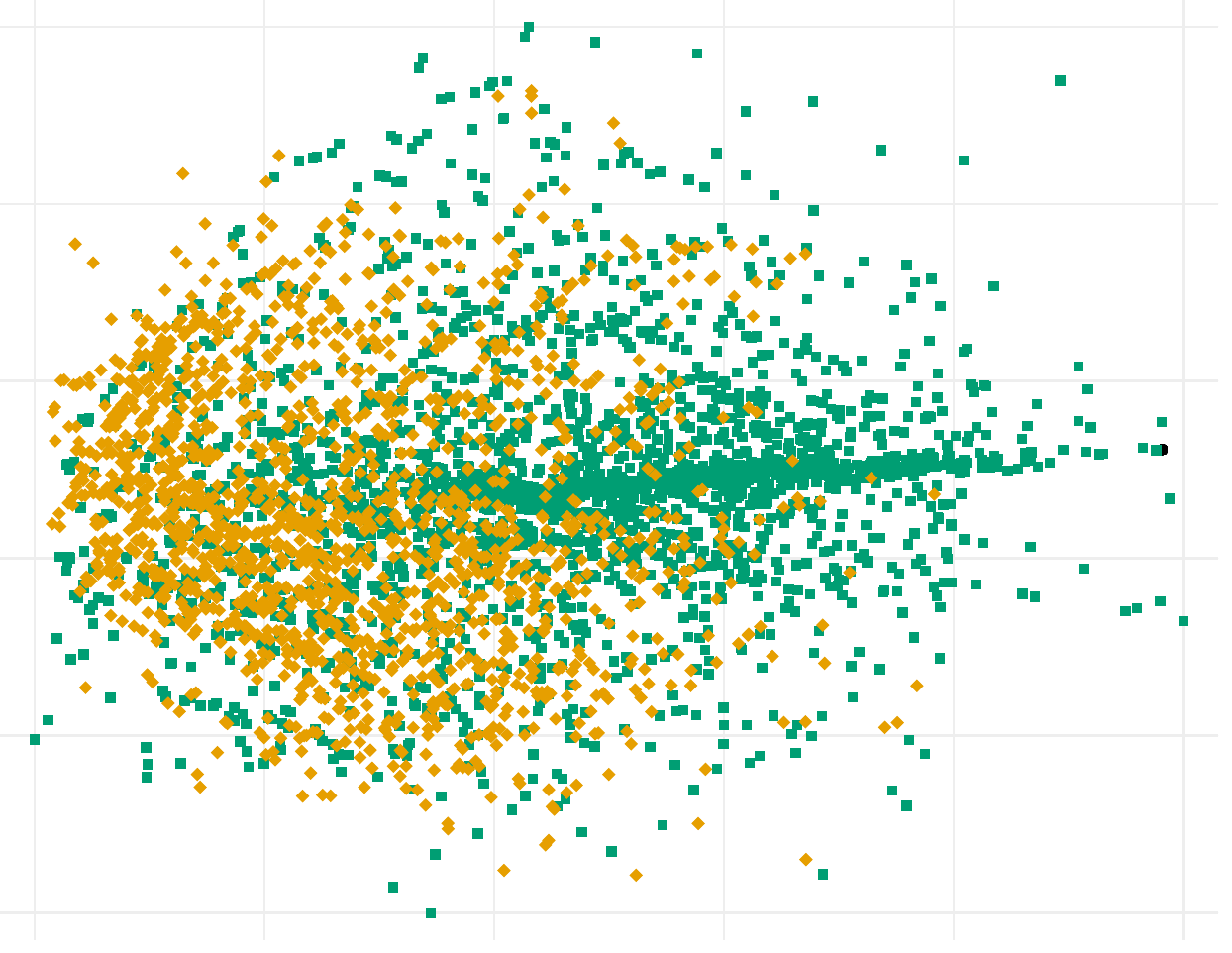}} 
  & \makecell{\includegraphics{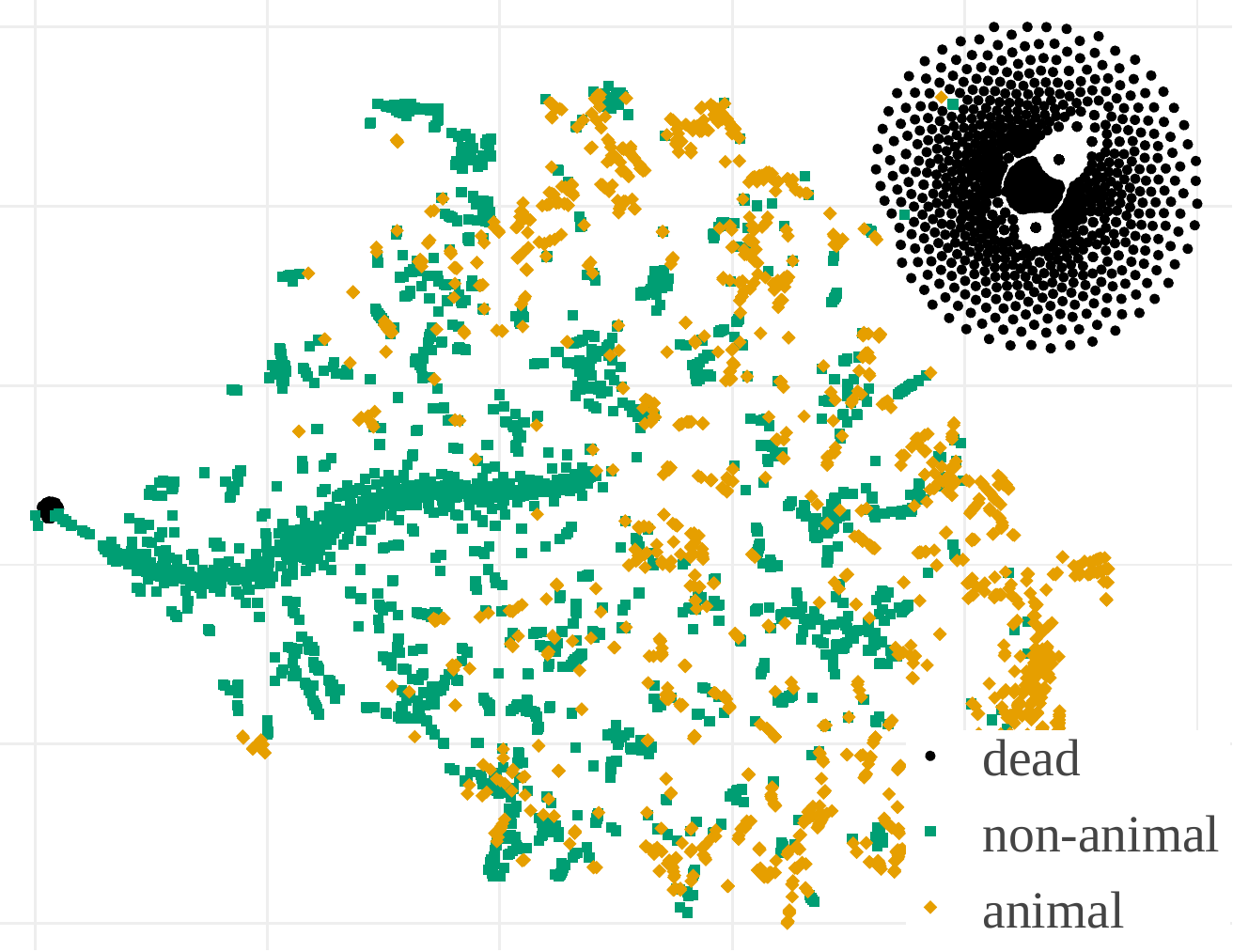}}\\

\end{tabular}
}
 \caption{
PCA and t-SNE visualization of the goal spaces for the IMGEP variants show that HGS has more area (PCA) and clusters (t-SNE) for non-animals compared to learned goal spaces (PGL and OGL) and vice versa for animals. t-SNE shows that the hand-defined goal space (HGS) and learned goal spaces (PGL and OGL) structure and cluster more the discovered patterns compared to random goal space (RGS).}
 
\label{fig::visualization_goal_spaces}
\end{figure}

The goal spaces of all IMGEP algorithms are visualized (Fig.~\ref{fig::visualization_goal_spaces}) via two-dimensional reductions: PCA \citep{Jolliffe1986} and t-Distributed Stochastic Neighbor Embedding (t-SNE) \citep{maaten2008visualizing}. 
The visualizations were constructed by using for each algorithm its goal space representations of all its discovered patterns from a single repetition experiment.
All goal representations were normalized so that the overall minimum value became 0 and the maximum value 1 for each dimension.
PCA was performed to detect the two principle components.
T-SNE was executed using the default standard Euclidean distance metric and perplexity set to 50.  

\subsection{Proportion of Discovered Patterns of Different Classes}

\begin{figure}[ht!]
  \centering
  \includegraphics[height=5.85cm]{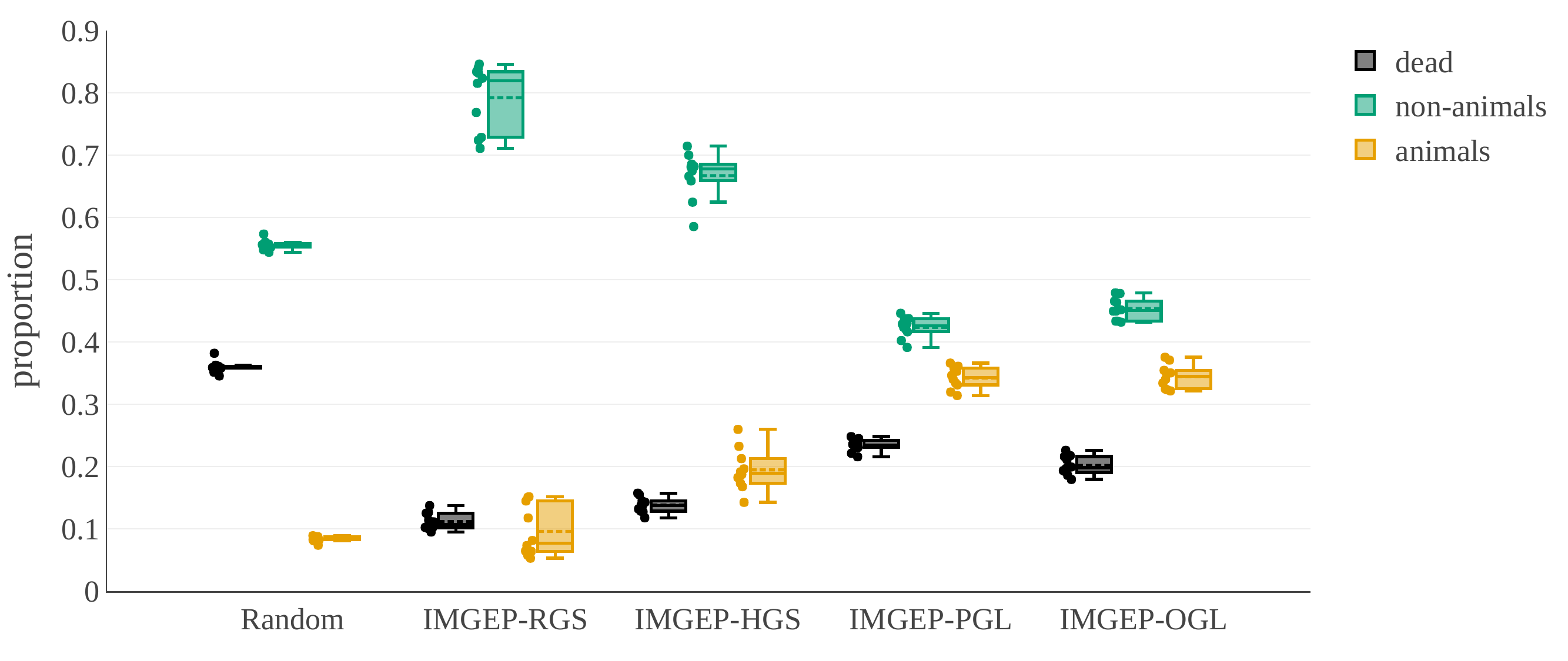}
  \caption{
    Proportion of patterns for each class and algorithm. 
    Each dot besides the boxplot shows the proportion of found patterns for each repetition ($n=10$).
    The box ranges from the upper to the lower quartile. 
    The whiskers represent the upper and lower fence.
    The mean is indicated by the dashed line and the median by the solid line.
    }
  \label{fig:results_number_of_identified_patterns}
\end{figure}

We used the measure of diversity of the found patterns to compare the performance of algorithms.
Another measure to compare the algorithms is the average proportion of dead, animals, non-animals patterns discovered by each algorithm (Fig.~\ref{fig:results_number_of_identified_patterns}).
For animal patterns the results follow the diversity results (Fig.~\ref{fig:results_diversity}), i.e. OGL and PGL find the highest proportion of animal patterns, followed by HGS, then RGS and Random. 
A corollary is that RGS, random and HGS find in proportion a higher percentage of non-animal patterns than OGL and PGL. 
As the number of \textit{different} non-animal patterns is as high for OGL and PGL, this shows the higher sample efficiency of OGL and PGL to find diverse patterns. 
In contrast, Random, RGS and HGS approaches tend to find non-animal patterns that are more similar to each other on average.

\begin{figure}[p!]
  \centering
  \textbf{Random Exploration}
  \includegraphics[width=\textwidth]{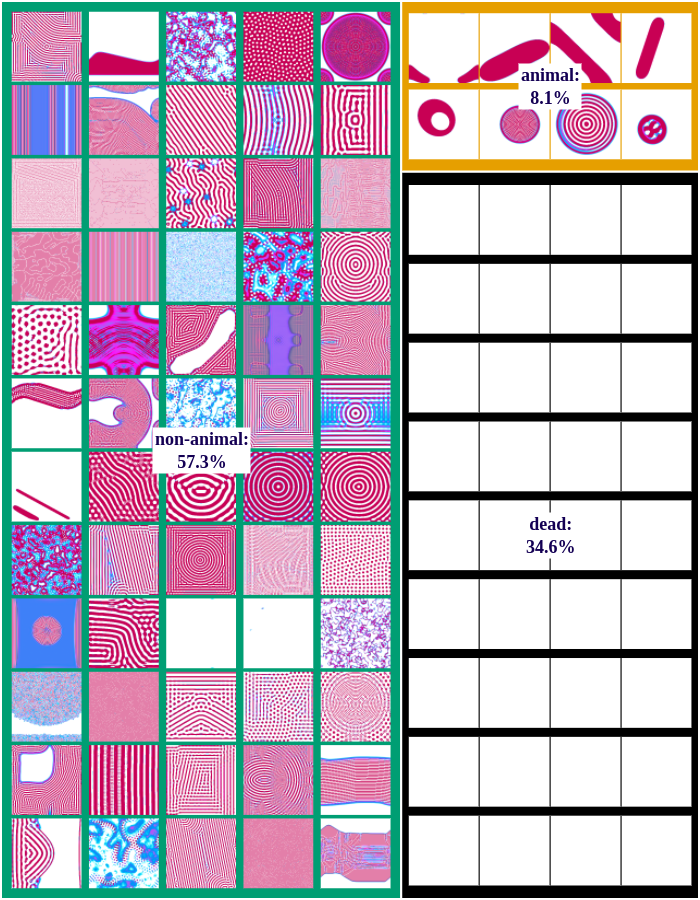}
  \caption{Randomly selected examples of patterns discovered by the random exploration algorithm during a single exploration with 5000 iterations.}
  \label{fig:identified_patterns_random}
\end{figure}

\begin{figure}[p!]
  \centering
  \textbf{IMGEP-HGS}
  \includegraphics[width=\textwidth]{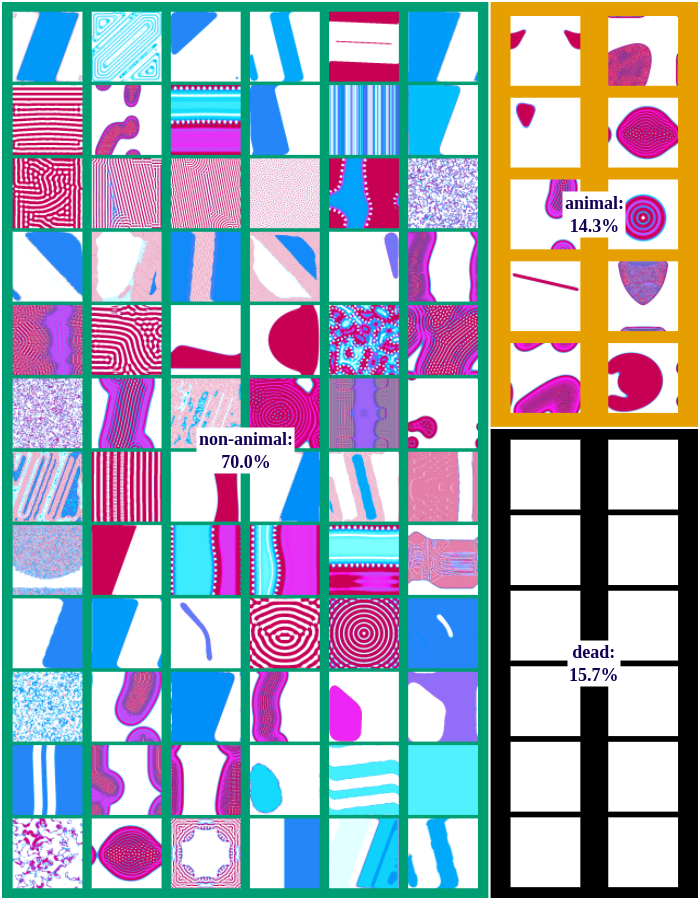}
  \caption{Randomly selected examples of patterns discovered by the IMGEP-HGS algorithm during a single exploration with 5000 iterations.}
  \label{fig:identified_patterns_hgs}
\end{figure}

\begin{figure}[p!]
  \centering
  \textbf{IMGEP-PGL}
  \includegraphics[width=\textwidth]{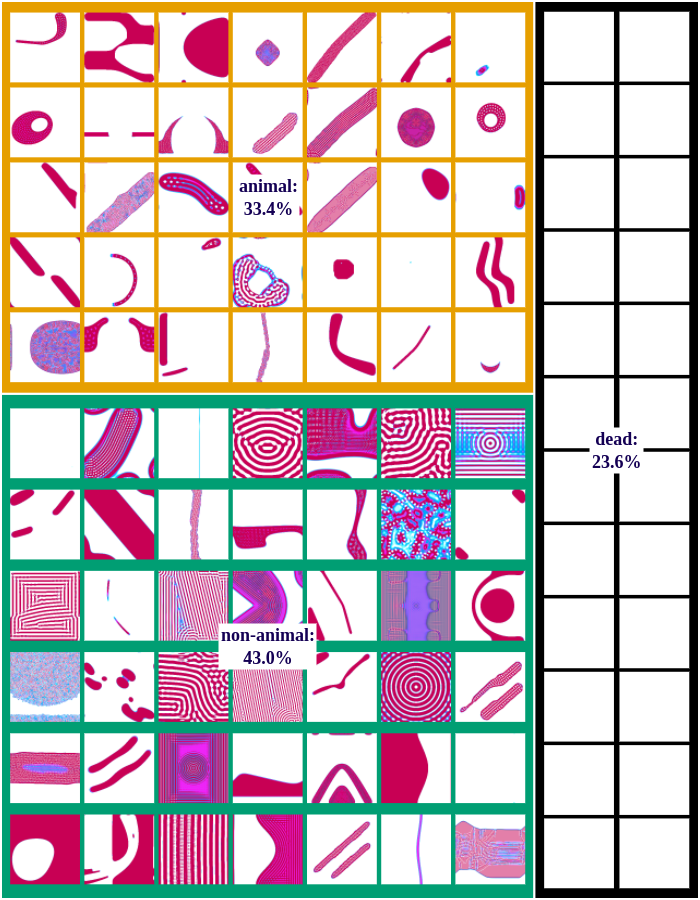}
  \caption{Randomly selected examples of patterns discovered by the IMGEP-PGL algorithm during a single exploration with 5000 iterations.}
  \label{fig:identified_patterns_pgl}
\end{figure}

\begin{figure}[p!]
  \centering
  \textbf{IMGEP-OGL}
  \includegraphics[width=\textwidth]{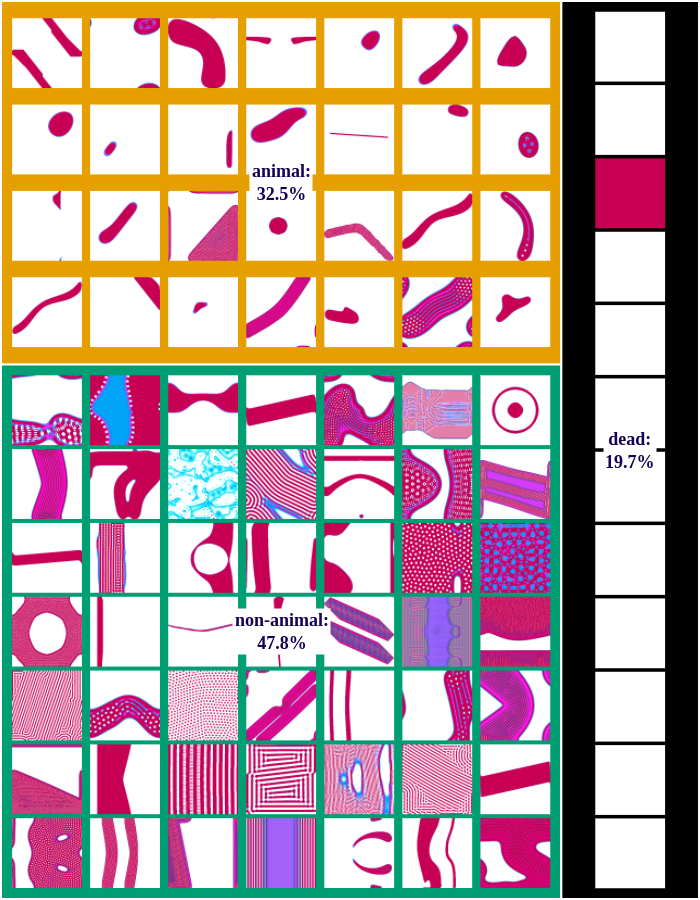}
  \caption{Randomly selected examples of patterns discovered by the IMGEP-OGL algorithm during a single exploration with 5000 iterations.}
  \label{fig:identified_patterns_ogl}
\end{figure}

\newpage
\section{Implementation Details and Hyperparameter Settings}
\label{c:sm_implementation_details}

This section provides implementation details and hyperparameter settings of all experiments. The content is organized in the following manner:
\begin{itemize}
    \item \ref{c:sm_lenia}: Settings of Lenia 
    
    \item \ref{c:sm_classifier}: Description of the classifiers for dead, animal and non-animal Lenia patterns
    
    \item \ref{c:lenia_statistical_measures}: Description of the statistical measures about Lenia patterns used as features for the IMGEP-HGS goal space and the analytic behavior space.
    
    \item \ref{c:sm_sampling_lenia_parameter}: Sampling mechanisms for Lenia's initial state via CPPNs and Lenia's other dynamic parameters.
    
    \item \ref{c:sm_hgs}: Details about the IMGEP-HGS
    
    \item \ref{c:sm_imgep_learned_goalspaces}: Details about IMGEPs utilizing deep variational autoencoders (RGS, PGL, OGL).
    
    \item \ref{c:sm_definition_analytic_spaces}: Description of the measurement method of diversity in the analytic parameter and behavior space
\end{itemize}



\subsection{Lenia Settings}
\label{c:sm_lenia}

A full description of Lenia can be found in \citep{chan2019lenia}.
For all experiments the following configurations of Lenia were used:
\begin{itemize}
	\item Grid size: $256 \times 256$ ($A \in [0,1]^{256 \times 256}$)
	\item Number of steps: $M = 200$
	\item Exponential growth mapping: $G(u;\mu,\sigma) = 2 \exp \left(- \frac{(u-\mu)^2}{2 \sigma^2}  \right)-1$
	\item Exponential kernel function: $K_C(r) = \exp \left( \alpha - \frac{\alpha}{4 r (1-r)}\right)$, with $\alpha=4$
	\item Kernel shell: $K_S(r;\beta) = \beta_{\lfloor Br \rfloor} K_C(Br~mod~1)$, with $\beta = (\beta_1, \beta_2, \beta_3)$
\end{itemize}
The controllable parameters of Lenia are the kernel size $R \in \sN$, time step $T \in \sN$, $\mu \in \sR$ and $\sigma \in \sR$ that control the growth mapping, and $\beta_1$, $\beta_2$, $\beta_3 \in \sR$ that control the kernel shell. 
Additionally the initial state $A^{t=1} \in [0,1]^{256 \times 256}$ controls the system dynamics.


\subsection{Classification of Lenia Patterns}
\label{c:sm_classifier}

We classified 3 types of Lenia patterns: dead, animals and non-animals.
The categories were used to analyze if the exploration algorithms showed differences in their behaviors by identifying different types of patterns. 
The classifiers only categorize the final pattern into which the Lenia system morphs after $M = 200$ time steps.

\textbf{Dead Classifier:}
For dead patterns the activity of all cells are either $0$ or $1$.

\textbf{Animal Classifier:}
A pattern is classified as an animal if it is a \textit{finite} and \textit{connected} pattern of activity.
Cells $x$, $y$ are connected as a pattern if both are active ($A(x) \geq 0.1$ and $A(y) \geq 0.1$) and if they influence each other.
Cells influence each other when they are within their radius of the kernel $K$ as defined by the parameter $R$. 

\begin{figure}[b!]
\centering
\setlength\tabcolsep{1pt}
\renewcommand{\arraystretch}{0.5}
\begin{tabular}{@{} ccc c|c ccc @{}}
 \multicolumn{3}{c}{(a) Infinite Pattern}
&
&
 \multicolumn{3}{c}{(a) Finite Pattern}  
  \\ 
 
 ~ & & & & & & \\
 
 pattern &
 \makecell{infinite \\ segmentation} &
 \makecell{finite \\ segmentation}  &
 &
 pattern &
 \makecell{infinite \\ segmentation} &
 \makecell{finite \\ segmentation}
 \\
 
 \fbox{\includegraphics[width=2.1cm]{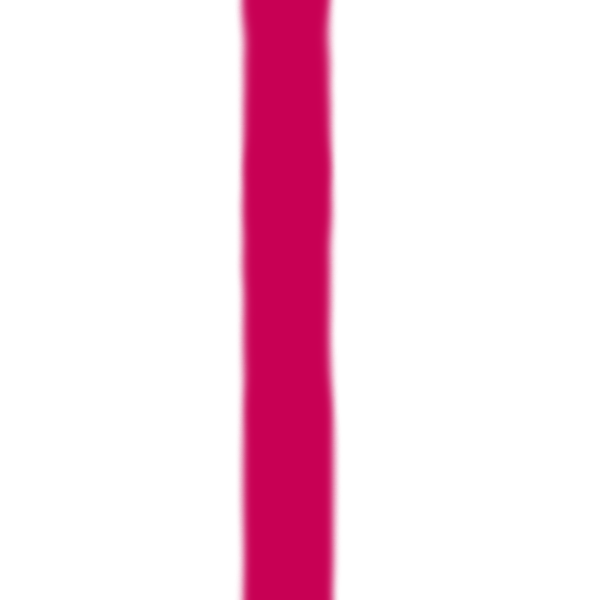}} &
 \fbox{\includegraphics[width=2.1cm]{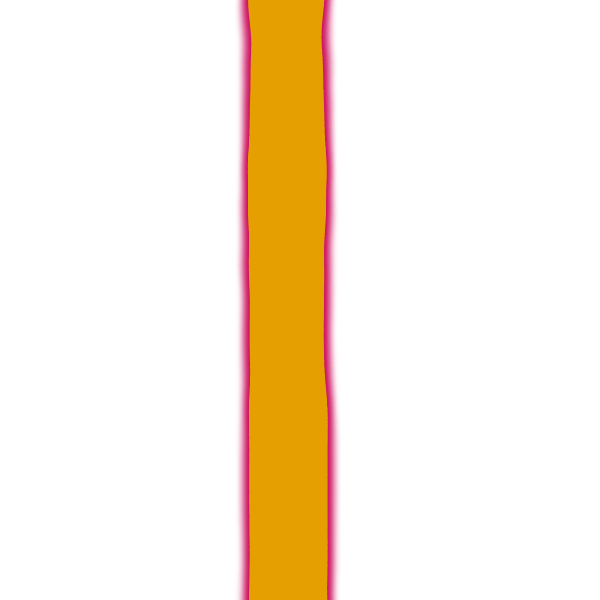}} &
 \fbox{\includegraphics[width=2.1cm]{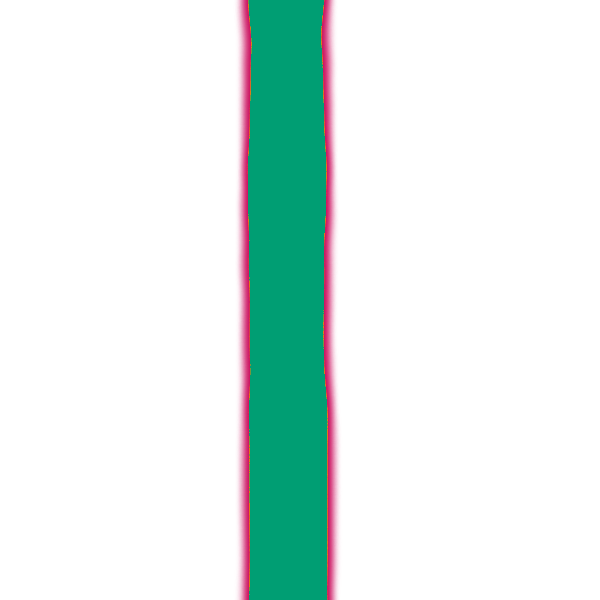}} & 
 &
 ~\fbox{\includegraphics[width=2.1cm]{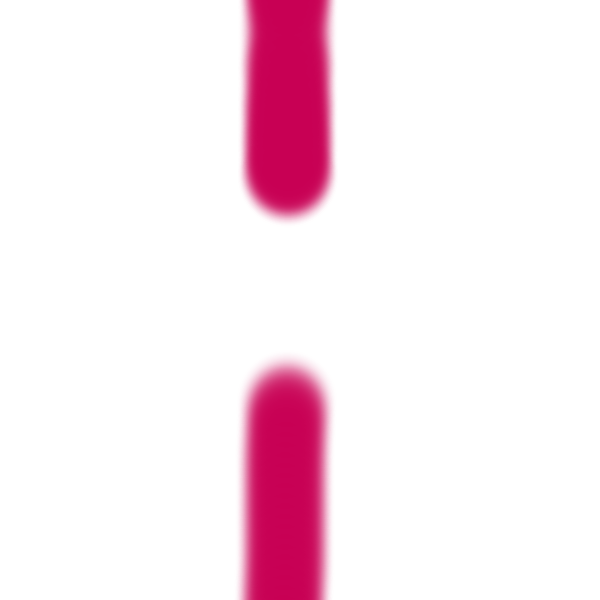}} &
 \fbox{\includegraphics[width=2.1cm]{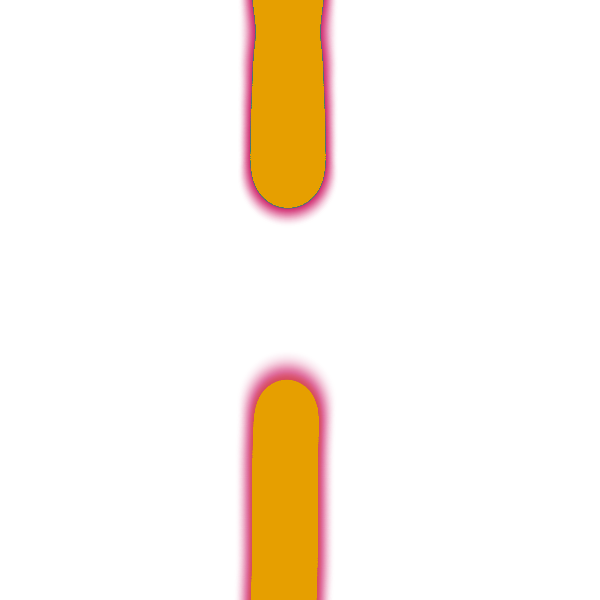}} &
 \fbox{\includegraphics[width=2.1cm]{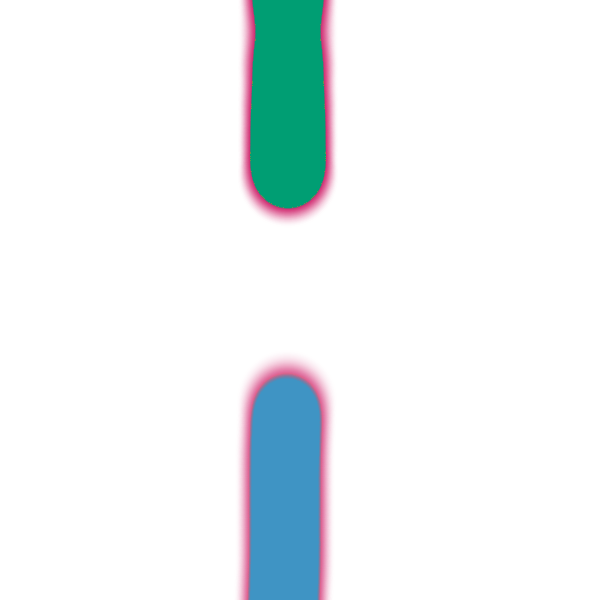}} 
 \\
 
\end{tabular}
 \caption{
 Classification of finite and infinite patterns. 
 Infinite patterns form loops between image borders which are identified if a segment is connected between two borders in the infinite and finite segmentation.
 Finite patterns form no loops.
 They have connected segments between borders in the infinite but not finite segmentation.
 Segments are colorized in yellow, green and blue.} 
 \label{fig:lenia_classifiers_animal}
\end{figure}

Furthermore, the connected pattern must be finite.
In Lenia finite and infinite patterns can be differentiated because the opposite borders of Lenia's cell grid are connected, so that the space is similar to a ball surface.
Thus, a pattern can loop around this surface making it infinite.
We identify infinite patterns by the following approach.
First, all connected patterns are identified for the case of assuming an infinite grid cell, i.e.\ opposite grid cell borders are connected.
Second, all connected patterns for the case of a finite grid cell, i.e.\ opposite grid cell borders are not connected, are identified.
Third, for each border pair (north-south and east-west) it is tested if cells within a distance of $R$ from both borders exists, that are part of a connected pattern for the infinite and finite grid cell case. 
If such a pattern exists than it is assumed to be infinite, because it loops around the grid cell surface of Lenia (Fig.~\ref{fig:lenia_classifiers_animal}, a).
All other patterns are considered to be finite (Fig.~\ref{fig:lenia_classifiers_animal}, b).
Please note that this method has a drawback.
It can not identify certain infinite patterns that loop over several borders, for example, if a pattern connects the north to east and then the west to south border (see the third animal in Fig.~\ref{fig:identified_patterns_random} for an example).


Moreover, there are two additional constraints that an animal pattern must fulfill.
First, the cells of the connected pattern $P = \{x_1, \ldots, x_n\}$ must have at least 80\% of all activation, i.e.\ $\sum_{x \in P} A(x) \geq 0.8 \sum_{\forall y} A(y)$.
Second, a pattern must exists for the last two time steps ($t = M$ and $t = M - 1$).
Both constraint are used to avoid that too small patterns or chaotic entities which change drastically between time steps are classified as animals.
See Figs.\ \ref{fig:identified_patterns_random}, \ref{fig:identified_patterns_hgs}, \ref{fig:identified_patterns_pgl} and \ref{fig:identified_patterns_ogl} for examples of animal patterns.

\textbf{Non-Animal Classifier:}
We also classified non-animal patterns which are all entities that were not dead and not an animal.
These patterns spread usually over the whole state space and are connected via borders.
See Figs.\ \ref{fig:identified_patterns_random}, \ref{fig:identified_patterns_hgs}, \ref{fig:identified_patterns_pgl} and \ref{fig:identified_patterns_ogl} for examples of non-animal patterns.

\subsection{Statistical Measures for Lenia Patterns}
\label{c:lenia_statistical_measures}

We defined five statistical measurements for the final patterns $A^{t=M}$ that emerge in Lenia.
The measures were used as features for hand-defined goal spaces of IMGEPs and to define partly the analytic behavior space in which the results of the exploration experiments were compared.

\textbf{Activation mass $M_A$}:
Measures the sum over the total activation of the final pattern and normalizes it according to the size of the Lenia grid: 
\begin{equation*}
	M_A = \frac{1}{L^2} \sum_x A^{t=M}(x) ~,
\end{equation*}
where $L^2 = 256 \cdot 256$ is the number of cells of the Lenia system.

\textbf{Activation volume $V_A$}:
Measures the number of active cells and normalizes it according to the size of the Lenia grid:
\begin{equation*}
	V_A = \frac{1}{L^2} \left|\left\{\forall x:A^{t=M}(x) > \epsilon\right\}\right| \textrm{ with }  \epsilon = 10^{-4}.
\end{equation*}

\textbf{Activation density $D_A$}: 
Measures how dense activation is distributed on average over active cells:
\begin{equation*}
	D_A = \frac{M_A}{V_A}.
\end{equation*}

\textbf{Activation asymmetry $A_A$}:
Measures how symmetrical the activation is distributed according to an axis that starts in the center of the patterns activation mass and goes along the last movement direction of this center.
This measure was introduced to especially characterize animal patterns.
The center of the activity mass is usually also the center of the animals and analyzing the activity along their movement axis measures how symmetrical they are.

As a first step, the center of the activation mass is computed for every time step of the Lenia simulation and the Lenia pattern recentered to this location. 
This ensures that the center is all the time correctly computed in the case the animal moves and reaches one border to appear on the opposite border in the uncentered pattern.
The center $(\bar{x},\bar{y})_t$ for time step $t$ is calculated by:
\begin{equation*}
    (\bar{x},\bar{y})_t =  \left( \frac{M_{10}}{M_{00}}, \frac{M_{01}}{M_{00}} \right) 
    \textrm{ with }
    M_{pq} = \sum_x \sum_y x^p y^q A^t(x,y)~,
\end{equation*}
where $M_{pq}$ measures the image moment (or raw moment) of order $(p+q)$ for $p,q \in \sN$.

Based on the center $(\bar{x},\bar{y})_t$ the pattern $A^t$ is recentered to $A_C^t$ by shifting the $x$ and $y$ indexes according to the center: 
\begin{equation}
    \label{eq:lenia_statistic_measures_centered_pattern}
    A^t_C(x,y) = A^t( (x-\bar{x})\,\mathrm{mod}\,{L}  ,(y-\bar{y})\,\mathrm{mod}\,{L}) ~,
\end{equation}
where $L$ is width and length of the Lenia grid and the indexing is $x,y = 0, \ldots, L-1$.
After each time step the center is recomputed and the pattern recentered: 
\begin{equation*}
    A^{t=1} \underset{\mathrm{recenter}}{\longrightarrow} A_C^{t=1} \underset{\mathrm{Lenia\;step}}{\longrightarrow} A^{t=2} \underset{\mathrm{recenter}}{\longrightarrow} A_C^{t=2} \underset{\mathrm{Lenia\;step}}{\longrightarrow} \ldots ~.    
\end{equation*}
Please note, the simulations and all figures of patterns in the paper are done with the uncentered pattern.
The centered version is only computed for the purpose of statistical measurements.

The recenter step by $(\bar{x},\bar{y})_t$ defines also the  movement direction of the activity center:
\begin{equation*}
    (m_x, m_y)_t = (\bar{x},\bar{y})_t - \left(x^\mathrm{mid},y^\mathrm{mid}\right) = \left(\bar{x}-x^\mathrm{mid},\bar{y}-y^\mathrm{mid}\right) ~,   
\end{equation*}
where $x^\mathrm{mid}, y^\mathrm{mid} = \frac{L-1}{2}$ are the coordinates for the grid's middle point.
A line can be defined that starts in the midpoint $\left(x^\mathrm{mid},y^\mathrm{mid}\right)$ of the final centered pattern $A_C^{t=M}$ and goes in and opposite to the movement direction of the activity mass center $(m_x, m_y)_{t=M}$.
This line separates the grid in two equal areas. 
The asymmetry is computed by comparing the activity amount in the grid right $M^{right}_A$ and left $M^{left}_A$ of the line.
The normalized difference between both sides is the final measure: 
\begin{equation*}
	A_A = \frac{1}{M_A} (M^{right}_A - M^{left}_A).
\end{equation*}

\textbf{Activation centeredness $C_A$}:
Measures how strong the activation is distributed around the activity mass center:
\begin{equation*}
	C_A = \frac{1}{M_A} \sum_x \sum_y w_{xy} \cdot A_C^{t=M}(x,y)
	~~\textrm{  with  }~~
	w_{xy} = \left(1 - \frac{d(x,y)}{\max_{y,x} d(x,y)}\right)^2 ~,
\end{equation*}
where $d(x,y) = \sqrt{(x-x^\mathrm{mid})^2 + (y-y^\mathrm{mid})^2}$ is the distance from the point $(x,y)$ to the center point $\left(x^\mathrm{mid},y^\mathrm{mid}\right)$.
$A_C^{t=M}(x,y)$ is the centered activation that is updated every time step as for the asymmetry measure (Eq.~\ref{eq:lenia_statistic_measures_centered_pattern}).
The weights $w_{xy}$ decrease the farer a point is from the center.
Thus, patterns that are concentrated around the center have a high value for $C_A$ close to $1$.
Whereas, patterns whos activity is distributed throughout the whole grid have a smaller value.
For patterns that are equally distributed ($\forall_{x,x'}: A(x) = A(x')$) is $C_A = 0$  defined as centeredness measure.

\subsection{Sampling of Parameters for Lenia}
\label{c:sm_sampling_lenia_parameter}

All algorithms explore Lenia by sampling the parameters $\theta$ that control Lenia. 
The parameters are comprised of the initial pattern $A^{t=1}$ and the parameters which control the dynamic behavior ($R, T, \mu, \sigma, \beta_1, \beta_2 ,\beta_3$).  
There are two operations to sample parameters: 
1) random initialization and
2) mutating an existing parameter $\theta$.
CPPNs are used for the random initialization and mutation of the initial pattern $A^{t=1}$.
The details of this process are described in the next section.
Afterwards, the initialization and mutation of Lenia's parameter that control its dynamics are described.

\subsubsection{Sampling of Start Patterns for Lenia via CPPNs}
\label{c:sm_initial_pattern_generation}

\begin{figure}[t!]
  \centering
  \includegraphics[scale=0.8]{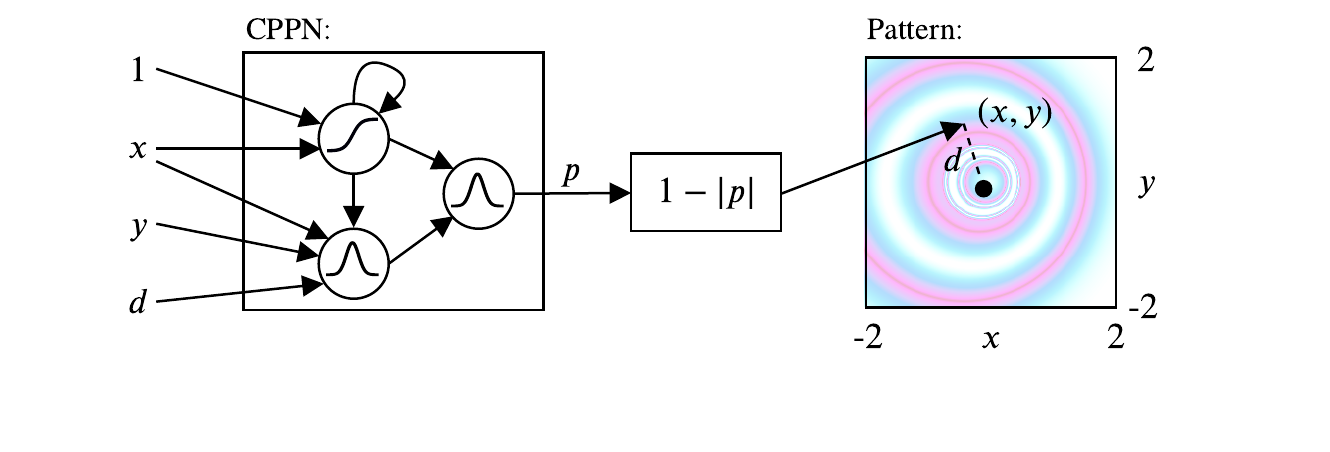}
  \caption{
    CPPNs are recurrent neural networks.
    Their input is a bias of $1$, the $x$ and $y$ coordinate of a grid cell and its center distance $d$. 
    Their output is the activity value of the grid cell.
    }
  \label{fig:cppn_illustration}
\end{figure}

Compositional Pattern Producing Networks (CPPNs) are recurrent neural networks \citep{stanley2006exploiting} which we to generate and mutate the initial state of Lenia $A^{t=1}$.
The CPPNs generate Lenia activity patterns cell by cell by taking as input a bias value, the $x$ and $y$ coordinate of the cell (mapped to $x=[-2, 2]$ and $y=[-2, 2]$) and its distance $d$ to the grid center (Fig.~\ref{fig:cppn_illustration}).
Their output $p$ defines the activity of the cell ($A(x,y) = 1 - |p|$) between $0$ and $1$ for the given $(x,y)$ coordinate.

CPPNs consist of several hidden neurons (typically 4 to 6 in our experiments) that can have recurrent and self connections. 
Each CPPN has one output neuron.
Two activation functions were used for the hidden neurons and the output neuron.
The first is Gaussian and the second is sigmoidal:
\begin{equation}
    \label{eq:cppn_activation_gauss}
    \textrm{gauss}(x) = 2 \exp\left(-(2.5 x)^2\right) - 1 ~,~~~~~~
    \textrm{sigm}(x) = 2 \left(\frac{1}{1 + \exp(-5 x)}\right) - 1 ~.
\end{equation}
To randomly initialize a Lenia initial pattern $A^{t=1}$ a CPPN is randomly sampled by sampling the number of hidden neurons, the connections between inputs and neurons and neurons to neurons, their connection weights and the activation functions for neurons.
Afterwards the initial pattern is generated by it.
In the history $\mathcal{H}$ of the IMGEPs is then the CPPN as part of the parameter $\theta$ added.
If the parameter is mutated, then the weights, connections and activation functions of the CPPN are mutated and the new initial pattern $A^{t=1}$ generated by it. 
A CPPN is defined over its network structure (number of nodes, connections of nodes) and its connection weights.

\begin{figure}[b!]
\centering
\setlength\tabcolsep{1pt}
\renewcommand{\arraystretch}{0.5}
\begin{tabular}{@{}cccccc@{}}

 Initialization &
 1\textsuperscript{st} Mutation &
 2\textsuperscript{nd} Mutation &
 3\textsuperscript{rd} Mutation &
 4\textsuperscript{th} Mutation &
 5\textsuperscript{th} Mutation \\

 ~ &
 ~ &
 ~ &
 ~ &
 ~ &
 ~ \\

 \fbox{\includegraphics[width=2.1cm]{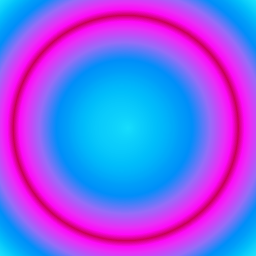}} &
 \fbox{\includegraphics[width=2.1cm]{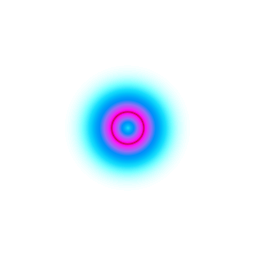}} &
 \fbox{\includegraphics[width=2.1cm]{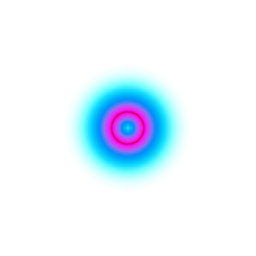}} &
 \fbox{\includegraphics[width=2.1cm]{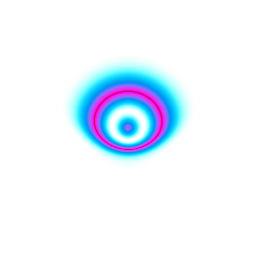}} &
 \fbox{\includegraphics[width=2.1cm]{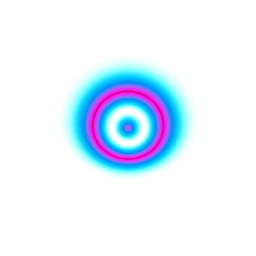}} &
 \fbox{\includegraphics[width=2.1cm]{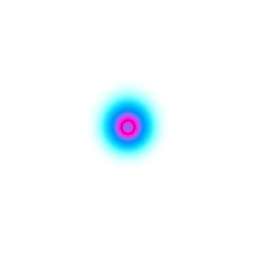}} \\

 \fbox{\includegraphics[width=2.1cm]{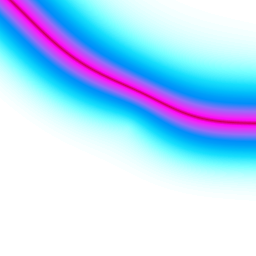}} &
 \fbox{\includegraphics[width=2.1cm]{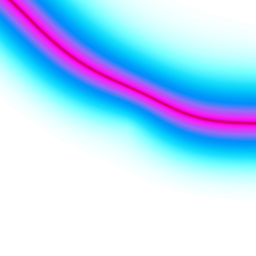}} &
 \fbox{\includegraphics[width=2.1cm]{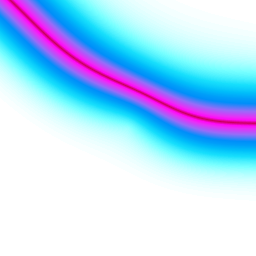}} &
 \fbox{\includegraphics[width=2.1cm]{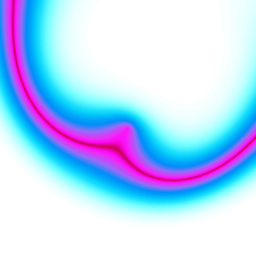}} &
 \fbox{\includegraphics[width=2.1cm]{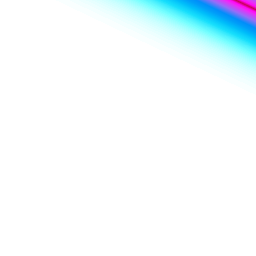}} &
 \fbox{\includegraphics[width=2.1cm]{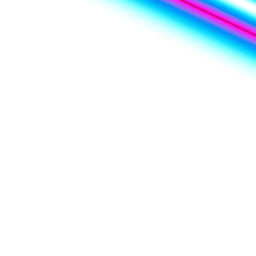}} \\

 \fbox{\includegraphics[width=2.1cm]{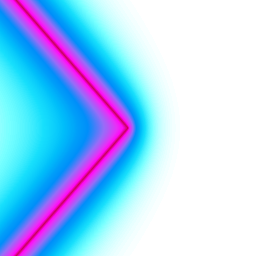}} &
 \fbox{\includegraphics[width=2.1cm]{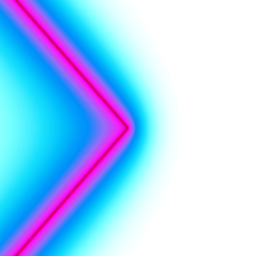}} &
 \fbox{\includegraphics[width=2.1cm]{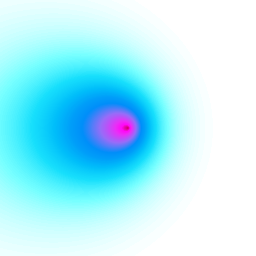}} &
 \fbox{\includegraphics[width=2.1cm]{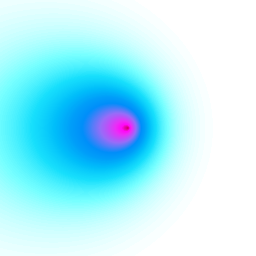}} &
 \fbox{\includegraphics[width=2.1cm]{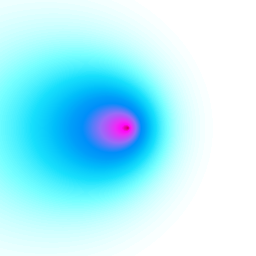}} &
 \fbox{\includegraphics[width=2.1cm]{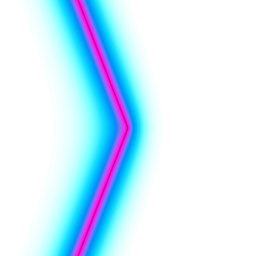}} \\

\end{tabular}
 \caption{
 CPPNs can generate complex patterns via their random initialization and successive mutations.
 Each row shows generated patterns by one CPPN and its mutations.} 
 \label{fig:cppn_evolution_examples}
\end{figure}

We used the \textit{neat-python}\footnote{\url{https://github.com/CodeReclaimers/neat-python}} package for the random generation and mutation of CPPNs.
It is based on the NeuroEvolution of Augmenting Topologies (NEAT) algorithm for the evolution of neural networks \citep{stanley2002efficient}.
The meta-parameters for the initialization and mutation of CPPNs are listed in Table~\ref{tbl:cppn_parameters}.
The random sampling and mutation of CPPNs allows to generate complex patterns as illustrated in Fig.~\ref{fig:cppn_evolution_examples}. 
Please note, the neat-python package allows also the setting and mutation of response and bias weights for each neuron. 
Those settings were not used for the experiments.
Moreover, we adjusted the sigmoid and Gaussian function in the neat-python package to the ones defined in Eq.~\ref{eq:cppn_activation_gauss} to be able to replicate similar images as in \cite{stanley2006exploiting}.

\subsubsection{Sampling of Lenia's Dynamic Parameters}
\label{c:sm_param_init_and_mutate}

The parameters that control the dynamics of Lenia ($R, T, \mu, \sigma, \beta_1, \beta_2 ,\beta_3$) are initialized and mutated via uniform and Gaussian distributions.
Table \ref{tbl:lenia_parameters} lists for each parameter the meta-parameters for their initialization and mutation.
Each parameter is initialized by an uniform sampling $\theta_i \sim \mathcal{U}(a, b)$ with $a$ and $b$ as upper and lower border.
An existing parameter $\theta_i$ is mutated by the following equation:
\begin{equation*}
    \theta_i \leftarrow \left[\theta_i + \mathcal{N}(0, \sigma_M)\right]^{a}_{b} ~,
\end{equation*}
where $\sigma_M$ is the mutation power and $[n]^a_b = \min(\max(n,a), b)$ is the clip function.
For natural numbers $\theta_i \in \sN$ the resulting value is rounded.

\begin{table}[t!]
    \centering
    \begin{tabular}{ l c }
        \toprule
        Parameter & Value \\ 
        \midrule
        Initial number of hidden neurons & $4$ \\
        Initial activation functions & gauss, sigm \\
        Initial connections & random connections with probability $0.6$ \\
        Initial synapse weight &  Gaussian distribution with $\mu=0$, $\sigma=0.4$ \\
        Synapse weight range &  $[-3, 3]$ \\
        Mutation neuron add probability & $0.02$ \\
        Mutation neuron delete probability & $0.02$ \\
        Mutation connection add probability & $0.05$ \\
        Mutation connection delete probability & $0.01$ \\
        Mutation rate of activation functions  & $0.1$ \\
        Mutation rate of synapse weights &  $0.05$ \\
        Mutation replace rate of synapse weights &  $0.06$ \\
        Mutation power of synapse weights $\sigma_M$ &  $1$ \\
        Mutation enable/disable rate of synapse weights &  $0.02$ \\
        \bottomrule
    \end{tabular}
    \vspace{0.2cm}
    \caption{Settings for the sampling of CPPNs to generate Lenia's initial states.}
    \label{tbl:cppn_parameters}
    \end{table}

\begin{table}[H]
    \centering
    \begin{tabular}{ c c c c }
        \toprule
        Parameter & Type & Value Range & Mutation $\sigma_M$\\ 
        \midrule
        $R$ & $\sN$ & $[2, 20]$ & $0.5$ \\
        $T$ & $\sN$ & $[1, 20]$ & $0.5$ \\
        $\mu$ & $\sR$ & $[0, 1]$ & $0.05$ \\
        $\sigma$ & $\sR$ & $[0.001, 0.3]$ & $0.01$ \\
        $\beta_1, \beta_2, \beta_3$ & $\sR$ & $[0, 1]$ & $0.05$ \\
        \bottomrule
    \end{tabular}
    \vspace{0.2cm}
    \caption{Settings for the initialization and mutation of Lenia system parameters $\theta$.}
    \label{tbl:lenia_parameters}
\end{table}


\subsection{IMGEP-HGS Details}
\label{c:sm_hgs}

The 5 statistical features that are given in Appendix~\ref{c:lenia_statistical_measures} were used to define the goal space of the IMGEP-HGS approach (Algorihm~\ref{algo:sm_IMGEP_baselines}). 
Goals in this space were sampled from a uniform distribution within the ranges defined in Table~\ref{tbl:sm_hgs_goalspace_ranges}.

    

\begin{table}[H]
    
    \centering
    \begin{tabular}{ l c c | l c c }
        \toprule
        Feature 		        & min & max & 
        Feature 		        & min & max \\ 
        \midrule
        
        mass $M_A$ 			    & $0$ & $1$  &
        asymmetry $A_A$  		& $-1$ & $1$  \\
        
        volume $V_A$ 		    & $0$ & $1$  &
        centeredness $C_A$  	& $0$ & $1$  \\
        
        density $D_A$			& $0$ & $1$  &
        & & \\
        
        \bottomrule
    \end{tabular}
    \vspace{0.1cm}
    \caption{HGS Goal Space Ranges}
    \label{tbl:sm_hgs_goalspace_ranges}
\end{table}


\subsection{IMGEPs with Deep Variational Autoencoders}
\label{c:sm_imgep_learned_goalspaces}

\subsubsection{VAE Framework and Implementation Details}

We applied deep variational autoencoders (VAEs) to learn latent representations of Lenia patterns.
VAEs have two components: a neural \textit{encoder} and \textit{decoder}.
The encoder $q(\mathbf{z}|\mathbf{x}, \chi)$ represents a given data point $x$ in a latent representation $\mathbf{z}$.
In variational approaches the encoder describes a data point by a representative distribution in the latent space of reduced dimension $d$.
A standard Gaussian prior $p(\mathbf{z})=\mathcal{N}(0,I)$ and a diagonal Gaussian posterior $q(\mathbf{z}|\mathbf{x}, \chi)=\mathcal{N}(\mu,\sigma)$ are used.
Given a data point $x$, the encoder outputs the mean $\mu$ and variance $\sigma$ of the representative distribution in the latent space.
The decoder $p(\mathbf{x}|\mathbf{z}, \psi)$ tries to reconstruct the original data $x$ from a sampled latent representation $\mathbf{z}$ for the distribution given by the encoder. 
Under these assumptions, training is done by maximizing the computationally tractable evidence lower bound (with $\beta = 1$):
\begin{equation}
    \label{eq::VAEobjective}
        \mathcal{L}(\chi, \psi) = \underbrace{\mathbb{E}_{\mathbf{z} \sim q_\chi(\mathbf{z}|\mathbf{x})}[\log p_\psi(\mathbf{x}|\mathbf{z})]}_a - \beta \times \underbrace{\mathbb{D}_{KL} [q_\chi(\mathbf{z}|\mathbf{x}) \Vert p(\mathbf{z})]}_b ~.
    \end{equation}
The first term ($a$) represents the expected reconstruction accuracy while the second ($b$) is the KL divergence of the approximate posterior from the prior. 
\begin{equation}
    \label{eq::factorizedBterm}
    b = \mathbb{D}_{KL} [\mathcal{N}(\mu(\mathbf{x}),\Sigma(\mathbf{x})) \Vert \mathcal{N}(0,I)] = \displaystyle{\sum_{i=1}^d} \: \underbrace{\mathbb{D}_{KL} [\mathcal{N}(\mu(\mathbf{x})_i,\sigma(\mathbf{x})_i) \Vert \mathcal{N}(0,1)]}_{b_i} ~.
\end{equation}

In particular, we used the $\beta$-VAE framework \citep{higgins2017beta} which re-weights the $b$ term by a factor $\beta > 1$, aiming to enhance the disentangling properties of the learned latent factors. 
The weight factor was set to $\beta = 5$ for all experiments.
All $\beta$-VAEs used for this paper use the same architecture (Table~\ref{tab::VAE_architecture}).
The encoder network has as input the Lenia pattern and as outputs for each latent variable $\mathbf{z}_i$ the mean $\mu_i$ and log-variance $\log(\sigma_i^2)$.
The decoder takes as input during the training for each latent variable a sampled value $z_i \sim \mathcal{N}(\mu_i, \sigma_i^2)$.
For validation runs and the generation of all reconstructed patterns shown in figures the decoder takes the mean $z_i = \mu_i$ as input.
Its output is the reconstructed pattern. 

\begin{table}[b!]
	\centering
    \resizebox*{1.0\textwidth}{!}{%
	\begin{tabular}{ll}
   & \\
   \textbf{Encoder} & \textbf{Decoder} \\
   \cmidrule(lr){1-1} \cmidrule(lr){2-2}
   Input pattern A: $256\times256\times1$ & Input latent vector z: $8\times1$ \\
   Conv layer: 32 kernels $4\times4$, stride $2$, $1$-padding + ReLU & FC layers : 256 + ReLU,  $16\times16\times32$ + ReLU \\
   Conv layer: 32 kernels $4\times4$, stride $2$, $1$-padding + ReLU & TransposeConv layer: 32 kernels $4\times4$, stride $2$, $1$-padding + ReLU \\
   Conv layer: 32 kernels $4\times4$, stride $2$, $1$-padding + ReLU & TransposeConv layer: 32 kernels $4\times4$, stride $2$, $1$-padding + ReLU \\
   Conv layer: 32 kernels $4\times4$, stride $2$, $1$-padding + ReLU & TransposeConv layer: 32 kernels $4\times4$, stride $2$, $1$-padding + ReLU \\
   FC layers : 256 + ReLU, 256 + ReLU, FC: $2\times8$ & TransposeConv layer: 32 kernels $4\times4$, stride $2$, $1$-padding \\
\end{tabular}}
    \vspace{0.2cm}
	\caption{$\beta$-VAE architecture for the pretrained and online experiments.}
\label{tab::VAE_architecture}
\end{table}

The training objective (Eq.~\ref{eq::VAEobjective}) results in the following loss function for a batch:
\begin{equation}
\mathrm{Loss}(x, \hat{x}, \mu, \sigma ) = -a + \beta \sum_{i=1}^d b_i ~,
\end{equation}
where $x$ are the input patterns, $\hat{x}$ are the reconstructed patterns, $\mu, \sigma$ are the outputs of the decoder network and $d$ is the number of latent dimensions. 
The reconstruction accurray part $a$ of the loss is given by a binary cross entropy with logits:
\begin{equation*}
a = \frac{1}{N} \sum_{n=1}^N \sum_{j=1}^{L^2} \left( x_{j,n} \cdot \log\sigma(\hat{x}_{j,n}) + (1 - x_{j,n}) \cdot \log(1 - \sigma(\hat{x}_{j,n})) \right) ~,  
\end{equation*}
where the index $j$ is for the single cells (pixel) of the pattern and $n$ for the datapoint in the current batch, $N$ is the batch size and $\sigma(x) = \frac{1}{1+e^{-x}}$.
The KL divergence terms $b_i$ are given by:
\begin{equation*}
b_i = \frac{1}{2 \cdot N } \sum_{n=1}^N \left( \sigma_{i,n}^2 + \mu_{i,n}^2 - \log(\sigma_{i,n}^2) - 1 \right) ~.
\end{equation*}

All $\beta$-VAEs were trained for 2000 epochs and initialized with pytorch default initialization. 
We used the Adam optimizer \citep{kingma2014adam} ($lr=1\mathrm{e}{-3}$, $\beta_1=0.9$, $\beta_2=0.999$, $\epsilon=1\mathrm{e}{-8}$, weight decay=$1\mathrm{e}{-5}$) with a batch size of 64.

The patterns from the datasets were augmented by random x and y translations (up to half the pattern size and with probability 0.3), rotation (up to 40 degrees and with probability 0.3), horizontal and vertical flipping (with probability 0.2). 
The translations and rotations were preceded by spherical padding to preserve Lenia spherical continuity. 

\subsubsection{IMGEP Variants}

\paragraph*{IMGEP-RGS (random goal space):} 
IMGEP with a goal space defined by an encoder network with random weights (Algorithm~\ref{algo:sm_IMGEP_baselines}).
The weights were initialized according to the Xavier method \citep{glorot2010understanding}.
The network architecture of the encoder is the same that the one of the VAEs used for IMGEP with learned goal spaces. In the other IMGEP algorithms (HGS/PGL/OGL), the goals are sampled uniformly within fixed-range boundaries that are chosen in advance. 
However, in the case of random goal spaces, we do not know in advance in which region of the space goals will be encoded. 
Therefore, we set the sampling range for each latent variable to the minimum and maximum value of the latent representations of all so far explored patterns.

\begin{algorithm}[b!]
\DontPrintSemicolon
Initialize goal space encoder $\mathcal{R}$ with handdefined features (HGS), random weights (RGS), or pretrained weights (PGL)\\
\For{$i\leftarrow 1$ \KwTo $N$}{
\If(\tcp*[f]{Initial random iterations to populate $\mathcal{H}$}){$i<N_{init}$} {
Sample $\theta \sim \mathcal{U}(\Theta)$
}
\Else(\tcp*[f]{Intrinsically motivated iterations}){
Sample a goal $g \sim \mathcal{G}(\mathcal{H})$ based on space represented by $\mathcal{R}$ \\
Choose $\theta \sim \Pi(g, \mathcal{H})$  \\
}
Perform an experiment with $\theta$ and observe $o$ \\
Append $(o, \theta, \mathcal{R}(o))$ to the history $\mathcal{H}$ \\
}
\caption{IMGEP-HGS, IMGEP-RGS and IMGEP-PGL}
\label{algo:sm_IMGEP_baselines}
\end{algorithm}

\paragraph*{IMGEP-PGL (prelearned goal space):} 
IMGEP with a goal space defined by a $\beta$-VAE that was trained before the exploration starts (Algorithm~\ref{algo:sm_IMGEP_baselines}).
The $\beta$-VAE was trained on a dataset consisting of 558 precollected Lenia patterns (training: 75\%, validation: 10\%, testing: 15\%). Half of the patterns (279) were manually identified animal patterns by \cite{chan2019lenia}. 
The other half (279) were randomly initialized CPPN patterns (see Section~\ref{c:sm_initial_pattern_generation}).
The best $\beta$-VAE model obtained during training (highest accuracy on the validation data) was used as goal space for the exploration. 
During the exploration, goals were uniformly sampled in a $[-3,3]^8$ hypercube. 
These values were chosen because the encoder of the $\beta$-VAE is trained to match a prior standard normal distribution.
Therefore, we can assume that most area of the covered goal space will fall into that hypercube.


\paragraph*{IMGEP-OGL (online learned goal space):}
IMGEP with a goal space that was online learned during the exploration via an $\beta$-VAE (Algorithm~1).
Every $K=100$ explorations the $\beta$-VAE model was trained for 40 epochs resulting in 2000 epochs in total (less if there is not enough data after the first $T$ runs to start the training). 
The datasets were constructed by collecting non-dead patterns during the exploration.
One pattern in every ten is added to the validation set (10\%).
All others are used for the training set. 
At the initial period of training, the training dataset has approximately 50 patterns and at the last period approximately 3425 patterns. 
The validation dataset only serves for checking purposes and has no influence on the learned goal space.
Importance sampling is used to give the patterns a different weight during the training.
A weighted random sampler is used that samples newly discovered patterns half of the time. 
Each pattern that has been added to the training dataset during the last period of 100 explorations has a probability of $\frac{0.5}{N}$ to be sampled (N is the total number of new patterns in the dataset). 
Older patterns are with probability $\frac{0.5}{|D_{\mathcal{T}}| - N}$.
As a result, newer discovered patterns have a higher weight and a stronger influence on the training. 

\subsubsection{VAE Training Results}

\begin{figure}[H]
\centering
\setlength\tabcolsep{2pt}
\renewcommand{\arraystretch}{1.0}
\begin{tabular}{@{}cc@{}}

IMGEP-PGL - $\beta$-VAE & IMGEP-OGL - $\beta$-VAE\\

\includegraphics[width=0.49\textwidth]{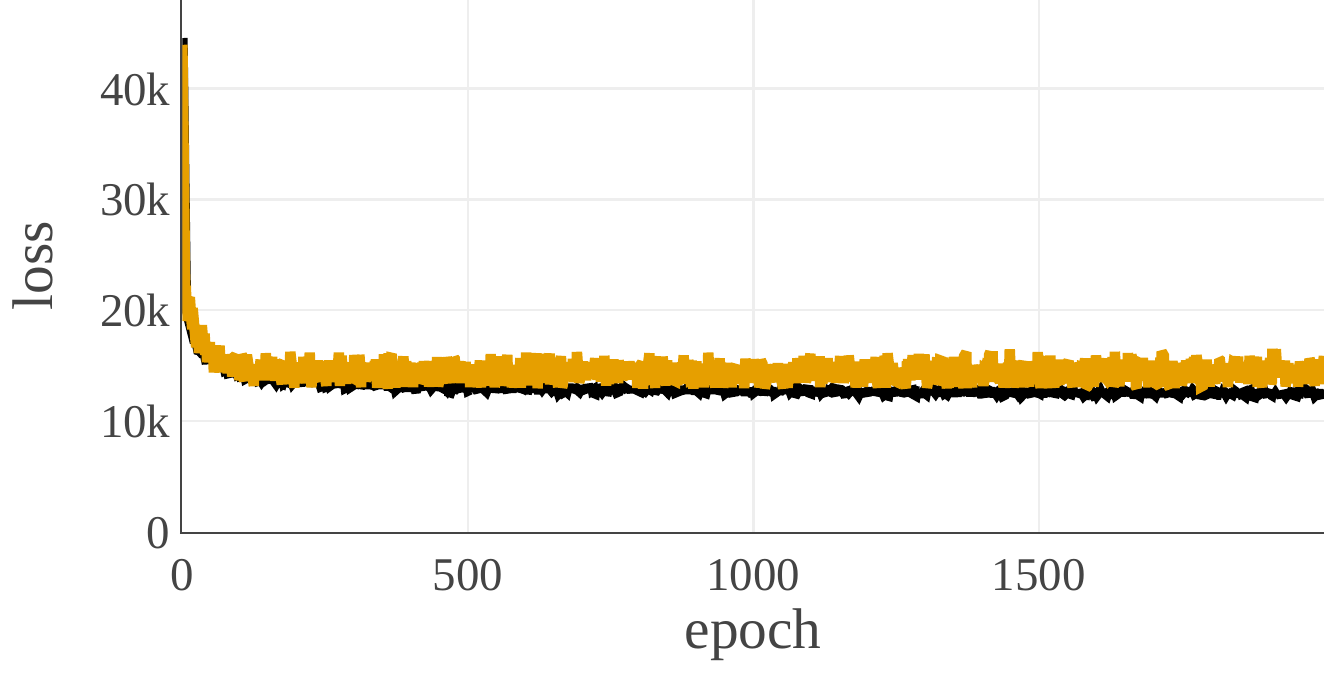} 
&
\includegraphics[width=0.49\linewidth]{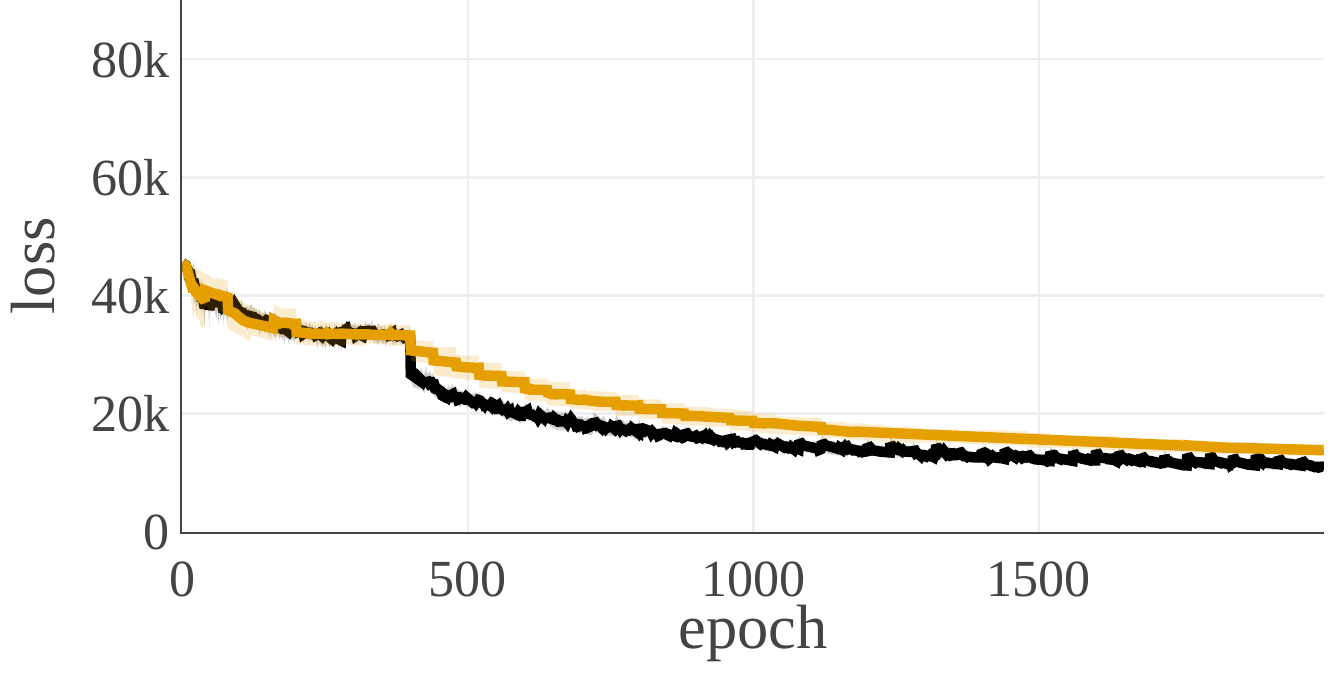} 
\\

\end{tabular}
 \caption{
 Averaged learning curves ($n=10$) of the $\beta$-VAEs for the IMGEP-PGL and OGL experiments.
 } 
 \label{fig:IMGEP-PGL_training}
\end{figure}

\begin{figure}[H]
\centering

a) Reconstruction Examples of the $\beta$-VAE used for the IMGEP-PGL
\vspace{0.1cm}

\setlength\tabcolsep{2pt}
\renewcommand{\arraystretch}{1.0}
\begin{tabular}{@{}ccc@{}}

 \fbox{\includegraphics[width=2.1cm]{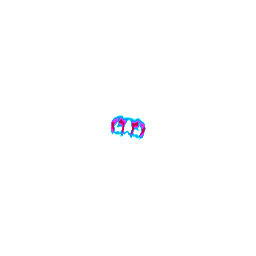} \includegraphics[width=2.1cm]{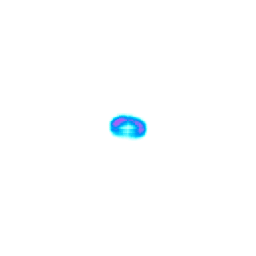}} &
 \fbox{\includegraphics[width=2.1cm]{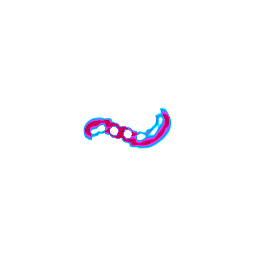}
 \includegraphics[width=2.1cm]{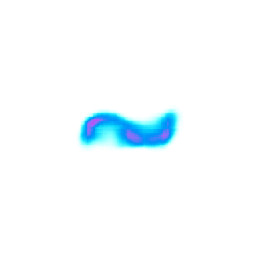}} &
 \fbox{\includegraphics[width=2.1cm]{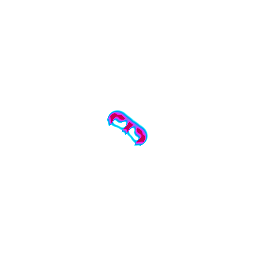}
 \includegraphics[width=2.1cm]{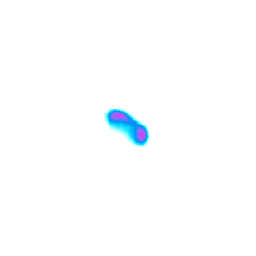}} \\


 \fbox{\includegraphics[width=2.1cm]{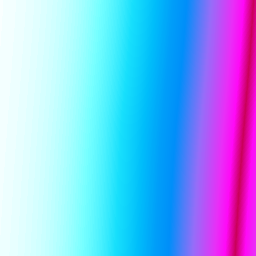}
 \includegraphics[width=2.1cm]{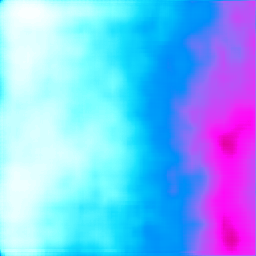}} &
 \fbox{\includegraphics[width=2.1cm]{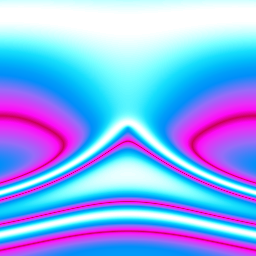}
 \includegraphics[width=2.1cm]{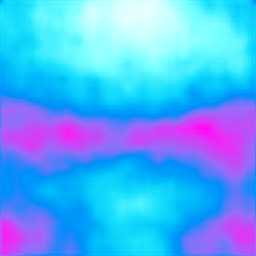}} &
 \fbox{\includegraphics[width=2.1cm]{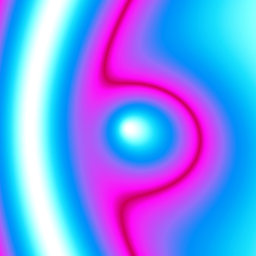}
 \includegraphics[width=2.1cm]{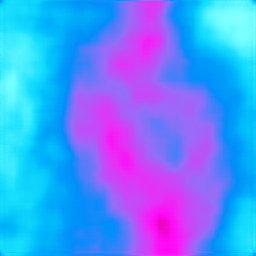}} \\

 
\end{tabular}

\vspace{0.2cm}

 b) Reconstruction Examples of the $\beta$-VAE used for the IMGEP-OGL
\vspace{0.1cm}

\setlength\tabcolsep{2pt}
\renewcommand{\arraystretch}{1.0}
\begin{tabular}{@{}ccc@{}}

 \fbox{\includegraphics[width=2.1cm]{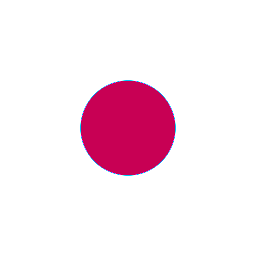}
 \includegraphics[width=2.1cm]{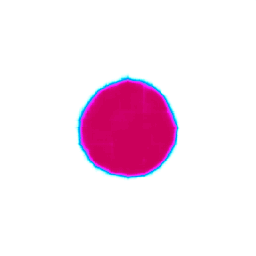}} &
 \fbox{\includegraphics[width=2.1cm]{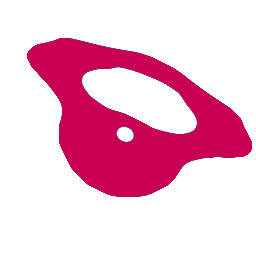} 
 \includegraphics[width=2.1cm]{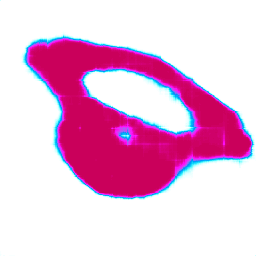}} &
 \fbox{\includegraphics[width=2.1cm]{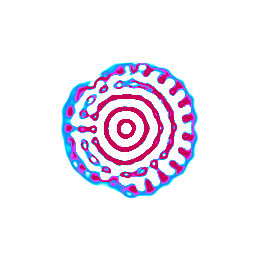}
 \includegraphics[width=2.1cm]{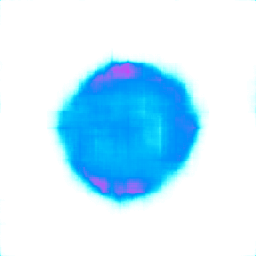}} \\


 \fbox{\includegraphics[width=2.1cm]{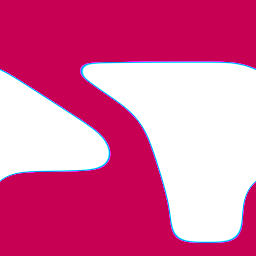}
 \includegraphics[width=2.1cm]{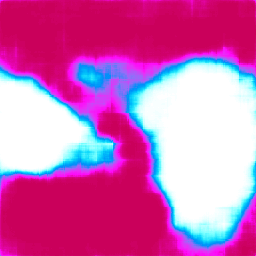}}  &
  \fbox{\includegraphics[width=2.1cm]{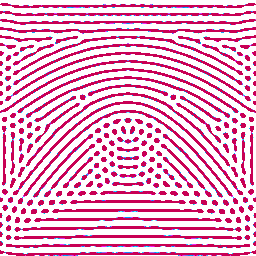}
 \includegraphics[width=2.1cm]{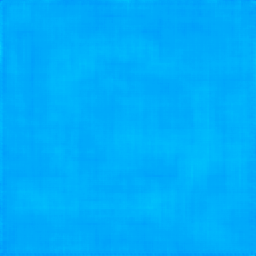}} &
 \fbox{\includegraphics[width=2.1cm]{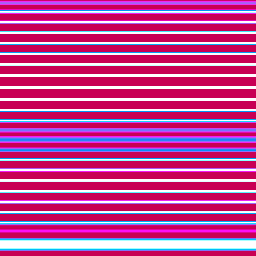}
 \includegraphics[width=2.1cm]{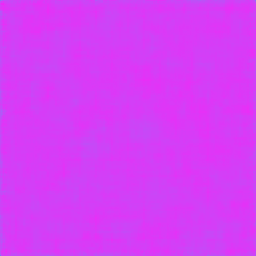}} \\
 
 \end{tabular}
 
 \caption{
 Examples of patterns (left) and their reconstruction (right) by $\beta$-VAE networks used for the IMGEP-PGL (a) and OGL (b). 
 The patterns are sampled from their validation dataset.
 For the PGL (a) the dataset is composed of animal patterns (row 1) from \cite{chan2019lenia} and randomly generated CPPN patterns (row 2). 
 For the OGL (b) it is composed of animal patterns (row 1) and non-animal patterns (row 2).
 } 
 \label{fig:IMGEP_reconstruction_examples}
\end{figure}

The $\beta$-VAE learning saturates for both the precollected dataset and the online collected dataset (Fig.~\ref{fig:IMGEP-PGL_training}).
Their ability to reconstruct patterns based on the encoded latent representation is also qualitatively similar.
For both datasets the $\beta$-VAEs are able to learn the general form of the activity pattern (Fig.~\ref{fig:IMGEP_reconstruction_examples}).
Nonetheless, the compression of the images to a 8-dimensional vector results in a general blurriness in the reconstructed patterns.
As a result, the $\beta$-VAEs are not able to encode finer details and textures of patterns.

\subsection{Measurement of Diversity in the Analytic Parameter and Behavior Space}
\label{c:sm_definition_analytic_spaces}

\subsubsection{Diversity Measure}
\label{c:sm_diversity_measure}

Diversity is measured by the area that explored parameters or observations cover in their respective spaces.
The parameter space consisted of the initial start state of Lenia ($A^{t=1} \in [0,1]^{256\times256}$) and the settings for Lenia's dynamics ($R,T,\mu,\sigma,\beta_1,\beta_2,\beta_3$).
The space consist therefore of $256^2$ dimensions, each for a single grid cell of the initial pattern, plus 7 dimensions for the dynamic settings.
The observation space consists of the final patterns $A^{t=200} \in [0,1]^{256\times256}$ resulting in $256^2$ dimensions for the space.
Each single exploration results in a new point in those spaces.

\begin{figure}[H]
\centering
 \includegraphics[width=10cm]{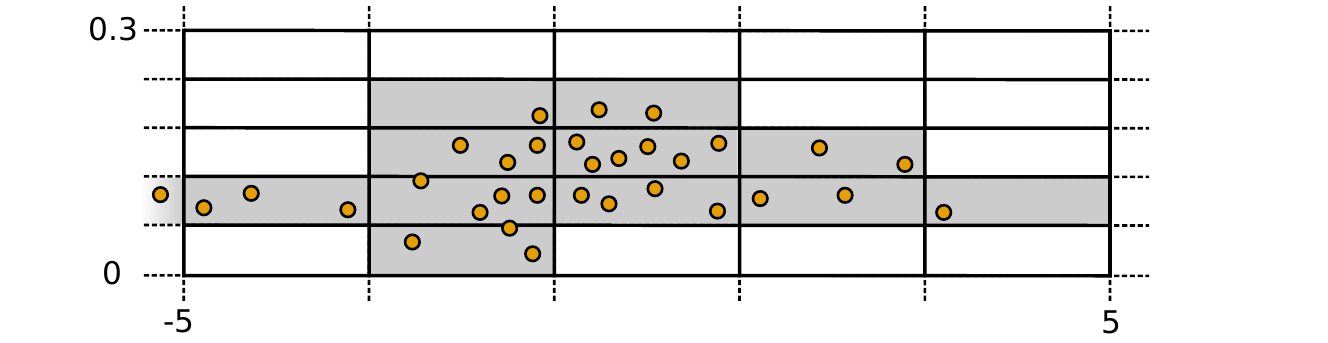}
  \caption{
 Illustration of the diversity measure in a two-dimensional space.
 Ranges for the dimensions are $[-5, 5]$ and $[0, 0.3]$. 
 The number bins per dimension is 7 resulting in $7^2=49$ discretized areas.
 The diversity is the number of areas (here 12) in which points exist (grey areas).
   } 
 \label{fig:diversity_measure}
\end{figure}

The diversity measures how much area the algorithms explored in those spaces (Fig.~\ref{fig:diversity_measure}).
The measurement is done by discretizing the space with a spatial grid and counting the number of discretized areas in which at least one point falls.
For the discretization a minimum and maximum border is defined for each space dimension.
Each dimension is then split in equally sized bins between those borders.
The areas with values falling below the minimum or above the maximum border are counted as two additional bins.

The number of dimensions of the original parameter and observation space are too large to measure diversity, because the initial and the final pattern have $256^2$ dimensions.
We constructed therefore an analytic parameter and behavioral space where the latent representations of a $\beta$-VAE were used to reduce the high-dimensional patterns to 8 dimensions.
5 bins (7 with the out of range values) per dimension were used for the discretization of those spaces for all experiments in the paper.

\subsubsection{Analytic Parameter and Behavior Space}

\begin{table}[b!]
    \centering

    \begin{tabular}{ c c c }
    
        Analytic Parameter Space Definition & ~ & Analytic Behavior Space Definition \\
        
        \begin{tabular}{ l c c }
            \toprule
            Parameter & min & max \\ 
            \midrule
            R  & $1$ & $20$ \\
            
            T & $2$ & $10$ \\
            
            $\mu$ &  $0$ & $1$ \\
            
            $\sigma$ & $0$ & $0.3$ \\
            
            $\beta_1, \beta_2, \beta_3$ & $0$ & $1$ \\
            
            $\beta$-VAE latent $1$ to $8$ (trained & \multirow{ 2}{*}{-5} & \multirow{ 2}{*}{5}\\
            on intial states $A^{t=1}$) &  & \\
            
            \bottomrule
        \end{tabular}
        &
        ~
        &
        \begin{tabular}{ l c c }
            \toprule
            Parameter & min & max \\ 
            \midrule
            mass $M_A$  & 0 & 1 \\
            
            volume $V_A$ & 0 & 1 \\
            
            density $D_A$ & 0 & 1 \\
            
            asymmetry $A_A$ & -1 & 1 \\
            
            centeredness $C_A$ & 0 & 1 \\
            
            $\beta$-VAE latent $1$ to $8$ (trained & \multirow{ 2}{*}{-5} & \multirow{ 2}{*}{5}\\
            on final patterns $A^{t=M}$) &  & \\

            \bottomrule
        \end{tabular}
        \\
    \end{tabular}
    
\caption{Features of the analytic parameter and behavior space with their min and max values.}
    \label{tbl:analytic_space_parameters}

\end{table}

 

\begin{figure}[t!]
\centering

a) Reconstruction Examples of the Analytic Parameter Space $\beta$-VAE
\vspace{0.1cm}

\setlength\tabcolsep{2pt}
\renewcommand{\arraystretch}{1.0}
\begin{tabular}{@{}ccc@{}}

 \fbox{\includegraphics[width=2.1cm]{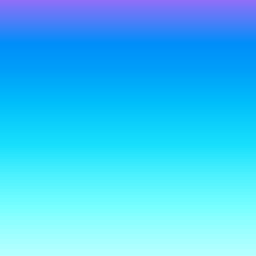} 
 \includegraphics[width=2.1cm]{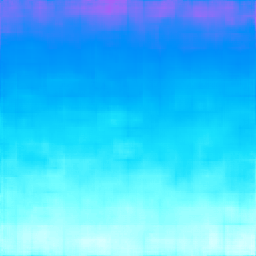}} &
 \fbox{\includegraphics[width=2.1cm]{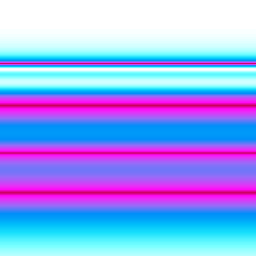}
 \includegraphics[width=2.1cm]{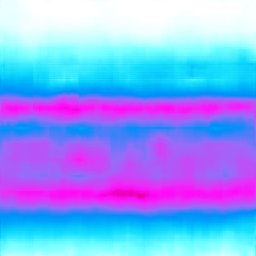}} &
 \fbox{\includegraphics[width=2.1cm]{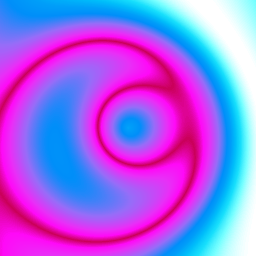}
 \includegraphics[width=2.1cm]{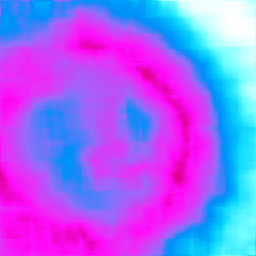}} \\

\end{tabular}

\vspace{0.1cm}

b) Reconstruction Examples of the Analytic Behavior Space $\beta$-VAE
\vspace{0.1cm}

\setlength\tabcolsep{2pt}
\renewcommand{\arraystretch}{1.0}
\begin{tabular}{@{}ccc@{}}

 \fbox{\includegraphics[width=2.1cm]{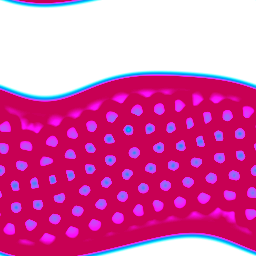} 
 \includegraphics[width=2.1cm]{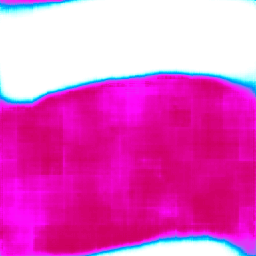}} &
 \fbox{\includegraphics[width=2.1cm]{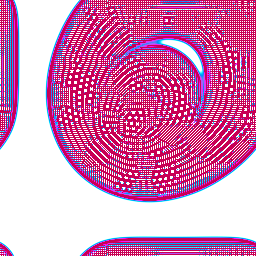}
 \includegraphics[width=2.1cm]{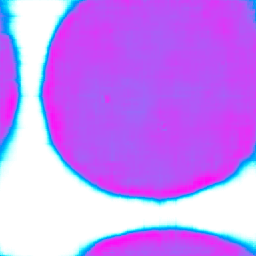}} &
 \fbox{\includegraphics[width=2.1cm]{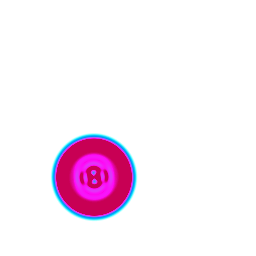}
 \includegraphics[width=2.1cm]{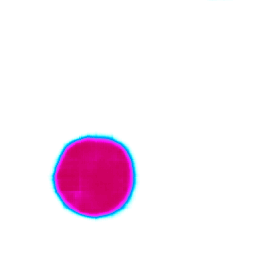}} \\

\end{tabular}

 \caption{
 Examples of patterns (left) and their reconstruction (right) by the $\beta$-VAE used for the analytic parameter (a) and behavior space (b). 
 The patterns are sampled from the validation dataset.
 } 
 \label{fig:analytic_parameter_space_reconstruction_examples}
\end{figure}

The analytic parameter space was constructed by the 7 Lenia parameters that control its dynamics and 8 latent representation dimensions of a $\beta$-VAE (Table~\ref{tbl:analytic_space_parameters}).
The $\beta$-VAE was trained on initial patterns $A^{t=1}$ used during the experiments.
The analytic behavior space was constructed by combining the 5 statistical measures for final Lenia patterns (Section~\ref{c:lenia_statistical_measures}) and also 8 latent representation dimensions of a $\beta$-VAE (Table~\ref{tbl:analytic_space_parameters}).
This $\beta$-VAE was trained on final patterns $A^{t=200}$ observed during the experiments.
The datasets for both $\beta$-VAEs were constructed by randomly selecting 42500 patterns (37500 as training set, 5000 as validation set) from the experiments of all algorithms and each of their 10 repetitions.
The $\beta$-VAEs used the same structure, hyper-parameters, loss function and learning algorithm as described in Section~\ref{c:sm_imgep_learned_goalspaces}. 
They were trained for more than 1400 epochs.
The encoder which resulted in the minimal validation set error during the training was used.
According to their reconstructed patterns they can represent the general form of patterns but often not individual details such as their texture (Fig.~\ref{fig:analytic_parameter_space_reconstruction_examples}).


    
        
        
        
        
        
        
        



\begin{figure}[b!]
\centering
\setlength\tabcolsep{1pt}
\begin{tabular}{@{}cc@{}}
 (a) Diversity in Parameter Space 
 &
 (b) Diversity in Behavior Space 
 \\
 \includegraphics[width=0.49\linewidth]{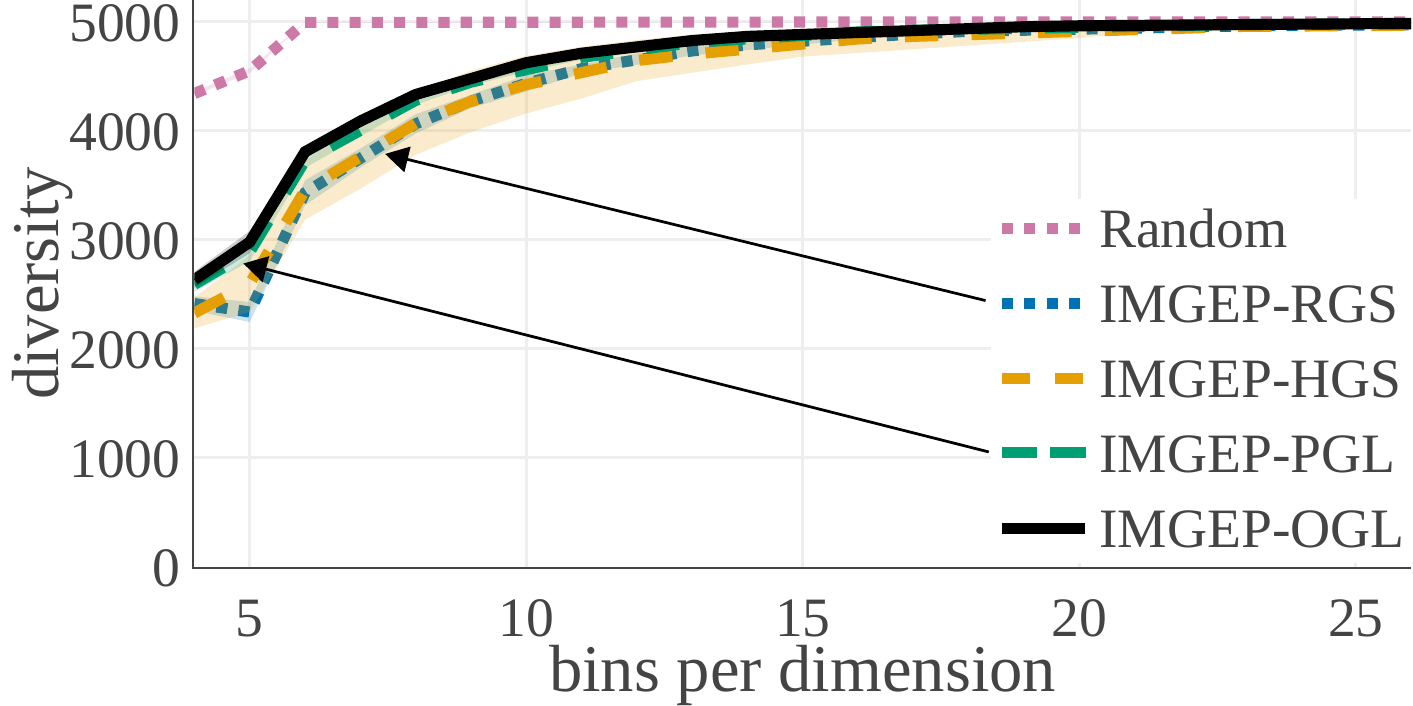} 
 &
 \includegraphics[width=0.49\linewidth]{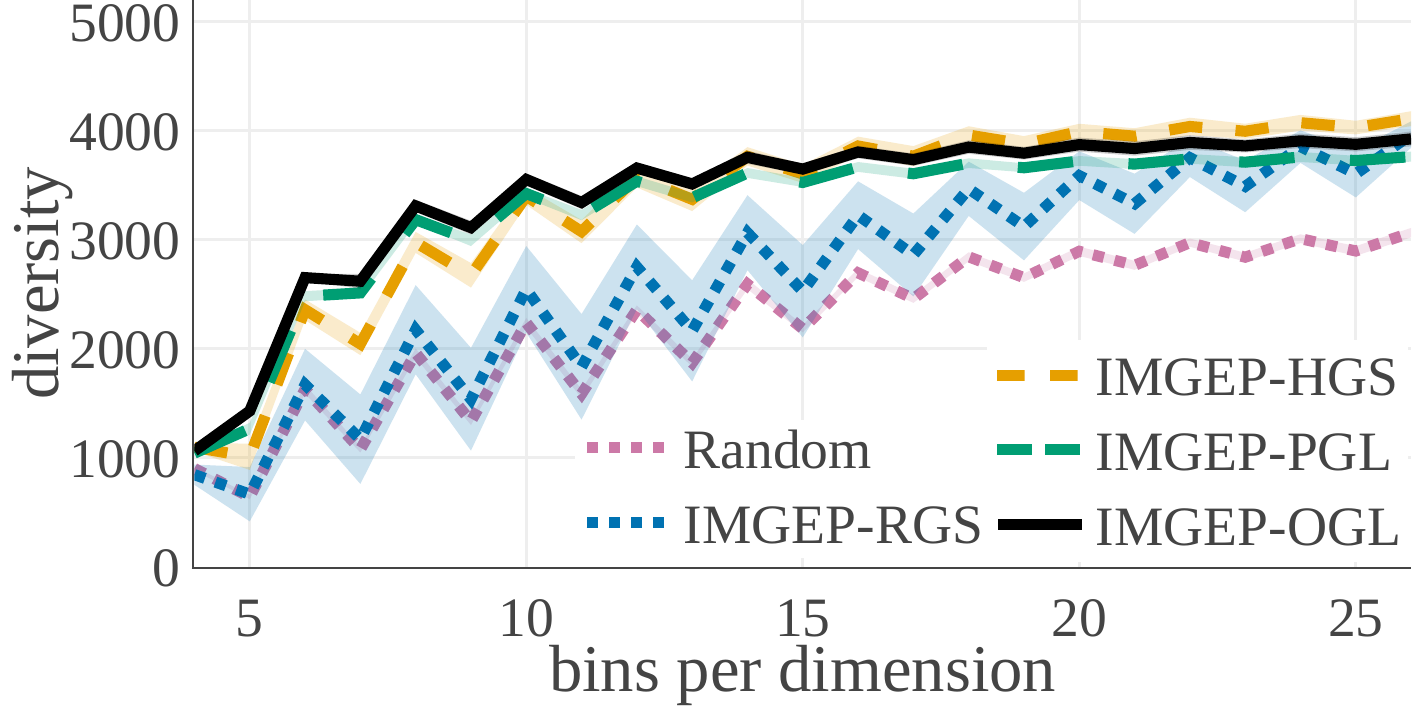} \\
 
 (c) Behavior Space Diversity for Animals \vspace{0.1cm}
&
(d) Behavior Space Diversity for Non-Animals
\\

 \includegraphics[width=0.49\linewidth]{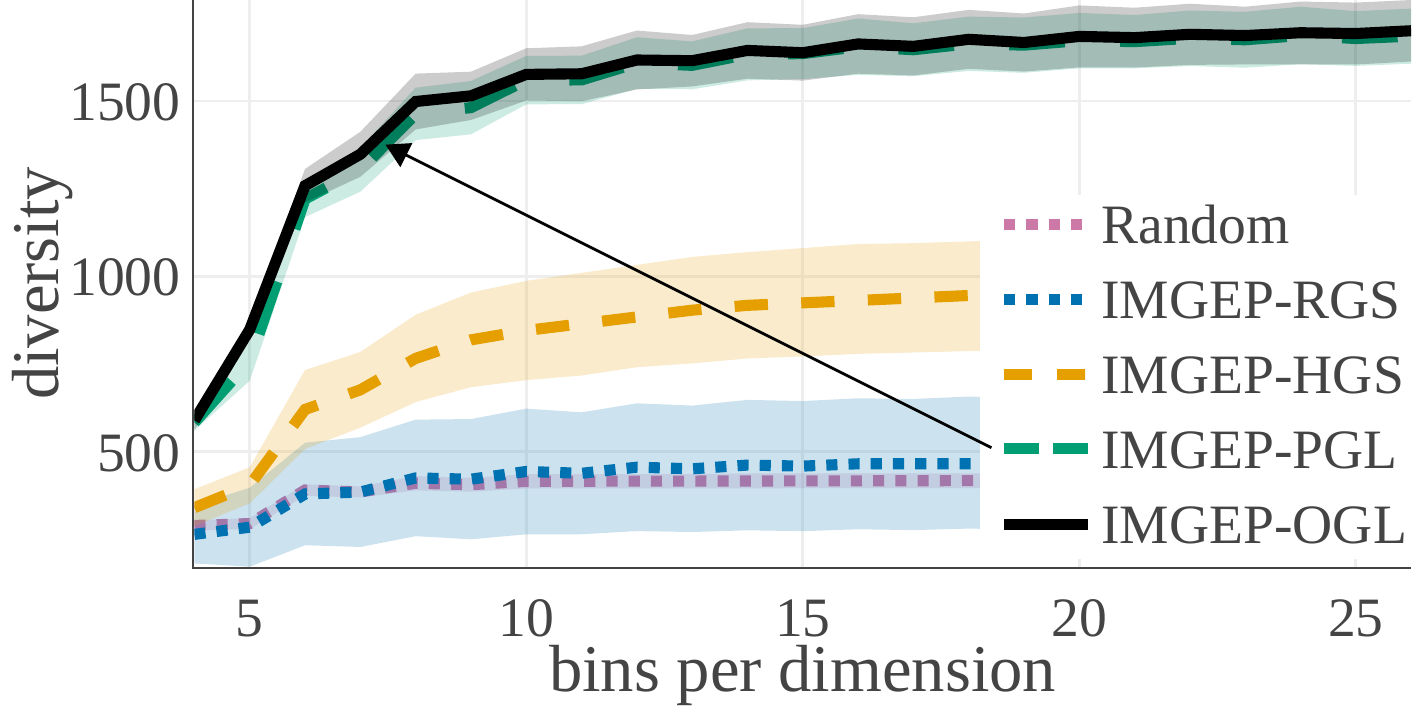}
 &
 \includegraphics[width=0.49\linewidth]{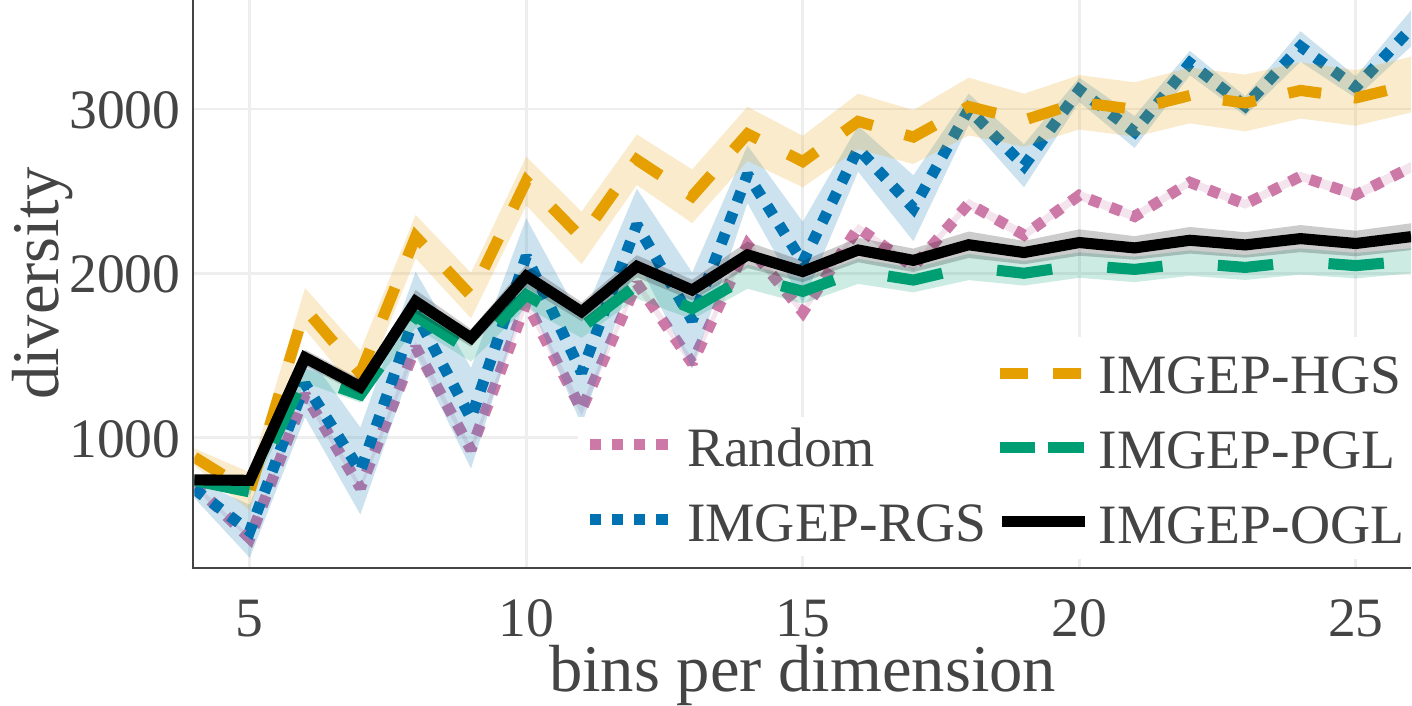}\\

\end{tabular}
 \caption{Influence of the number of bins per dimensions on the diversity measure.
 Depicted is the average diversity ($n=10$) with the standard deviation as shaded area.
 }
 \label{fig:results_diversity_dependend_on_binnumber_animal_nonanimal}
\end{figure}

\subsubsection{Dependence of the Diversity on the Number of Bins per Dimension}

The diversity measure has as parameter the number bins per dimension.
We analyzed the influences of this parameter on the diversity measure
(Fig.~\ref{fig:results_diversity_dependend_on_binnumber_animal_nonanimal}).
Although the diversity difference between algorithms depends on the number of bins per dimension for each space, the order of the algorithms is generally invariant to it.
Only if the number of bins per dimension grows large (\textgreater 10) the order of the algorithms changes in some cases.
The order starts to follow the same order as seen for the proportion of identified patterns (Fig.~\ref{fig:results_number_of_identified_patterns}).
In this case the discretization of the space becomes too fine.
Each pattern falls into its own discretized area. 
We chose therefore a smaller number of bins per dimension of 7 to compare the algorithms in a meaningful way.

\end{document}